\RequirePackage{latexml}
\iflatexml
  \documentclass{article}
  \usepackage{amsmath,amssymb,amsthm,array,authblk}
\usepackage[numbers,square]{natbib}
\usepackage{hyperref}
\providecommand{\Needspace}[1]{}
\newtheorem{theorem}{Theorem}
\newtheorem{lemma}[theorem]{Lemma}
\newtheorem{proposition}[theorem]{Proposition}
\newtheorem{corollary}[theorem]{Corollary}
\newenvironment{keywords}{\par\noindent\textbf{Keywords:} }{\par}
\newif\ifanonsubmission
\ifdefined\COLTreview\anonsubmissiontrue\fi
\title{Polynomial-Time Learning of Random Automata from One Label per Word}
\ifanonsubmission
  \author{Anonymous}
\else
  \author[1]{Yilin Cheng}
  \affil[1]{Fudan University, Shanghai, China\\Zhongguancun Academy, Beijing, China}
\fi
\date{}

\else
\ifdefined\COLTreview
\documentclass[anon]{colt2026}
\else
\documentclass{colt2026}
\fi
\usepackage{times}
\usepackage{array,needspace}
\jmlrvolume{}
\jmlryear{2026}
\jmlrpages{}
\jmlrproceedings{Preprint,}{Preprint,}
\renewcommand{\copyright}{{\fontfamily{lmr}\selectfont\textcopyright}}
\jmlrworkshop{}
\editor{}
\setcitestyle{numbers,square}
\allowdisplaybreaks[1]
\title[Polynomial-Time Learning of Random Automata from One Label per Word]{Polynomial-Time Learning of Random Automata\\from One Label per Word}
\ifanonsubmission
\jmlrproceedings{Submission,}{Submission,}
\hypersetup{pdftitle={Polynomial-Time Learning of Random Automata from One Label per Word},pdfauthor={}}
\else
\coltauthor{\Name{Yilin Cheng}\\
\addr Fudan University, Shanghai, China\\
\addr Zhongguancun Academy, Beijing, China}
\hypersetup{pdftitle={Polynomial-Time Learning of Random Automata from One Label per Word},pdfauthor={Yilin Cheng}}
\fi
\definecolor{paperlink}{rgb}{0.12,0.25,0.39}
\hypersetup{linkcolor=paperlink,citecolor=paperlink,urlcolor=paperlink}
\fi
\newcommand{\PrA}{\mathbb P_A}
\newcommand{\E}{\mathbb E}
\newcommand{\ind}{\mathbf 1}

\newcommand{\LeanArtifactURL}{https://github.com/Yilin-Cheng23/random-automata-lean/tree/aa71e6964fbe26627926770155acf7244a7b8d51}

\newtheorem{fairlabelcor}{Corollary}

\begin{document}
\maketitle

\begin{abstract}
We study a binary deterministic finite automaton whose $2n$ transitions
are chosen independently and uniformly from its $n$ states. The learner
observes independent uniform words of length $m$, each labeled with a
single bit indicating acceptance. For every fixed $\beta>0$ we give a polynomial-time algorithm that works
outside a set of transition tables of probability $o(n^{-\beta})$.
On each remaining table, for any start state and accepting set, the
predictor has error at most $\epsilon$ on fresh uniform words of length $m$
with probability at least $1-\gamma$ over the training sample and the
learner's random choices. Its sample size and
running time are polynomial in $n$, $m$, $1/\epsilon$ and $\log(2/\gamma)$,
with degree depending only on $\beta$. Although each word reveals only one output, comparing label moments at two
consecutive history lengths lets us estimate the expected product of the
signed outputs of two continuations from the same hidden state.

We also refute Conjecture~4 of Fish and Reyzin on meeting times.
For two distinct states fixed before the table is drawn and every fixed $c>0$, the
fraction of words of length $\lfloor cn\rfloor$ that send them to different
states converges in probability to $e^{-c}$. On a typical table, driving this
fraction down to $1/n$ takes asymptotically $n\log n$ letters.
\ifanonsubmission\else Selected proof steps are
\href{\LeanArtifactURL}{formalized in Lean}.\fi
\end{abstract}

\section{Introduction}
Consider a deterministic finite automaton that reads a word one letter at a
time. A passive learner observes the input word and whether it is accepted,
but none of the states visited along the way. Fish and Reyzin~\cite{FR} asked
how efficiently such a learner can predict acceptance when the transition
table is random and the input words are uniform. For a fixed start state,
exhaustive search ranges over $n^{2n}2^n$ choices of the transition table and
the accepting set. For polynomial word lengths $m=\Omega(n)$, their
Proposition~5 gives a conditional $2^{O(n)}$-time algorithm at
inverse-polynomial error: if a common suffix of linear length makes any two
trajectories coalesce with sufficiently high probability, then Fourier
coefficients supported on that suffix suffice for prediction.

Our polynomial-time learner takes a different route. Its reconstruction
step needs the responses to two continuations from the same hidden state,
yet each training word reveals the response to just one. Our learner estimates the expected product of two such
signed outputs by comparing label moments at two consecutive history lengths.
The meeting-time law of Section~\ref{sec:front-pairs} also explains why the
synchronization premise of Proposition~5 fails: after $\lfloor cn\rfloor$
letters, a fraction tending to $e^{-c}$ of all words still send two
prescribed starts to different states, for every fixed $c>0$.

\paragraph{Model.}
The $2n$ entries of $f_0,f_1:[n]\to[n]$ are independent and uniform. For a
word $w=w_1\cdots w_m$ we write $f_w(x)$ for the state reached from $x$ after
reading $w$. Let $b:[n]\to\{0,1\}$ be the indicator of the accepting set,
$\xi=2b-1$ its sign encoding, and $s$ the designated initial state. At
training length $m$, each example is an independent pair
\[
                (W,b(f_W(s))),\qquad W\sim{\rm Unif}(\{0,1\}^m).
\]
All examples share the same transitions and accepting set. For a fixed
labeled target $M$, the risk of a predictor is
$R_{\ell,M}(\widehat f)=\Pr_{|V|=\ell}[\widehat f(V)\ne b(f_V(s))]$, where $V$
is a uniform word of length $\ell$. Guarantees that are uniform over
accepting sets allow $b$ to be chosen after the table is drawn; in the
appendices the vector of statewise bits is called an accepting mask.
Logarithms are natural unless a base is indicated.

\begin{theorem}[Learning at the observed length]\phantomsection\label{thm:poly-learning}
For every fixed $\beta>0$ there is a passive algorithm $\mathcal A_\beta$
such that, for each $n$, a set $\mathcal G_{n,\beta}$ of transition tables
of probability $1-o(n^{-\beta})$ has the following property.
Fix any table in $\mathcal G_{n,\beta}$, any accepting set and start,
any $m\ge1$ and any $\epsilon,\gamma\in(0,1/2)$.
The only inputs to $\mathcal A_\beta$ are $n,m,\epsilon,\gamma$ and
independent examples $(W,b(f_W(s)))$ as above. With probability at least
$1-\gamma$ over the training sample and the learner's random choices, it outputs a predictor $\widehat f$
with $R_{m,M}(\widehat f)\le\epsilon$. Its sample size and bit running
time are polynomial in $n,m,1/\epsilon,\log(2/\gamma)$, with degree
depending only on $\beta$.
\end{theorem}
Taking $\gamma=n^{-\beta}/2$ gives joint failure probability at most
$n^{-\beta}$ over the table, the sample and the learner's random choices for all sufficiently large $n$. Training and
test words always have the same length $m$.

\begin{fairlabelcor}[Fair accepting bits]\phantomsection\label{cor:fair-baseline}
There is an absolute $c>0$ such that for each fixed integer $D\ge1$
there is a two-batch learner $\mathcal A'_D$ with the following guarantee.
For each prescribed start, $1\le m\le n^D$ and
$\epsilon,\gamma\in(0,1/2)$, it receives only $n,m,\epsilon,\gamma$ and
independent terminal-labeled words. With independent fair accepting bits,
its probability of returning a predictor of risk greater than $\epsilon$
is at most $O_D(n^{-c})+\gamma$ over the table, the accepting bits, the training sample and any random choices of the learner.
Its sample size and bit running time are polynomial in
$n,1/\epsilon,\log(2/\gamma)$ for fixed $D$.
\end{fairlabelcor}
Appendix~\ref{sec:quantified-target} proves this baseline directly. Its
sample bound is far from practical: even at $m=n$, the size of the first
batch has degree $573722$ in $n$, up to logarithmic factors
(Section~\ref{sec:window-cost}).

\paragraph{How the learner uses one label.}
Fix a short history and two continuation tests. For each test we estimate
the mean signed label among words ending in that history followed by the test.
The product of the population means corresponds to two independent hidden
starts reading the same history and then their respective tests. We average
these products over histories, first at one length and then at the next.
Pairs that remain distinct contribute in nearly the same proportion at the
two lengths. A weighted difference therefore estimates the expected product
of the tests on a \emph{single} state (Section~\ref{sec:front-moments}). These products suffice to separate
recurrent states with different future labels (Section~\ref{sec:front-core}).
A separate difficulty arises when the table has a small closed component,
which can absorb the run early in the word; Section~\ref{sec:front-transfer}
shows how to retain the relevant information from the input prefix in that
case. Appendix~\ref{sec:iid-clean-completion} combines the two cases under
the original random-table law and specifies the window lengths, precisions
and sample sizes.

\section{Meeting times under a common word}\label{sec:front-pairs}
Let $\tau$ be the first time at which a common random word maps two
prescribed distinct initial states to the same state. Conditional on the
table, let $q_i(t)=\Pr_W(\tau>t\mid A,i)$ be the fraction of length-$t$ words
that send the two starts to different states. In the binary i.i.d. model,
for fixed $c\ge0$ and $\beta>0$,
\begin{equation}\label{eq:front-pair-statement}
 q_i(\lfloor cn\rfloor)\ \xrightarrow{\PrA}\ e^{-c},\qquad
 q_i(\lfloor\beta n\log n\rfloor)=n^{-\beta+o_{\Pr}(1)}.
\end{equation}
Both limits are in probability over the transition table. On a typical
table, once a logarithmic initial window has passed, a prescribed pair
that is still distinct meets on the next letter with conditional probability
$(1+o(1))/n$. A pair chosen because its two transition rows coincide meets
on the first letter. Quattropani and
Sau~\cite[preprint, Thm.~2.2 and Cor.~2.5]{QS} proved a geometric meeting law
for two walks driven by \emph{independent} words, on a table with distinct
outgoing targets in each row. Their law refutes the linear-length assertion
for independent walks, and they also simulated the common-word case.
Equation~\eqref{eq:front-pair-statement} concerns the common input of Fish
and Reyzin's conjecture, in which both walks reuse the same transitions;
Appendix~\ref{sec:tails} proves it. The learning algorithm needs a related
quantity, namely the distribution of the pairs that survive a logarithmic
window, and the next section explains how terminal labels give access to it.

\section{Extracting same-state products from terminal labels}\label{sec:front-moments}
Write $\pi$ for the stationary distribution on the large recurrent component,
$T_a g(x)=g(f_a(x))$, and $m_v(g)=\pi T_vg$. For $g=T_u\xi$, $m_v(g)$ is the expected terminal sign when a state drawn
from $\pi$ reads $vu$. In the training data, conditioning on the suffix
$vu$ leaves a uniform random prefix from the prescribed start. If that
prefix is long enough for mixing, the conditional mean approximates
$m_v(g)$ to the prescribed inverse-polynomial accuracy. For short training
lengths the learner tabulates full words. The reconstruction step needs
$m_v(gh)$, where both tests act on the same hidden state.

On a fixed table define
\[
 S_l=\mathbb E_{|w|=l}(\pi T_w)^\top(\pi T_w),\qquad
 J_l(v;g,h)=\mathbb E_{|w|=l}m_{wv}(g)m_{wv}(h).
\]
The learner estimates the two means in each product separately. The
off-diagonal row $u_l$ of $S_l$ evolves under the substochastic kernel $Q$ on
ordered distinct pairs driven by a common letter, while both marginals of
$S_l$ equal $\pi$. Put $x_l=(u_{l+1}\mathbf1)/(u_l\mathbf1)$ and
$q_l=1-x_l$. If, over one step, the off-diagonal row changes mainly in its
total mass and little in its shape, then subtracting $x_lS_l$ from $S_{l+1}$
leaves essentially the diagonal. Precisely,
\[
 \left\|\frac{S_{l+1}-x_lS_l}{q_l}
                  -\operatorname{diag}(\pi)\right\|_{\rm ent}
 \le {2\|u_{l+1}-x_lu_l\|_1\over q_l},
\]
where $\|\cdot\|_{\rm ent}$ is the sum of absolute entries. The factor of two accounts for the off-diagonal remainder and the diagonal correction needed to preserve the marginals. For a sufficiently long logarithmic
$l$ the right-hand side is below any prescribed inverse polynomial, while
$q_l\asymp1/n$. Applying this to $T_vg$ and $T_vh$ gives the multiplication
rule of Figure~\ref{fig:windows}.
\begin{figure}[ht]
\centering
\begin{tabular}{@{}lc@{}}
History before $v$ & Product of two terminal means \\
$|w|=l$ & $J_l=\mathbb E_w[m_{wv}(g)m_{wv}(h)]$ \\
$|w|=l+1$ & $J_{l+1}=\mathbb E_w[m_{wv}(g)m_{wv}(h)]$ \\
\hline
$(J_{l+1}-x_lJ_l)/(1-x_l)$ & $\displaystyle\longrightarrow\quad
\pi T_v(gh)=\sum_z(\pi T_v)_z\,g(z)h(z)$
\end{tabular}
\caption{Two windows on the same transition table. Above the line, each
product of means corresponds to two independent stationary starts that read
the same history $w$ and suffix $v$. Weighting and subtracting the two rows
cancels the pairs that are still distinct, up to the remainder bounded in the
text. Below the line, both tests act on a single state $z$.}
\label{fig:windows}
\end{figure}

The finite-sample cost of this rule is governed by $q_l$. If both estimates
of $J$ are within $\eta$ of their population values and the clipped ratio
$\widehat x\in[0,1]$ is within $d\le q_l/2$ of $x_l$, then the quotient
changes by at most $4\eta/q_l+4d/q_l^2$. Indeed, its denominator is at least
$q_l/2$, the change caused by replacing $x_l$ alone is
$(\widehat x-x_l)(J_{l+1}-J_l)/[(1-\widehat x)q_l]$, and both values of $J$
lie in $[-1,1]$. Simultaneous accuracy of the primitive means suffices, even
when their word bins overlap. Appendix~\ref{sec:poly-learning-proof} bounds
the approximation at the population level.

The ratio $x_l$ can itself be estimated from the data. For the sign readout,
$B_l=J_l(\varnothing;\xi,\xi)$ satisfies $1-B_l=u_l(1-\xi_i\xi_j)$ because
$\xi_i^2=1$, so $(1-B_{l+1})/(1-B_l)$ estimates $x_l$. On cores with
nonconstant labels the mismatch mass in the denominator has an
inverse-polynomial floor, even when the accepting set is chosen after the
table is drawn (Appendix~\ref{sec:history-absolute-gap}); a constant core is
recognized from its terminal sign frequencies. Applying the multiplication
rule recursively estimates any fixed-degree product of short sign tests, and
every estimate uses only word bins of ordinary labeled words at the training
length.

\paragraph{Why the off-diagonal remainder is small.}
Consider two replicas of a common-word pair on the same exposed table, driven
by independent words. Once the four trajectories occupy four fresh states,
an ordered match of the two endpoint pairs requires, to leading order, two
collisions with already named states. If $i$ and $j$ letters remain to be
read after the two joins, the replica words must agree on their last
$\max(i,j)$ letters. Hence the coefficient of $1/n^2$ is
\[
         \sum_{i,j\ge0}2^{-\max(i,j)}
            =\sum_{j\ge0}(2j+1)2^{-j}=6 .
\]
The same coefficient appears for a single row of the pair block and for its
average row, so subtracting the two second moments controls the squared
Frobenius norm of the centered block. For this, both moment bounds must be
taken on the same table event. Conditioning on the entire exposed forest of
a successful transcript leaves all other cells independent, and two excess
edges in a later local search require two further hits; this yields the
matching lower bound (Lemma~\ref{lem:poly-centered}). Changing one outgoing
row perturbs the centered block only through the trajectories that read it,
so concentration over the independent rows upgrades the bound to any fixed
target confidence on tables without small terminal components
(Appendix~\ref{sec:centered-edits}).

\section{Recovering the recurrent responses}\label{sec:front-core}
Two states are predictively equivalent if every continuation produces the
same output from both. The learner first identifies these recurrent
equivalence classes, together with their transitions and accepting signs. At
this stage the classes are determined only up to relabeling, and the initial
state plays no role.

The features are products of tests that begin with different first letters.
A centered Gram estimate, combined with a bounded-degree graph of exceptional
pairs, provides for each recurrent predictive class a linear functional that
isolates its row up to a small error, and fixed tensor powers suppress the
remaining cross terms. The resulting dictionary matrix $V$, with one row per
class, has full row rank and $\sigma_{\min}(V)\ge n^{-C}$, where the exponent
depends on the prescribed confidence. The short-word class signatures used
below also separate all classes.

Rare cancellations require some care. Fix a logarithmic window and consider
a matching of $h$ pairs whose ordered successor rows differ but whose
survival probability is very small. Two consecutive binary forks force at
least $3h$ collisions with already named destinations, while there are at
most $n^{2h}$ candidate matchings; a separate count bounds the number of such
pairs incident to a single state. Consequently, outside a target exception
of probability $o(n^{-\beta})$, only boundedly many exceptional pairs with
unequal rows remain. Each of them either reaches the region with a positive
survival coefficient along a bounded path, or stays in this finite set until
it merges. In both cases, any difference between the response languages of
two recurrent classes is witnessed by a word of logarithmic length
(Appendix~\ref{sec:exception-pair-matching}).

\paragraph{From observable matrices to discrete states.}
List the sign tests on words of length at most $L+1$, where $L=C\log n$ is a
distinguishing depth. Let $s_j$ denote test $j$ evaluated on the recurrent
classes and $\omega$ their stationary weights. The multiplication rule
estimates
\[
 D=V^\top\operatorname{diag}(\omega)V,\qquad
 D_j=V^\top\operatorname{diag}(\omega s_j)V.
\]
For a proposed pattern $a_1,\ldots,a_d$ of initial signs, form
$K(a)=(d-\tfrac12)D-\sum_{j\le d}a_jD_j$. In the corresponding diagonal
factor, a class matching the pattern has coefficient $-1/2$ and every other
class has coefficient at least $3/2$. Hence an absent pattern gives a
positive semidefinite matrix, whereas for a realized pattern the
pseudoinverse applied to the matching class coordinate gives a negative
Rayleigh quotient of magnitude at least $\min\omega\,\sigma_{\min}(V)^2/2$.
A rational positive-definiteness test with a small known shift therefore
separates the two cases using the estimated matrices. At most $n$ branches of
the sign-pattern tree survive at each level. The sign of the empty test gives
the output of a class, and prepending a letter to every shorter test
identifies its successor. Once the transitions are decoded, prediction follows these discrete
transitions along the test word. There is no further multiplication of
estimated matrices to amplify their errors.

At the prescribed start, a short prefix determines where the run enters this
machine. For exact prediction of long words, on a table event whose exception
has probability $O(1/n)$, every $g$-letter prefix reaches the core, and the
remaining transient behavior of prefixes can be absorbed into a machine with
at most $n$ states (Appendix~\ref{sec:proper-exact}). For prediction with
positive error it suffices to discard a small fraction of prefixes and to
compare the candidate initial core states on fresh labeled examples. 

\section{Learning with small closed components}\label{sec:front-transfer}
With probability of order $1/n$, the table contains a state at which both
letters are self-loops, so a learner that promises failure probability
$n^{-2}$ must handle such tables. Terminal examples never reveal when the run
entered a closed component, and inside a larger closed component the final
state may keep moving. One can still learn the mean label by asking which
part of it can depend on the prefix after the run has entered the component.

Let $T$ be a closed set of $s$ states, and suppose for the moment that its
two internal maps are given, without state identifiers or outputs. For
suffixes of length $r$, let $E_r$ be the span of the constant function and
of the $s^2$ indicators that a specified state of $T$ reaches another
specified state. If the run is already inside $T$, its entire suffix
response lies in $E_r$. Revealing the next input bit enlarges this space from
$E_r$ to $L^2(\{0,1\})\otimes E_{r-1}$. For any normalized new orthogonal
direction $d$, the observable response $Yd(W_{j+1:m})$ has mean zero given a
prefix that ends inside $T$, and its second moment given the whole prefix is
at most one. The original task therefore reduces to $O((m+1)(s^2+1))$
regressions whose targets vanish on $T$, and their errors add in squared
loss.

In each such regression, an exterior state with appreciable residual signal
admits a short suffix with a detectable mean. Inserting a mixing block before
this suffix makes the observable conditional response proportional to the
one-state survival coefficient for avoiding $T$, and a threshold on this
coefficient identifies the states from which the residual becomes small
quickly. Two candidate hidden origins that agree on all tests then face three
possibilities along a common continuation: they merge, one reaches the
threshold before the other, or their remaining residual contributes little
to prediction. In the second case an observable short word separates them.
This yields an algorithm that follows the remaining distinctions using the
prefix of the actual test word (Appendix~\ref{sec:closed-edit-resilience}).

On a single clean base table the argument holds simultaneously for all such
modifications: $T$ may be chosen after inspecting the table, $|T|\le n^{1-a}$
for a fixed $a>0$, and the outgoing rows of $T$ may be replaced by arbitrary
maps into $T$. The learner needs the abstract internal action, or it can
enumerate all actions in time $s^{O(s)}$. In the i.i.d. model a prescribed
$s$-set is closed with probability $(s/n)^{2s}$. By saturating the small
closed sets and resampling only their outgoing rows, we show that outside a
table exception of probability $o(n^{-\beta})$ there is a clean completion
with $s$ bounded in terms of $\beta$, so the enumeration costs only a
constant depending on $\beta$. The full construction and its quantifiers are
given in Appendices~\ref{sec:stochastic-clean-stratum}--\ref{sec:iid-clean-completion}.

\begin{center}
\fbox{\begin{minipage}{.94\linewidth}
\textbf{Algorithm 1. Learning at the observed length.}
Fix $\beta$ and use the windows, tolerances and list caps of
Appendix~\ref{sec:binary-schedule}. The input consists of
$n,m,\epsilon,\gamma$ and terminal-labeled words.
\begin{enumerate}
\item Build candidates for the recurrent core from suffix-bin means:
calibrate the two-window difference with the terminal signs, form $D$ and
$D_j$, and decode the recurrent responses with the shifted sign-pattern tests
of Section~\ref{sec:front-core}, including the case of a constant response.
Fit the prefix predictions on fresh words.
\item In addition, try every binary action on at most $\lceil\beta+2\rceil$
states. For each trial, form the coarse suffix response and the orthogonal
detail tasks from the observed words and labels, fit and validate their
readouts separately, then add them, clip and threshold. When the prescribed
windows exceed the word length, use direct word bins instead.
\item Compare all resulting classifiers on fresh complete words and return
one with the smallest empirical classification error.
\end{enumerate}
\end{minipage}}
\end{center}

\paragraph{Where the confidence enters.}
Let $S_k$ be the event that every terminal component has more than $k$
states. For a fixed exponent $A$, the clean-table estimate takes the form
$\Pr(S_k\cap G_A^c)=o(n^{-A})$, where $G_A$ involves only the transitions.
For typical tables, the centered-block estimate follows from an elementary
search with one surplus arc. For the smaller failure probability,
Appendix~\ref{sec:centered-edits} allows a bounded number of surplus arcs and
then uses concentration under independent row edits.
Appendix~\ref{sec:iid-clean-completion} applies this estimate with
$A=2\beta+10$ to a completion that keeps every exterior row, and
Lemma~\ref{lem:deletion-root-gap} is applied on the same completion at the
actual deleted set. Once the table is fixed, fresh labeled words suffice to
select a candidate with training failure probability $\gamma$.

\section{Window lengths and sample cost}\label{sec:window-cost}

For the baseline construction at $m=n$ with independent fair signs
(Appendix~\ref{sec:poly-learning-proof}), observing one specified suffix of
length $L$ costs a factor $2^L$ in the sample size. With
$d=\lceil10\log_2 n\rceil$, $R=7d+3$ and $t=4096R$, we have
$L=2t+4d+6=57348d+24582$, so the first batch has size
$O(n^{573722}(\log n+\log(8/\gamma)))$, where $\gamma$ is the training
failure probability (denoted $\delta$ in
Appendix~\ref{sec:poly-learning-proof}). The ridge matrix has
$p=2^{t+1}=O(n^{286720})$ rows, and a direct dense inversion takes
$O(p^3)=O(n^{860160})$ arithmetic operations. These figures refer to a single
batch and a single operation in this particular schedule. For
Theorem~\ref{thm:poly-learning} the exponents also depend on the cut-off for
small components, the block contraction and the precision required by the
decoder; Appendix~\ref{sec:binary-schedule} gives the cost of each stage.

\paragraph{Measuring the approximation at different lags.}
For each size $n$, we drew 24 independent tables and computed the population
error $\|(S_{t+1}-x_tS_t)/(1-x_t)-\operatorname{diag}\pi\|_{\rm ent}$.
Starting from $\pi\otimes\pi$, we propagated the pair distribution under
the two letter maps, averaging exactly over common words.
Table~\ref{tab:lag-panel} gives the median over tables with a unique closed
class, positive surviving mass and $q_t>10^{-12}$. We used the same draws
at all four lags.

\begin{table}[ht]
\centering\small
\caption{Population diagonal-extraction error (sum of absolute entries).
The last column gives the number of tables for which the error is defined at
each displayed lag.}
\label{tab:lag-panel}
\begin{tabular}{rrrrrr}
\hline
$n$ & $t=3$ & $t=10$ & $t=20$ & $t=40$ & Defined out of 24\\
\hline
6  & 0.261 & 0.00413 & $1.31\times10^{-5}$ & $1.14\times10^{-10}$ & 20 at all lags\\
10 & 1.15  & 0.0290  & $3.71\times10^{-4}$ & $7.75\times10^{-8}$ & 20 at all lags\\
16 & 1.72  & 0.121   & $2.57\times10^{-3}$ & $1.48\times10^{-6}$ & 22, 22, 22, 21\\
32 & 4.30  & 0.309   & $8.14\times10^{-3}$ & $6.33\times10^{-6}$ & 20 at all lags\\
64 & 8.77  & 0.732   & 0.0202          & $1.60\times10^{-5}$ & 24 at all lags\\
\hline
\end{tabular}
\end{table}

\paragraph{What the decoder recovers.}
Of 72 independent tables (24 each at $n=6,10,16$), 61 had a unique recurrent class
with nonconstant labels. We tested all 61 at every lag. To isolate decoding
from the conditioning argument in Section~\ref{sec:front-core}, we used
single sign tests up to length $\lceil\log_2 n\rceil$, with one extra
letter to identify signatures and successors. For the oracle shift
$(\min_x\omega_x)\sigma_{\min}(V)^2/4$, we took the masses and features
from the hidden table, summing the masses of identical signatures and
keeping one normalized row per signature. The pair and triple recursions
average over the common letters.
At lags $3,10,20,40,60,80,120$ the decoder recovered $4,30,58,58,58,58,58$
of the 61 targets, and with exact hidden moments it recovered all $61$ at the
same shift.

From lag 20 on, the same three cores failed the label-ratio guard
$1-x>10^{-5}$, and a backward search from the diagonal in each core's pair
graph shows that exactly these three are non-synchronizing. In draw
$(n,\mathrm{rep})=(6,18)$ the two letters act as the identity and a swap. In
draw $(6,14)$, merging two identical transition rows leaves a permutation
action on three states (the two merged states carry different labels). In
draw $(16,7)$ each letter exchanges the blocks $\{7,15\}$ and $\{8,10\}$, so
starts in opposite blocks never meet. The exact-moment comparison supplies the hidden moments directly.

A random binary table fails to synchronize with probability
$\Theta(1/n)$~\cite{Berlinkov}, enough to matter for every $\beta\ge1$.
On the clean event with no small terminal components,
\eqref{eq:global-rho} gives $\rho(Q)<1$. Every pair then merges under some
word, and successive mergers reduce the image to one state.
Appendix~\ref{sec:iid-clean-completion} bounds the exceptions on that
stratum. On the remaining tables covered by the guarantee, the learner
enumerates bounded closed actions and uses the observed prefix. The
experiment tests only the recurrent-core decoder.

\ifanonsubmission\else
\paragraph{Lean sources.}
The \href{\LeanArtifactURL}{Lean repository} contains 22 modules that
formalize selected proof steps, including the four-source coefficient and
the deterministic bounds used after deleting states.
\fi

\section{Related work}
\paragraph{Average-case learning of automata.}
Freund et al.~\cite{Freund} observe a label at every step of a random walk.
Angluin and Chen~\cite{AC} observe the state reached by each word, so their
statistical queries can estimate a vector of state probabilities. For
i.i.d. random transitions, their recovery bound for the recurrent component
assumes a full-rank observation matrix and depends on its pseudoinverse.
Here each word reveals one bit. The random transition table determines
whether the resulting moments distinguish recurrent responses and how
accurately they must be estimated.

Maranzatto~\cite[Secs.~5.2--5.4]{Mar} studies biased input walks. The
learner observes a label after each prediction and can reset at the cost of
one mistake. Section~5.4.2 of that work also uses two independent uniform
letter maps to study the cycles of one letter. Our learner observes just
the final label of each independent word and predicts at the same length.

\paragraph{Moment methods and learning from generated strings.}
Observable operator models and spectral algorithms~\cite{Jaeger,HKZ,BM}
use related moment factorizations. Here the common input to two trajectories
allows us to estimate a hidden product from terminal labels.
Angluin et al.~\cite{AEKR} prove statistical-query lower bounds for
target-dependent input distributions. Our input distribution is uniform at
a fixed length. Mossel and Roch~\cite{MR} show that learning general hidden
Markov models is at least as hard as learning parity with noise and give an algorithm under a
nonsingularity condition. Clark and Thollard~\cite{CT} learn the string
distribution generated by a probabilistic deterministic automaton, with
bounds depending on state distinguishability and the maximum expected string
length over states. In our problem the input distribution is known and the
unknown function assigns an acceptance bit to each word.

\section{Discussion and open problems}
The first batch in Section~\ref{sec:window-cost} is large because long
suffixes are rare and reconstruction amplifies errors in their estimated
means. Table~\ref{tab:lag-panel} gives small population errors at much
shorter windows on the tables tested. Can one prove a sharper bound on
$\|u_{l+1}-x_lu_l\|_1/(1-x_l)$ at those lengths? To reduce the sample
exponent, such a bound must meet the precision requirements of calibration
and decoding.

A second question is whether the polynomial degree can be chosen
independently of the target-confidence exponent $\beta$. Our enumeration
grows with the largest hidden closed component the learner must handle.
It is also open whether the two-window method recovers the recurrent
responses for alphabets with more than two letters.

\clearpage
\appendix
\renewcommand{\theHsection}{app.\arabic{section}}
\noindent\textbf{Proof route for Corollary~\ref{cor:fair-baseline}.}
\[
 \underbrace{\text{first meetings}}_{\text{Appendix A}}
 \ \longrightarrow\
 \underbrace{\text{moments and readout}}_{\text{Appendix~\ref{sec:poly-learning-proof}}}
 \ \xrightarrow{\ \text{length schedule (\ref{sec:poly-lengths})}\ }\
 \underbrace{\text{target failure }O_D(n^{-c})}_{\text{Appendix~\ref{sec:quantified-target}}}.
\]
The arbitrary-$\beta$ guarantee uses the separate completion argument in Appendix~\ref{sec:iid-clean-completion}.
\section{First meetings under a common word}
\paragraph{Two prescribed states.}
For $i=\{x,y\}$ with $x\ne y$, let $\tau$ be the first meeting time.
Write $q_i(t)=\Pr_W(\tau>t\mid A,i)$ for the surviving fraction of words.

\begin{theorem}[Shared-word meeting time]\phantomsection\label{thm:law}
For every prescribed distinct pair \(i=i_n\) and every fixed \(c\ge0\),
\[
 q_i(\lfloor cn\rfloor)\xrightarrow{\PrA} e^{-c},
 \qquad \frac{1}{n}\E_W[\tau\mid A,i]\xrightarrow{\PrA}1.
\]
At $t=\lfloor\beta n\log n\rfloor$ the survival probability is
$n^{-\beta+o_{\Pr}(1)}$ for every fixed $\beta>0$.
\end{theorem}

Already $\lambda=1$ in Conjecture~4 contradicts the fixed-$c$ limit.
A pair chosen for identical transition rows, however, meets on the first letter.

\subsection{First meetings in a logarithmic window}\label{sec:lag}
On distinct unordered pairs let $Q$ be the substochastic transition kernel with the diagonal killed. Put $f_t(i)=\Pr_W(\tau=t\mid A,i)$. We first estimate $f_r$ at a single lag, then use the survival recurrence to obtain bounds at later times. Set
\[
 \ell=\lfloor(\log_2 n)/5\rfloor,\quad d=\lceil10\log_2 n\rceil,\quad
 r=4d+1,\quad k=\lceil(\log n)^3\rceil,\quad M=2kr+2k+10 .
\]
A named vertex is a previously specified starting state, queried source or revealed destination. Across \(k\) replicas of \(r\) steps there are at most \(2kr\) cell queries; including the initial states and any prescribed terminal pair leaves the total named count below \(M\) on every continuation. Different replicas have independent words and the same table.

\begin{lemma}\phantomsection\label{lem:tame}
Except on an event of probability $O(n^{-1/5})$, from every vertex and for every set $H$ of size at most $M$, the probability that the next $d$ steps of one uniform-input walk all stay in $H$ is at most $n^{-3}$.
\end{lemma}
\begin{proof}
Let $G$ be the event that each outward breadth-first search revealing all arcs from vertices at distance $<\ell$ has at most one arc beyond its discovery tree. At most $N=2^{\ell+1}$ entries are probed per root. At each new probe the probability of hitting a named vertex is at most $(N+1)/n$. Two such probes cost at most $\binom N2((N+1)/n)^2$. Sum over the $n$ roots. On $G$ there are at most $\ell+1$ words of length $\ell$ from $x$ to a specified $y$: specifying the number of traversals of the sole extra arc specifies the intervening directed tree segments. This counts parallel labeled arcs as separate arcs. The probability to land in $H$ after $\ell$ steps is at most $M(\ell+1)2^{-\ell}\le n^{-1/6}$ for all large $n$. Apply the Markov property at successive block ends up to time $d$.
\end{proof}

Reveal table entries only at their first query, and expose input bits only on demand. A new probe is called a return when its destination is already named. Reading a previously revealed cell never counts as a return. Within a step use a fixed order and mark each destination immediately. An unqueried cell is uniform conditional on the revealed history. A trial starts from a \emph{safe} partial table if from every named vertex the chance of following only currently exposed entries for $d$ fair bits is at most $n^{-3}$. This property is computable from the partial table by killing the walk at an unknown entry. On $G$ every partial history considered here passes the test, by Lemma \ref{lem:tame}. Safety checks are performed on partial tables so that the unqueried cells retain their conditional product law.

\begin{lemma}\phantomsection\label{lem:acquire}
Call a distinct pair fresh when neither current vertex has ever been queried as a source. Start from a safe partial history for which the number of named vertices stays at most \(M\) on every continuation of the next \(2d\) steps. The probability of remaining unmerged through $2d$ steps without obtaining a fresh pair is at most
\[
 \eta=4dM/n+2n^{-3}+d^2 2^{-d}.
\]
\end{lemma}
\begin{proof}
Exclude a return in these steps, at cost at most $4dM/n$. Write $H$ for the initially named set, which contains both endpoints of every initially exposed arc. On this event, a path staying inside $H$ for its first $d$ steps must use only initially known entries. The probability of this happening to either coordinate, intersected with no return, is at most $2n^{-3}$ by safety.

Both coordinates now exit $H$ by time $d$. The first to exit discovers an extending simple path outside $H$. If the two exit together, their new destinations are distinct fresh sources. If the second exits on a new entry, both are again fresh. The only other exit uses the departure arc previously taken by the first walk: until the second exits this is the only exposed arc from $H$ to its complement. The second walk follows the first one's path with a strictly positive time lag. Until divergence it remains behind, and the leader keeps discovering new vertices. The first different input letter at a vertex of this path takes an unqueried entry, after which both vertices are fresh. This argument applies on survival \emph{through} the end of the window. Failure up to $2d$ therefore forces equality of length-$d$ substrings starting immediately after two distinct exit times in $\{1,\dots,d\}$. For fixed offsets even overlapping substrings agree with probability $2^{-d}$: each equality fixes the later bit from an earlier bit. Sum over the offsets using the unconditional distribution of the trial's future word given its initial history.
\end{proof}

\begin{lemma}\phantomsection\label{lem:upper}
Let $a=(1+1/\log n)/n$ and $L=(\log n)^{12}$. With high probability,
\[
 \max_i f_r(i)\le a,\qquad \max_{i,j}Q^r(i,j)\le L/n^2.
\]
In addition there exists \(b_n=1/n+O((\log n)^{10}/n^2)\) with \(\E[\ind_G f_r(i)^2]\le b_n^2\) for each prescribed \(i\).
\end{lemma}
\begin{proof}
Try to obtain a fresh pair during the first $2d$ steps, then during the next $2d$ steps if still necessary. Abort on an unsafe partial table at the start of any attempt actually made. A truncated success is an ordinary first meeting at step $r$ (or an ordinary surviving arrival at the specified pair, for the entry bound) for which no such abort occurs. Failure to acquire a fresh pair is charged below and does not itself abort a trial. Success under consideration requires survival until strictly before the last step. By iterated conditioning the cost of failing both attempts while remaining eligible is at most $\eta^2$. After acquisition there are at most $2r$ probes before the last step. Two or more returns among them have conditional probability at most $\binom{2r}{2}(M/n)^2$.

Up to a meeting, an initially fresh coordinate stays at fresh sources until a return on one of its own probes. With at most one return strictly before the final step, at least one of the two final probes is therefore fresh. Its target is independent uniform even conditional on the final letter, and on the other target when that entry is different. Given such a pre-step history with distinct current states, and given the final letter, the two table cells are distinct; equality has probability exactly \(1/n\). The restriction on returns ends before these final probes. The truncated one-replica success probability is bounded by
\[
 b_n=1/n+\eta^2+\binom{2r}{2}(M/n)^2.
\]
For landing at a particular distinct unordered pair, no pre-final return after acquisition leaves two fresh probes, giving probability at most $2/n^2$. One pre-final return has probability at most $2rM/n$ and leaves target probability at most $2/n$. Charging two returns or two failed attempts as before gives an upper bound $b'_n=O((\log n)^{10}/n^2)$.

For each fixed table, ordinary replica successes are independent over words. On $G$ ordinary successes pass every safety test, so $\E[\ind_G f_r(i)^k]$ is at most the probability of $k$ successive truncated successes. Iterating the uniform history-conditional bound gives $b_n^k$, without conditioning unused entries on $G$. Two replicas give the stated second moment. The same argument gives $\E[\ind_G (Q^r(i,j))^k]\le (b'_n)^k$. Apply Markov's inequality and union bounds of sizes at most $n^2$ and $n^4$; $n^2(b_n/a)^k=o(1)$ and $n^4(n^2b'_n/L)^k=o(1)$.
\end{proof}

\subsection{From first meetings to survival probabilities}\label{sec:tails}
Let $S_j=q_i(jr)$ and $u=ra$. On the event of Lemma \ref{lem:upper}, for $t\ge r$,
\[
 \Pr_i(\tau=t)=(Q^{t-r} f_r)(i)\le a q_i(t-r).
\]
Hence $S_{j+1}\ge S_j-uS_{j-1}$ for $j\ge1$. For a prescribed distinct pair, the expected probability of meeting by step $r$ is $O(r^2/n)$ by exposing the paths until the first repeated vertex. Write $\gamma=n^{-1/4}$; with high probability $S_1\ge1-\gamma$. The same bound applies to the pair reached by the two different first letters from $s$, treating equality of these destinations as immediate failure: expose those two entries first and charge a repeated named vertex throughout the exploration. For large $n$, $u/(1-\gamma)\le\gamma$. Whenever $S_j/S_{j-1}\ge1-\gamma$, the next ratio is at least $1-u/(1-\gamma)\ge1-\gamma$. Thus
\begin{equation}
 S_j\ge (1-\gamma)\big(1-u/(1-\gamma)\big)^{j-1},\qquad j\ge1. \label{eq:low}
\end{equation}

\begin{lemma}\phantomsection\label{lem:bad}
With high probability, the set $B=\{i:n f_r(i)<1-1/\log n\}$ has size at most $n^{2-1/10}$.
\end{lemma}
\begin{proof}
For each fixed $i$, require every destination up to time $r-1$ to be new, then require the first final destination to be new and the second to equal it. For a fixed word its probability is exactly \((n-2)_{2r-1}/n^{2r}\), where the numerator is a falling factorial: the two paths discover \(2r-1\) distinct new vertices through \(2r\) distinct cells. This is at least \((1-O(r^2/n))/n\). The revealed cells in each such transcript prescribe a directed graph whose underlying undirected graph is a tree consisting of two internally disjoint paths ending at their common terminal vertex. The remaining entries are independent uniform. For each search root, start an auxiliary undirected multigraph with the
\emph{entire} planted tree and the search root, initially an isolated
vertex if it lies outside the tree. The auxiliary graph is a forest.
Run the ordinary breadth-first search, adding each newly queried
unplanted arc to the auxiliary graph. A fresh target preserves cycle
rank, and a hit on a named vertex increases it by at most one. The
actual search graph is a subgraph of the auxiliary multigraph, including
any traversed planted arcs. Two excess arcs in the search therefore
require two named-target hits on newly queried unplanted cells. Given
the transcript each such cell still has a uniform destination.
With at most $N=2^{\ell+1}$ probes per root,
\[
 \Pr(G^c\mid\text{clean transcript})
 \le n\binom N2\left(\frac{N+2r+3}{n}\right)^2=O(n^{-1/5}).
\]
This holds uniformly over the planted word and its named vertices. Thus $\E[\ind_G n f_r(i)]\ge1-O(n^{-1/5}+r^2/n)$. Together with the two-replica bound this gives $\E[\ind_G(n f_r(i)-1)^2]=O(n^{-1/5}+(\log n)^{10}/n)$. Sum over pairs and apply Markov's inequality to the count exceeding the squared threshold.
\end{proof}

Set $b=(1-1/\log n)/n$ and $v=L n^{-1/10}$. For $t\ge2r$, Lemmas \ref{lem:upper} and \ref{lem:bad} give
\[
 (Q^{t-r}\ind_B)(i)\le v q_i(t-2r),\qquad
 (Q^{t-r} f_r)(i)\ge b\big(q_i(t-r)-v q_i(t-2r)\big).
\]
Since $q_i$ is nonincreasing,
$q_i(t)\le(1-b)q_i(t-1)+bv q_i(t-2r)$.
For $\lambda=1-b(1-2v)$ and all large $n$, $\lambda^{1-2r}\le2$. Induction using $q_i(t)\le1$ initially gives
\begin{equation}
 \max_i q_i(t)\le \lambda^{t-2r},\qquad t\ge2r. \label{eq:up}
\end{equation}
For a prescribed pair, one can also read the conditional hazard directly
from these bounds. On the event $S_1\ge\theta=1-\gamma$, the block
ratios above give $S_{j+1}/S_j\ge\theta$ for every $j\ge0$.
For $t\ge2r$, monotonicity within blocks therefore gives
$q_i(t-r)/q_i(t-1)\le\theta^{-2}$ and
$q_i(t-2r)/q_i(t-1)\le\theta^{-3}$.
Dividing the two first-meeting estimates by $q_i(t-1)>0$ yields
\[
 b(1-v\theta^{-3})
 \ \le\ \Pr_i(\tau=t\mid\tau>t-1,A)
 \ \le\ a\theta^{-2}.
\]
Both endpoints are $(1+o(1))/n$ uniformly in $t$. Since $q_i(2r-1)\ge\theta^2$,
multiplying the conditional survival probabilities gives
$\log q_i(t)=o(1)-(1+o(1))t/n$ for $t\ge2r$.
At $t=\lfloor cn\rfloor$ this yields $e^{-c}$ survival; at
$t=\lfloor\beta n\log n\rfloor$ it gives $n^{-\beta+o(1)}$.
Summing~\eqref{eq:low} and~\eqref{eq:up}, using monotonicity within blocks,
gives $(1-o(1))n\le\E_W[\tau\mid A,i]\le 2r+(1-\lambda)^{-1}=(1+o(1))n$.
This proves Theorem~\ref{thm:law}.

\begin{corollary}
Write $p(A)=\lim_{t\to\infty} \Pr_W(W_{1:t}\text{ is not a reset word}\mid A)^{1/t}$ for the synchronization parameter of \cite{Protasov}. Then $n(1-p(A))\to 1$ in probability.
\end{corollary}
\begin{proof}
For a fixed table, the probability within the limit lies between $\|Q^t\|_\infty=\max_i q_i(t)$ and $\binom n2\|Q^t\|_\infty$ by a union over starting pairs. The spectral radius formula gives $p(A)=\rho(Q)$. At fixed $n$ on the joint good event, take roots and let $t\to\infty$ in (\ref{eq:low}) for one prescribed pair and in (\ref{eq:up}) for all pairs. This gives $(1-u/(1-\gamma))^{1/r}\le p(A)\le\lambda$. Both endpoints are $1-(1+o(1))/n$ as $n\to\infty$.
\end{proof}

\section{Reconstruction from terminal moments}
\subsection{Polynomial learning from terminal labels}\label{sec:poly-learning-proof}

We first give the fixed-schedule construction at $m=n$. Later confidence refinements use the same observable arrays. All examples below are
independent uniform words of length $n$, accompanied by their terminal signs.
The start state $s$ is fixed before the target is sampled. The good event for
the labeled target will be independent of the accuracy and confidence
parameters $\epsilon,\delta$.

Let $T_a(x,f_a(x))=1$ be the letter matrices, acting on column tests by
$(T_ag)(x)=g(f_a(x))$, and put $P=(T_0+T_1)/2$. For a word
$v=v_1\cdots v_l$, write $T_v=T_{v_1}\cdots T_{v_l}$. In this appendix
$Q$ is the common-word kernel on the $N=n(n-1)$ \emph{ordered} distinct
pairs, killed on equality. Its ordered entries are bounded by the
corresponding unordered entries in Lemma~\ref{lem:upper}. Set
\[
 d=\lceil10\log_2 n\rceil,\qquad r=4d+1,\qquad
 w=2d+1,\qquad R=3w+d.
\]
The random digraph has, with high probability, a unique closed class
$\mathcal K$ and a stationary row law $\pi$ satisfying
\begin{equation}\label{eq:poly-stationary}
 \max_x\pi_x=n^{-1+o_{\Pr}(1)},\qquad
 \min_{x\in\mathcal K}\pi_x\ge n^{-2}.
\end{equation}
Here the minimum is over positive stationary masses. These facts follow
from \cite[Theorem~2]{ABP}; for binary outdegree, the exponent governing
the minimum positive mass is strictly less than two.

\subsubsection{Four sources and the centered ordered-pair block}

Use the reveal-on-first-query filtration of Section~\ref{sec:lag}. A
\emph{return} is a new cell whose destination is already named; reading
an exposed cell is free. Process distinct cells in a fixed order within
each step, naming their destinations immediately. Conditional on any
concrete history, unqueried cells are independent uniform states.

\begin{lemma}[Four fresh sources]\phantomsection\label{lem:poly-four}
Suppose four distinct vertices have no exposed outgoing cells and there
are $m$ named vertices, including these four. Paths 1 and 2 receive a
uniform word $W$ of length $l$, and paths 3 and 4 receive an independent
word $W'$ of length $l$. Success means that the final ordered endpoints
satisfy $1=3$ and $2=4$, with the two endpoints distinct. Conditional on
any concrete prior history, success with at most two further returns has
probability
\begin{equation}\label{eq:poly-four-exact}
 \sum_{k,h=0}^{l-1}2^{-\max(k,h)}
       \frac{(n-m)_{4l-k-h-2}}{n^{4l-k-h}}.
\end{equation}
For $m,l=O(\log n)$, the total success probability is at most
$6/n^2+O((m+l)^6/n^3)$. With four distinct starting vertices and no
previous queries, the contribution in \eqref{eq:poly-four-exact} is at least
\[
 \frac{6-(4l+6)2^{-l}}{n^2}\bigl(1-O(l^2/n)\bigr).
\]
\end{lemma}

\begin{proof}
Initially declare the four sources to be the frontier. At each subsequent
time the frontier comprises the occupied vertices newly named in the
preceding step. All outgoing cells of a frontier vertex are unqueried.
Expose frontier proposals first, and count a split if both letters are
requested at one frontier vertex. If $F$ is the frontier size, then
\begin{equation}\label{eq:poly-frontier}
 F_{\mathrm{next}}=F+\#\{\text{splits}\}
   -\#\{\text{returns from the frontier}\}
   +\#\{\text{new destinations from other sources}\}.
\end{equation}
Every newly named destination is occupied at the end of its step.
A successful match ends with $F_{\mathrm{end}}\le2$. Starting from four vertices and allowing at most two returns in \eqref{eq:poly-frontier} therefore forces $F_{\mathrm{end}}=2$,
exactly two frontier returns, no splits, and no new discoveries from
other sources.

A path that leaves the frontier cannot reenter it under these restrictions.
An exposed arc targets an old state, and a new cell from a source outside
the frontier would cause either an additional return or a new discovery
from such a source. Since all four paths finish on the frontier, none
ever leaves it. Each return consequently hits a destination generated
earlier in the same step. There are exactly two synchronous joins and
neither splits later. The prescribed final match forces these joins to
be $1=3$ and $2=4$.

Let $k,h$ be the numbers of letters strictly after the two joins. The last
$\max(k,h)$ letters of $W,W'$ must agree, an event of probability
$2^{-\max(k,h)}$. For compatible fixed words there are $4l-k-h$ distinct
cell queries, two with prescribed equalities, and all other destinations
must be distinct new states. This gives \eqref{eq:poly-four-exact}, also
when the joins occur in the same step. The chance of at least three
returns is at most
\[
 \binom{4l}{3}\left(\frac{m+4l}{n}\right)^3
\]
by sequential conditioning, padding the query list when necessary.
Finally,
$\sum_{k,h<l}2^{-\max(k,h)}
 =\sum_{j<l}(2j+1)2^{-j}=6-(4l+6)2^{-l}$.
The falling factorial differs from its leading power by relative
$O(l^2/n)$ when $m=4$, proving the lower bound.
\end{proof}

Let $G$ be the event used in Lemma~\ref{lem:tame}: every breadth-first
exploration to depth $\ell=\lfloor(\log_2 n)/5\rfloor$ has at most one
arc beyond its discovery tree. Then $\PrA(G^c)=O(n^{-1/5})$. A partial
table is safe if, from each named vertex, the probability of following
only its exposed cells for $d$ steps is at most $n^{-3}$. Lemma~\ref{lem:tame}
implies that on $G$ all the partial tables below are safe. Safety is
checked on the revealed history, preserving the product law of unused cells.

\begin{lemma}[Acquiring four sources]\phantomsection\label{lem:poly-four-acquire}
Start at a safe partial table, with $O(\log n)$ named vertices on every
continuation. In $w=2d+1$ steps, the probability that both replicas
survive but their four final vertices are not distinct unqueried sources
is $O((\log n)^2/n)$. The replicas may initially share vertices.
\end{lemma}

\begin{proof}
A return anywhere in the window costs $O(w\log n/n)$. On its complement,
a coordinate still inside the initially named set after $d$ steps has
followed exposed cells throughout; safety bounds this event by $n^{-3}$
for each of the four coordinates. After a path exits that set, new
vertices have unique labeled incoming edges and it cannot reenter the
set or revisit an outside vertex. If a final vertex coincides with
another path's current or earlier outside vertex, the unique incoming
chain traces both visits to the same exit arc. The path ending there
took at least $d$ further steps, so both visits have the same last $d$
letters. For different replicas these substrings are independent.
For paths within one replica, simultaneous exits along the same arc
would kill that pair; unequal exit times force equality of offset
substrings of the same word. At fixed offsets this also has probability
$2^{-d}$, including overlapping substrings, by the substring bijection
in Lemma~\ref{lem:acquire}. Sum over the six coordinate pairs and
$O(d^2)$ offsets. This union bound uses fixed offsets, without
conditioning word bits on the actual exit times. Every previously
queried source was occupied by a path, so the same argument covers a
final vertex that was queried earlier in the window. The total cost is
$O(w\log n/n)+4n^{-3}+O(d^2 2^{-d})$.
\end{proof}

Put
\[
 K=Q^R,\qquad a=N^{-1}\mathbf1^\top K,\qquad B=K-\mathbf1 a.
\]
All matrix norms below are Euclidean operator or Frobenius norms unless
specified otherwise.

\begin{lemma}[Centered ordered block]\phantomsection\label{lem:poly-centered}
With high probability,
\[
 \|B\|_{\mathrm{op}}\le\|B\|_{\mathrm{F}}\le n^{-1/20},
 \qquad \|a\|_2\le\frac{\log n}{n}.
\]
In fact $n\|a\|_2=O_{\Pr}(1)$.
\end{lemma}

\begin{proof}
For an ordered pair $i$, the squared row norm $\|K_i\|_2^2$ is the
conditional probability of an ordered terminal match between two
independent replica words starting at $i$ on the same table. Try up to
three windows of Lemma~\ref{lem:poly-four-acquire}, stopping at death or
the first acquisition of four fresh sources, and aborting unsafe
attempts. If $\eta=O((\log n)^2/n)$, three unsuccessful safe attempts
while both pairs survive cost at most $\eta^3$ by iterated conditioning.
The stopping rule uses only the exposed history. We first bound this
truncated experiment under the original law and then insert $\ind_G$,
since the truncation agrees with the experiment on $G$.
After an acquisition, the remaining length lies between $d$ and $R$.
Lemma~\ref{lem:poly-four} applies to the actual exposed history. On $G$
there is no safety abort, hence, uniformly in $i$,
\begin{equation}\label{eq:poly-row-moment}
 \E[\ind_G\|K_i\|_2^2]
 \le \frac6{n^2}+O\!\left(\frac{(\log n)^6}{n^3}\right).
\end{equation}
All conditional bounds here use observable histories, rather than
conditioning unused cells on $G$.

The quantity $\|a\|_2^2$ instead starts the replicas at two independent
uniform ordered distinct pairs. Their four sources are distinct with
probability $1-O(1/n)$. Restrict to the two-return transcripts counted
in \eqref{eq:poly-four-exact} with length $R$. Each transcript plants a
forest in the underlying undirected multigraph: the four paths join in
the prescribed pairs and share their continuations. Given its words
and answers, all unqueried cells remain independent uniform.

For each root, initialize an auxiliary exploration with this entire
forest, adding the root as an isolated vertex if necessary. Two excess
arcs in the root's breadth-first search imply cycle rank at least two
in the auxiliary multigraph. Starting from a forest, this needs two
fresh probes that hit named vertices. There are at most
$N_0=2^{\ell+1}$ new probes, each with conditional hit probability at most
$(N_0+4R+5)/n$. A union bound over roots gives the uniform relative estimate
\[
 \PrA(G^c\mid\text{two-return transcript})
 \le n\binom{N_0}2\left(\frac{N_0+4R+5}{n}\right)^2
 =O(n^{-1/5}).
\]
Consequently Lemma~\ref{lem:poly-four} yields
\begin{equation}\label{eq:poly-mean-moment}
 \E[\ind_G\|a\|_2^2]
 \ge\frac{6-O(n^{-1/5}+R^2/n+R2^{-R})}{n^2}.
\end{equation}
For each fixed table the variance identity is
\[
 \|B\|_{\mathrm{F}}^2=\sum_i\|K_i\|_2^2-N\|a\|_2^2.
\]
Both moment estimates use the same event $G$ and have leading term
$6/n^2$. Subtracting \eqref{eq:poly-mean-moment} from the sum of
\eqref{eq:poly-row-moment} gives
$\E[\ind_G\|B\|_{\mathrm{F}}^2]
 =O(n^{-1/5}+(\log n)^6/n)$.
Markov's inequality proves the first bound. Jensen's inequality and
\eqref{eq:poly-row-moment} give $n\|a\|_2=O_{\Pr}(1)$, and therefore
the stated high-probability bound on $a$.
\end{proof}

\subsubsection{Mixing, history rank, and an observable product}

The block $K$ commutes with exchange of the pair endpoints, and its
average row $a$ is symmetric. For $f:[n]\to[-1,1]$ let
$d_f(x,y)=f(x)-f(y)$. Killing at equality gives
\[
 (Q^l d_f)(x,y)=P^lf(x)-P^lf(y).
\]
On antisymmetric functions $K$ acts as $B$ and preserves antisymmetry.
Since $\|d_f\|_2\le2n$, Lemma~\ref{lem:poly-centered}, followed by
averaging $y$ against $\pi$, gives simultaneously for every $k\ge1$
\begin{equation}\label{eq:poly-mixing}
 \sup_x\|P^{kR}(x,\cdot)-\pi\|_{\mathrm{TV}}
 \le2n^{1-k/20}.
\end{equation}
This bound includes starts outside $\mathcal K$; additional steps preserve
it by total-variation contraction.

Write $\mu_v=\pi T_v$ and define
\[
 S_l=\E_{|v|=l}\mu_v^\top\mu_v.
\]
Let $u_l$ be the row of ordered off-diagonal entries of $S_l$ and
$H_l=u_l\mathbf1$. Both marginals of $S_l$ are $\pi$, and
$u_l=u_0Q^l$. For every fixed $C$, uniformly for $l\le CR$,
\begin{equation}\label{eq:poly-history-mass}
 H_l=1-o_{\Pr}(1).
\end{equation}
Indeed a prescribed distinct pair meets within $CR$ steps only after a
fresh return, of annealed probability $O((CR)^2/n)$. The sum of these
meeting probabilities over ordered pairs is at most $n(\log n)^4$
with high probability. Its contribution from the table-selected starting
law $\pi\otimes\pi$ is bounded by
$\|\pi\|_\infty^2 n(\log n)^4=o_{\Pr}(1)$ using
\eqref{eq:poly-stationary}. The initial diagonal has mass at most
$\|\pi\|_\infty$. This proves \eqref{eq:poly-history-mass} without
requiring independence between $\pi$ and the table.

The inequalities preceding \eqref{eq:up}, by linearity, apply to every
nonnegative initial sublaw on distinct pairs. Together with
\eqref{eq:poly-history-mass}, they imply
\begin{equation}\label{eq:poly-loss}
 H_l-H_{l+1}=(1+o_{\Pr}(1))/n
\end{equation}
at every fixed multiple $l=kR\ge2r$. These are the only first-meeting
estimates needed here; their proofs are in Sections~\ref{sec:lag}--\ref{sec:tails}.

Fix
\[
 t=4096R,\qquad h=t+1,\qquad x_t=H_{t+1}/H_t,\qquad q=1-x_t.
\]
To compare the shapes of $u_t$ and $u_{t+1}$, normalize the rows
$p_i=u_{iR}/H_{iR}$ and $p'_i=u_{1+iR}/H_{1+iR}$. Both update by
$\mathcal L(y)=yK/(yK\mathbf1)$. All denominators along these sequences
exceed $1/2$ by \eqref{eq:poly-history-mass}. Lemma~\ref{lem:poly-centered}
gives
\[
 \|\mathcal L(y)\|_2
 \le2\log n/n+2n^{-1/20}\|y\|_2.
\]
After 40 blocks both sequences have norm at most $5\log n/n$.
For probability rows $x,y$, put $z=\mathcal L(y)$. The exact difference is
\[
 \mathcal L(x)-\mathcal L(y)
 =\frac{(x-y)B(I-\mathbf1z)}{xK\mathbf1}.
\]
Since $\|\mathbf1z\|_{\mathrm{op}}=\sqrt N\|z\|_2$, subsequent updates
contract these differences by at most $n^{-1/25}$ for large $n$.
Passing to the absolute-entry norm costs at most $\sqrt N\le n$, so
\[
 \|u_{t+1}-x_tu_t\|_1
 \le2n\,n^{-(4096-40)/25}.
\]
Together with \eqref{eq:poly-loss} and the equal marginals of the $S_l$,
this proves
\begin{equation}\label{eq:poly-diagonal}
 q\asymp n^{-1},\qquad
 C=\frac{S_{t+1}-x_tS_t}{q},\qquad
 \|C-\operatorname{diag}(\pi)\|_{\mathrm{ent}}\le n^{-150}.
\end{equation}
Here $\|M\|_{\mathrm{ent}}=\sum_{x,y}|M(x,y)|$. To see the last assertion,
the off-diagonal absolute mass is at most
$\|u_{t+1}-x_tu_t\|_1/q$. Since $C\mathbf1=\pi^\top$, the diagonal
error is at most the same quantity. In particular, positive
semidefiniteness of $S_t$ and \eqref{eq:poly-stationary} imply, on core
coordinates,
\begin{equation}\label{eq:poly-history-rank}
 S_h\succeq c n^{-3}I
\end{equation}
for an absolute $c>0$.

For bounded tests $g_0,g_1$ and a word $v$, put
\[
 m_v(g)=\pi T_vg,\qquad
 J_l(v;g_0,g_1)=\E_{|w|=l}m_{wv}(g_0)m_{wv}(g_1).
\]
Equation~\eqref{eq:poly-diagonal} gives the uniform identity with error
\begin{equation}\label{eq:poly-product}
 \left|m_v(g_0g_1)-
 \frac{J_{t+1}(v;g_0,g_1)-x_tJ_t(v;g_0,g_1)}{1-x_t}\right|
 \le n^{-150}.
\end{equation}
Indeed $J_l=(T_vg_0)^\top S_l(T_vg_1)$ and
$T_v(g_0g_1)=(T_vg_0)(T_vg_1)$ entrywise.

The ratio $x_t$ is observable using the single random accepting mask.
Let $\xi$ be its sign vector and set
\[
 B_l=\E_{|w|=l}m_w(\xi)^2=\xi^\top S_l\xi.
\]
Conditional on the table, $\E_\xi B_t=\operatorname{tr}S_t=1-H_t$.
As $0\le B_t\le1$, \eqref{eq:poly-history-mass} and Markov's inequality
imply $1-B_t\ge1/2$ on a labeled-target event of probability tending to
one. Since $\xi_x^2=1$, \eqref{eq:poly-diagonal} then yields
\begin{equation}\label{eq:poly-calibration}
 \left|\frac{1-B_{t+1}}{1-B_t}-x_t\right|
 \le2q n^{-150}.
\end{equation}

\subsubsection{The two-branch feature Gram matrix}

Set $j=r$. Index a feature by a pair $\omega=(w_0,w_1)$ of length-$(j+1)$
words and put
\[
 Z_\omega(x)=\xi_{f_{0w_0}(x)}\xi_{f_{1w_1}(x)},\qquad
 k=2^{2(j+1)}.
\]
Two physical states belong to the same row class if their ordered
successor rows $(f_0(x),f_1(x))$ agree. On $\mathcal K$, choose one
representative per row class and let $V$ have entries
$V(x,\omega)=Z_\omega(x)/\sqrt k$. Features are constant on a row class.

\begin{lemma}[Two-branch Gram bound]\phantomsection\label{lem:poly-feature}
With probability tending to one over the table and the signs,
\[
 VV^\top\succeq(1/2-o_{\Pr}(1))I.
\]
\end{lemma}

\begin{proof}
For a uniform length-$l$ word $W$, write
\[
 D_l(x,y)=\Pr_W(f_W(x)=f_W(y)\mid A),\qquad
 G_l(x,y)=\E_W\xi_{f_W(x)}\xi_{f_W(y)}.
\]
Conditional on $A$, $G_l-D_l$ is a centered quadratic form in the
independent signs. For $x\ne y$, the symmetric coefficient matrix has
squared Frobenius norm at most $(\log n)^{12}/(2n^2)$ at $l=j$ by
Lemma~\ref{lem:upper}. The same bound holds at $j+1$, by applying the
$j$-step bound after the first step. Its operator norm is at most its
Frobenius norm. The Hanson--Wright inequality~\cite{RV} and a union bound
over roots give
\begin{equation}\label{eq:poly-sign-concentration}
 \max_{x,y}|G_l(x,y)-D_l(x,y)|
 \le C\log^7 n/n=:\eta_n,\qquad l=j,j+1,
\end{equation}
with high probability. At equal roots the difference is zero.

We next record the required table-only exposure bounds. For $x\ne y$ put
\[
 m_{xy}=\prod_{a=0}^1D_{j+1}(f_a(x),f_a(y)).
\]
Uniformly for distinct $x,y,z$ where applicable,
\begin{align}
 \E m_{xy}&=O(j^4/n^2),\label{eq:poly-two-return}\\
 \E[m_{xy}m_{yz}]&=O(j^8/n^4),\label{eq:poly-four-return}\\
 \E\left[\ind_{(f_0(x),f_1(x))\ne(f_0(y),f_1(y))}
       \prod_{v\in\{0,1\}^2}D_j(f_v(x),f_v(y))\right]
       &=O(j^6/n^3).\label{eq:poly-three-return}
\end{align}
All these estimates are annealed on the original table, before selecting
core representatives. Here is a return-count proof that also treats
the shared root in \eqref{eq:poly-four-return}.

Assign a separate contrast coordinate to each required meeting: put
$+1$ on its first path and $-1$ on its second, with all tail words
independent across coordinates. At a vertex, collect these signed
coefficients into a row vector. In a synchronous round, a fresh frontier
vertex has no queried outgoing cells. Its outgoing proposal vectors
sum to its current coefficient row. If their span has rank $p_l$ and
the current fresh rows have rank $r_l$, then $p_l\ge r_l$. A nonreturn
proposal goes to a new state. Its vector can be recovered from that
state's final coefficient by subtracting subsequent return proposals
to the same state. Previously exposed arcs cannot target a state first
created in this round. Thus, if $c_l$ fresh-cell returns occur in the
round, the next frontier coefficients satisfy
\begin{equation}\label{eq:poly-rank-return}
 r_{l+1}\ge p_l-c_l\ge r_l-c_l.
\end{equation}
This argument processes distinct cells sequentially and marks targets
immediately, so it permits several coordinates at the same source.
When all required meetings occur the coefficient row is zero everywhere.

For \eqref{eq:poly-two-return}, the initial fixed-letter proposals span
the two contrast coordinates, one for each letter. Extinction requires
at least two returns by \eqref{eq:poly-rank-return}. For
\eqref{eq:poly-four-return}, the coefficient rows at $x,y,z$ within each
letter block are $(1,0),(-1,1),(0,-1)$. They span rank two. The two
letter blocks occupy disjoint coordinates and give initial proposal
rank four, requiring at least four returns.

For \eqref{eq:poly-three-return}, first reveal the two fixed first letters
with one contrast per letter, and let $c$ be the number of returns at
this level. If $c\le1$, the remaining fresh rank is at least $2-c$.
Duplicate each coordinate for the fixed second letter. The second-level
proposals split into two disjoint coordinate blocks, doubling this rank.
Extinction needs at least $c+2(2-c)\ge3$ returns in total. If $c=2$ and
there is no later return, every coefficient at a newly named first-level
destination must already vanish. Each of the two nonreturn proposals
can then only be canceled by its opposite proposal in the same
first-letter coordinate. This forces $f_a(x)=f_a(y)$ for both $a$,
the excluded identical-row case. Cases $c\ge3$ already have the required
number of returns. Each calculation uses $O(j)$ queries and states.
Sequential conditioning bounds the chance of $q$ returns by
$O(j^{2q}/n^q)$, proving \eqref{eq:poly-two-return}--\eqref{eq:poly-three-return}.

Consider the symmetric matrix with diagonal one and off-diagonal
entries $m_{xy}$, restricted to distinct row classes. With high probability
the edges $m_{xy}>n^{-1/3}$ form a matching. Indeed, by
\eqref{eq:poly-four-return}, the probability of any two incident such
edges is at most $n^3 n^{2/3}O(j^8/n^4)=o(1)$. The remaining entries have
squared sum at most
\[
 n^{-1/3}\sum_{x\ne y}m_{xy}=o_{\Pr}(1)
\]
by \eqref{eq:poly-two-return}.

For $v\in\{0,1\}^2$, let $d_v=D_j(f_v(x),f_v(y))$. Summing
\eqref{eq:poly-three-return} over unequal-row pairs shows that their
sum of products $\prod_vd_v$ tends to zero in probability. Hence uniformly
over those pairs,
\[
 \min_vd_v\le\left(\sum_{\text{unequal rows }x,y}\prod_vd_v\right)^{1/4}
 =o_{\Pr}(1).
\]
Since
$D_{j+1}(f_a(x),f_a(y))
 =[D_j(f_{a0}(x),f_{a0}(y))+D_j(f_{a1}(x),f_{a1}(y))]/2$,
one branch is at most $1/2+o_{\Pr}(1)$ and the other at most one.
Thus every heavy edge has weight at most $1/2+o_{\Pr}(1)$. A matrix
supported on a matching decomposes into two-by-two blocks, so the
structural matrix has least eigenvalue at least $1/2-o_{\Pr}(1)$.

The actual Gram entry is
$\prod_aG_{j+1}(f_a(x),f_a(y))$. By
\eqref{eq:poly-sign-concentration}, its off-diagonal error is at most
$\eta_n(d_0+d_1)+\eta_n^2$, where now
$d_a=D_{j+1}(f_a(x),f_a(y))$. A meeting from prescribed distinct roots
needs a return, so
\[
 \sum_{x\ne y}D_{j+1}(f_a(x),f_a(y))^2=O_{\Pr}(nj^2).
\]
To obtain this bound, use $D^2\le D$ and the one-return probability
$O(j^2/n)$, including the initial fixed-letter query. The squared
Frobenius perturbation is therefore
$O_{\Pr}(\eta_n^2nj^2+n^2\eta_n^4)=o_{\Pr}(1)$.
Diagonal Gram entries are exactly one. This proves the lemma.
\end{proof}

\subsubsection{Estimating the observable arrays}

Let $c_*$ be the number of row classes on $\mathcal K$ and let $M$ be
their physical-state indicator matrix. Each row of $M$ has a single one;
its columns are disjoint and nonempty. Put
\[
 p=2^h,\qquad d_h=2^{-\lceil h/2\rceil},\qquad
 F(u,\cdot)=d_h\pi T_uM\quad (|u|=h).
\]
Since $1/2\le pd_h^2\le1$, \eqref{eq:poly-history-rank} gives
\[
 F^\top F=pd_h^2M^\top S_hM\succeq c'n^{-3}I.
\]
The row classes have well-defined deterministic letter transitions $U_a$.
The vector $b_a$ with entries $b_a([x])=\xi_{f_a(x)}$ is also well-defined.
States with the same successor row may have different current signs;
their transitions and these last-letter tests agree. Define
\begin{equation}\label{eq:poly-arrays}
 H=FV,\qquad H_a=FU_aV,\qquad g_a=Fb_a.
\end{equation}
Lemmas~\ref{lem:poly-feature} and \eqref{eq:poly-history-rank} imply
\begin{equation}\label{eq:poly-singular}
 s_*:=\sigma_{\min,+}(H)\ge n^{-2}
\end{equation}
for large $n$. All five arrays in \eqref{eq:poly-arrays} have norm at most
one: each matrix entry is bounded by $d_h/\sqrt k$, each vector entry
by $d_h$, and $pd_h^2\le1$.

For each feature $\omega=(w_0,w_1)$, apply \eqref{eq:poly-product} to
$g_0(x)=\xi_{f_{0w_0}(x)}$ and $g_1(x)=\xi_{f_{1w_1}(x)}$.
Use $v=u$ and $v=ua$ to recover entries of $H$ and $H_a$, respectively.
Both cases have normalization $d_h/\sqrt k$, including the length-$(h+1)$
word $ua$. Ordinary means $m_{ua}(\xi)$, multiplied by $d_h$, recover
$g_a$. Every factor on the right of \eqref{eq:poly-product} is an
ordinary terminal mean conditioned on a specified suffix.

Here is a finite-sample implementation. Let
\[
 L=2t+j+5,\qquad
 N_1=\left\lceil C2^L n^{242}\bigl(L+\log(8/\delta)\bigr)\right\rceil.
\]
For every suffix of length at most $L$, estimate its terminal mean by
the empirical sign average in its bin, using zero for an empty bin.
There are fewer than $2^{L+1}$ such bins, each with probability at least
$2^{-L}$. Conditional on the bin count, its signs are i.i.d. with the
appropriate conditional mean. Chernoff's and Hoeffding's inequalities
and a union bound give simultaneous error at most $n^{-120}$ with
probability at least $1-\delta/2$. Independence between bins of different
lengths is unnecessary.

If $n\ge L+2500R$, the unobserved padding before every suffix has at least
$2500R$ letters. Equation~\eqref{eq:poly-mixing} bounds the difference
between its conditional terminal mean and the stationary mean by
$4n^{-124}$. If $n<L+2500R$, use the full length-$n$ truth table instead:
as $L+2500R=O(\log n)$, its $2^n$ entries are polynomially many in this
regime. Coupon collection learns them with confidence $1-\delta$.

Compute $\widehat B_t,\widehat B_{t+1}$ from the empirical means. If
$1-\widehat B_t<1/4$, return a constant. Otherwise put
$\widehat x_t=(1-\widehat B_{t+1})/(1-\widehat B_t)$, and return a constant
if $1-\widehat x_t<n^{-2}$. On the good target and simultaneous sampling
events neither guard activates for large $n$. Equations
\eqref{eq:poly-calibration} and the sampling bound give
$|\widehat x_t-x_t|=O(n^{-120})$. Each empirical $J_l$ has product error
$O(n^{-120})$. In \eqref{eq:poly-product}, division by $q$ costs $O(n)$,
and perturbing the ratio costs at most $O(n^2)$ times its error. Clip
the resulting unnormalized product estimates to $[-1,1]$. Their errors
are $O(n^{-118})$.

After multiplication by $d_h/\sqrt k$, the Frobenius error of a whole
matrix is at most its largest unnormalized error, since $pd_h^2\le1$.
The same observation with factor $d_h$ controls the vectors. Thus the
estimated arrays satisfy simultaneously
\begin{equation}\label{eq:poly-array-error}
 \|\widehat H-H\|_{\mathrm{op}},\quad
 \|\widehat H_a-H_a\|_{\mathrm{op}},\quad
 \|\widehat g_a-g_a\|_2\le\tau:=n^{-100}.
\end{equation}
This normalization makes the fixed history length $h=t+1$ sufficient
for both conditioning and numerical precision.

\subsubsection{A stable linear representation in rational arithmetic}

The full column rank of $F$ and full row rank of $V$ imply
$VH^\dagger F=I_{c_*}$. Put $K_a=H_aH^\dagger$. For every nonempty word
$v$,
\[
 K_v=FU_vVH^\dagger.
\]
Because $\|F\|\le1$, $\|U_v\|\le\sqrt n$, $\|V\|\le\sqrt n$, and
$\|H^\dagger\|=1/s_*$, every true letter product, including the empty
product, has norm at most
\[
 B_*=n/s_*\le n^3.
\]
For a core row class $i$, its readout $e_i^\top VH^\dagger$ has norm at
most $1/s_*$, since the corresponding dictionary row has norm one.
For a suffix $z=z_1\cdots z_l$ of positive length this readout applied to
\begin{equation}\label{eq:poly-exact-feature}
 K_{z_1}\cdots K_{z_{l-1}}g_{z_l}
\end{equation}
returns its exact terminal sign from class $i$.

Use the rational ridge inverse
\[
 \lambda=n^{-26},\qquad
 R_\lambda(X)=X^\top(XX^\top+\lambda I)^{-1},\qquad
 \widehat K_a=\widehat H_a R_\lambda(\widehat H).
\]
Singular-value decomposition gives
$\|R_\lambda(H)-H^\dagger\|\le\lambda/s_*^3$.
The resolvent identity and \eqref{eq:poly-array-error} give
\[
 \|R_\lambda(\widehat H)-R_\lambda(H)\|
 \le\tau/\lambda+3\tau/\lambda^2,
 \qquad \|R_\lambda(\widehat H)\|\le(2\sqrt\lambda)^{-1}.
\]
Consequently $\|\widehat K_a-K_a\|\le a_*=O(n^{-20})$.
Expanding a product at the error positions, using the bound $B_*$ on
every intervening true product, bounds the error at length $l$ by
\[
 B_*\bigl[(1+B_*a_*)^l-1\bigr].
\]
The estimated end vector adds at most
$\tau B_*(1+B_*a_*)^l$. Multiplication by any true readout costs at most
$1/s_*$. The total error is therefore $o(1)$ uniformly for $l\le n$.
In particular, for large $n$ the estimated version of
\eqref{eq:poly-exact-feature} has signed margin at least $1/2$ under
the true readout, for every suffix and every core row class.

All normalizations are rational: $d_h$ is a power of two and
$\sqrt k=2^{j+1}$. Empirical means, matrix products, and the ridge inverse
are computable with polynomially many rational operations on arrays of
polynomial dimension. Rational inversion has polynomial bit complexity.
For word evaluation, take a common denominator for the two letter
matrices. A product of at most $n$ matrices then has bit length at most
$n$ times the single-letter bit budget plus $O(n\log p)$. The guards,
clipping, and positive ridge ensure that the procedure is defined on
every data set. These observations give polynomial bit complexity, as
well as polynomial arithmetic complexity.

\subsubsection{Fitting the starting readouts and proving the risk bound}

Choose the prefix length
\[
 b=R\left\lceil20\left(1+\log_n\frac{64}{\epsilon}\right)\right\rceil.
\]
Rounding $\epsilon$ down to a dyadic tolerance lets this ceiling be
computed by integer comparisons with powers of $n$. On the same table
event, \eqref{eq:poly-mixing} gives
\begin{equation}\label{eq:poly-bad-prefix}
 \Pr_{|u|=b}(f_u(s)\notin\mathcal K)\le\epsilon/8,
 \qquad 2^b\le(n/\epsilon)^{O(1)}.
\end{equation}
If $b\ge n$, learn the full truth table by coupon collection. Its sample
size $O(2^n(n+\log(1/\delta)))$ and running time are polynomial in
$n,1/\epsilon,\log(1/\delta)$ because $2^n\le2^b$ in this regime.

Suppose $b<n$. For a nonempty suffix $z$ define the rational feature
vector
\[
 \widehat\phi(z)=
 \widehat K_{z_1}\cdots\widehat K_{z_{|z|-1}}\widehat g_{z_{|z|}}.
\]
Take a second, independent sample batch and partition it by the first
$b$ letters. In each prefix bin, solve the rational linear inequalities
requiring signed training margins at least one for a linear readout of
$\widehat\phi$. Use a constant predictor if the bin is empty or infeasible.
On every prefix ending in $\mathcal K$, twice its true readout is feasible
for all examples by the margin bound above. Hence the returned predictor
has zero training error on the fixed union of these good bins.

Conditional on batch one, all possible readouts form a class of
halfspaces with separate blocks for the $2^b$ prefixes. Including an
affine coordinate covers the constant fallback and gives VC dimension
at most $d_*=(p+1)2^b$. Multiplying the error indicators by the fixed
indicator of the good bins preserves this VC bound. This restricted
error class has an element identically zero, and the returned predictor
is consistent on it. The realizable VC bound therefore shows that
\[
 N_2=O\!\left(
 \frac{d_*\log(2/\epsilon)+\log(8/\delta)}{\epsilon}
 \right)
\]
examples suffice to make the risk on good bins at most $\epsilon/2$,
with probability at least $1-\delta/2$. The other bins have total mass
at most $\epsilon/8$ by \eqref{eq:poly-bad-prefix}. The learner does not
need their identity or any hidden state.

Both batches and all computation have size polynomial in
$n,1/\epsilon,\log(1/\delta)$. Their two failure probabilities sum to at
most $\delta$. A truth-table procedure on a fixed finite initial range
of $n$ preserves these bounds. All target events used in the proof
depend only on the table and labels, and have probability $1-o(1)$;
so the same target event works for every $\epsilon,\delta$ at $m=n$ with fair accepting signs.
Appendix~\ref{sec:iid-clean-completion} gives the extension to every
accepting set and each fixed target-confidence exponent. In particular, setting
$\epsilon=n^{-\alpha}$ gives polynomial complexity for each fixed
$\alpha>0$, with joint target-and-sample failure at most $\delta+o(1)$.

\subsubsection{A polynomial exceptional-target bound for the original schedule}
\label{sec:quantified-target}

\begin{corollary}\phantomsection\label{cor:quantified-target}
There is an absolute constant $c>0$ such that, for each fixed integer $D\ge1$,
the original schedule in Appendix~\ref{sec:poly-lengths} fixes a two-batch algorithm $\mathcal A'_D$ before the target is drawn. It admits
good-target sets $\mathcal G^{\mathrm{qt}}_{n,D}$ with
\[
 \Pr_{A,b}\big((\mathcal G^{\mathrm{qt}}_{n,D})^c\big)=O_D(n^{-c}).
\]
Here the accepting bits are independent fair bits. For each prescribed
start and $m\le n^D$, $\mathcal A'_D$, given only the parameters and terminal-labeled words, has conditional risk at most
$\epsilon$ with training probability at least $1-\delta$, using samples
and bit operations polynomial in $n,1/\epsilon,\log(2/\delta)$ for fixed
$D$. The same target event covers all these lengths and accuracies, so the joint probability of failure over the target and sample is at most $O_D(n^{-c})+\delta$.
\end{corollary}

\begin{proof}
We retain all window lengths, precision parameters, and fitting rules
of the original schedule. The proof uses weaker stationary-mass bounds
and a different bound on the mean ordered-pair row, while preserving
the numerical threshold $\sigma_{\min,+}(H)\ge n^{-2}$ needed by that
algorithm. All estimates below are for sufficiently large $n$; a fixed
initial range is covered by the truth-table convention.

\paragraph{Centered block and mixing.}
Let $G$ be the breadth-first-search event used in
Lemma~\ref{lem:poly-centered}. Its displayed truncated moment estimates
give
\[
 \Pr(G^c)=O(n^{-1/5}),\qquad
 \E[\ind_G\|B\|_F^2]
       =O(n^{-1/5}+n^{-1}\log^6 n),\qquad
 \E[\ind_G\|a\|_2^2]=O(n^{-2}).
\]
The last assertion follows by Jensen's inequality from
\eqref{eq:poly-row-moment}. Thus the event
\begin{equation}\label{eq:qt-block-event}
 G,\qquad \|B\|\le n^{-1/16},\qquad
 \|a\|_2\le n^{-1+1/100}
\end{equation}
fails with probability at most
\[
 O(n^{-1/5}+n^{-3/40}+n^{-7/8}\log^6 n+n^{-1/50}).
\]
The action of $Q^R$ on antisymmetric tests is the action of $B$.
Consequently, for any stationary distribution $\pi$ of $P$,
\begin{equation}\label{eq:qt-mixing}
 \sup_x\|P^{kR}(x,\cdot)-\pi\|_{\mathrm{TV}}
       \le2n^{1-k/16}.
\end{equation}
This also proves uniqueness of $\pi$, convergence from every state,
and uniqueness of its closed irreducible support $\mathcal K$.
No separate probability estimate for a giant recurrent class is needed.

\paragraph{A stationary minimum with a polynomial tail.}
Write $\eta_2<1$ for the constant in the binary diameter theorem of
\cite{ABP}, and put $a_0=(1-\eta_2)/2>0$.
The proof of its Theorem~1, Section~4.1, pp.~10--12, gives, for some
$\gamma_0>0$,
\[
 \Pr\!\left(\operatorname{diam}(A)>
             (1+\eta_2+a_0)\log_2n\right)=O(n^{-\gamma_0}).
\]
Take $\varepsilon_0=\min\{a_0/2,1/20\}$. Lemma~4 of
\cite{ABP} gives per-vertex failure $O(n^{-1-\delta_{\rm ABP}})$
at this fixed slack. Proposition~5 and the union bound on p.~12
then allow $\gamma_0=\min\{1,\delta_{\rm ABP}\}>0$.
Choose $x$ with $\pi_x\ge1/n$. For every $y\in\mathcal K$ there is a
path from $x$ to $y$ of length at most this diameter. Stationarity along
the path, whose probability is at least $2^{-\operatorname{diam}(A)}$,
therefore gives, with $\zeta=(1-\eta_2)/2$,
\begin{equation}\label{eq:qt-pi-min}
 \pi_y\ge\tfrac12 n^{-3+\zeta}.
\end{equation}
This argument uses only the diameter tail and
\eqref{eq:qt-mixing}; it does not require a tail rate for the
stationary-extreme limit theorem.

\paragraph{Stationary-history survival by independent burn-in.}
Set $t=4096DR$ and $t_{\max}=t+1$. On each fixed table start two
independent uniform states, drive them by independent words of length
$128R$, and then drive them by a common word of length $t_{\max}$.
Let $C_A$ be their probability of having equal endpoints in this
experiment. Revealing the two paths on the random table uses
$O_D(\log n)$ cells and states. Equal initial states have probability
$1/n$; otherwise final equality requires a fresh-cell target to hit a
previous state. Sequential conditioning hence gives
\begin{equation}\label{eq:qt-burn-collision}
 \E_A C_A=O_D(\log^2n/n).
\end{equation}
On \eqref{eq:qt-block-event}, the law of either state after its
independent burn-in is within $2n^{-7}$ of $\pi$, by
\eqref{eq:qt-mixing}. Their joint law is within $4n^{-7}$ of
$\pi\otimes\pi$. Applying the common-word kernel and observing equality
preserves this total-variation bound. Thus
\[
 1-H_{t_{\max}}\le C_A+4n^{-7}.
\]
Markov's inequality applied to \eqref{eq:qt-burn-collision} proves
\begin{equation}\label{eq:qt-history-event}
 H_l\ge1-n^{-1/4}\quad(0\le l\le t_{\max})
\end{equation}
outside an additional event of probability
$O_D(n^{-3/4}\log^2n)$. Monotonicity of $H_l$ makes this one event
simultaneous in $l$. In particular this estimate accounts for the
dependence of $\pi$ on the transition table.

\paragraph{First-meeting loss and normalized contraction.}
The high-moment proof of Lemma~\ref{lem:upper}, with
$k=\lceil(\log n)^3\rceil$, bounds its two additional failures by
$n^2(b_n/a)^k$ and $n^4(n^2b'_n/L)^k$. Both decrease faster than
every inverse power of $n$; the already counted failure of $G$ is the
remaining term. The displayed squared-deviation estimate in the proof
of Lemma~\ref{lem:bad} bounds the failure of its cardinality assertion by
\[
 O(n^{-1/10}\log^2n+n^{-9/10}\log^{12}n).
\]
Its heat-kernel and first-meeting inequalities apply by linearity to
any nonnegative initial sublaw, including the table-dependent stationary
off-diagonal law. Together with \eqref{eq:qt-history-event}, at time
$t+1$ they give
\[
 q=1-H_{t+1}/H_t\asymp n^{-1}.
\]

All denominators in the two normalized-history sequences exceed
$1/2$. Their norm recursion from \eqref{eq:qt-block-event} is
\[
 \|\mathcal L(y)\|_2
   \le2n^{-1+1/100}+2n^{-1/16}\|y\|_2.
\]
After 40 blocks the norm is at most $5n^{-1+1/100}$.
The exact difference identity in the original proof consequently gives
the contraction factor
$C n^{-1/16+1/100}\le n^{-1/25}$ for all large $n$.
The original $4096D$ blocks therefore still imply
\begin{equation}\label{eq:qt-diagonal}
 \left\|\frac{S_{t+1}-x_tS_t}{q}
                  -\operatorname{diag}(\pi)\right\|_{\mathrm{ent}}
     \le n^{-150D}.
\end{equation}
Indeed the bound before division by $q$ is
$2n\,n^{-(4096D-40)/25}$, leaving the required slack.

For the fair accepting signs,
$\E_\xi B_t=\operatorname{tr}S_t=1-H_t\le n^{-1/4}$.
Hence $B_t\le1/2$ except with conditional probability at most
$2n^{-1/4}$. This verifies the same calibration guard and precision
bounds on a labeled-target event with a polynomial failure rate.

\paragraph{Feature conditioning.}
The displayed annealed moment estimates in the two-branch Gram proof
give the following rates at any fixed sufficiently small numerical
threshold. Two heavy edges incident to the same vertex have probability
$O(j^8n^{-1/3})$. The light entries have squared Frobenius norm at most
$n^{-1/3}\sum_{x\ne y}m_{xy}$, whose expectation is
$O(j^4n^{-1/3})$. The sum of grandchild collision products over unequal
rows has expectation $O(j^6/n)$. Markov's inequality makes the light
norm small and every heavy edge at most $1/2$ plus a small constant,
with these polynomial failure bounds. The matching-block argument
therefore gives a fixed positive lower bound for the structural Gram
matrix.

The pair heat-kernel event has the rate just recorded. On this event
the conditional Hanson--Wright and union bounds, with a sufficiently
large absolute constant, give
$\max|G_l-D_l|\le C\log^7n/n$ with failure $O(n^{-2})$.
Finally the one-return estimate gives
\[\E\sum_{x\ne y}D_{j+1}(f_a(x),f_a(y))^2=O(nj^2).\]
Requiring this sum to be at most $n^{3/2}$ costs $O(j^2n^{-1/2})$.
On that event the squared Frobenius sign perturbation is at most
\[
 O(n^{-1/2}\log^{14}n+n^{-2}\log^{28}n)=o(1).
\]
Choosing the fixed structural thresholds sufficiently small yields
$VV^\top\succeq\tfrac14 I$ outside the polynomially bounded union of
the preceding events. All bounds hold on physical pairs before any
row representatives are selected.

\paragraph{The original numerical schedule still applies.}
Positive semidefiniteness of $S_t$, \eqref{eq:qt-diagonal},
$q\asymp1/n$, and \eqref{eq:qt-pi-min} give, on core coordinates,
\[
 S_{t+1}=x_tS_t+q\operatorname{diag}(\pi)+E_t
       \succeq c_1n^{-4+\zeta}I,
 \qquad \|E_t\|\le q n^{-150D}.
\]
Aggregation of row classes and the rational normalization imply
$F^\top F\succeq c_2n^{-4+\zeta}I$. Consequently
\[
 \sigma_{\min,+}(FV)\ge c_3n^{-2+\zeta/2}\ge n^{-2}
\]
for sufficiently large $n$. This is precisely the numerical lower
bound used by the original ridge and word-product error estimates.
Its padding and prefix bounds also remain valid, because
\eqref{eq:qt-mixing} is stronger than the mixing bound they use.
Thus the original sample and computation guarantees follow unchanged.

Every target event in this proof is independent of
$\epsilon,\delta$. Taking their union, one may for example choose
\[
 c=\min\{\gamma_0/2,1/100\}>0.
\]
The constants may depend on the fixed length exponent $D$ through
\eqref{eq:qt-burn-collision}. This proves the asserted target bound and
the conditional-to-joint failure statement. The argument concerns the
original schedule; its stationary-mass replacement by
\eqref{eq:qt-pi-min} requires separate numerical bookkeeping for a
smaller schedule that uses the stronger stationary minimum.
\end{proof}

\subsection{A single transition-table condition for accurate learning}
\label{sec:one-block-learner}
On $N=n(n-1)$ ordered distinct pairs let $Q$ be the killed
common-letter transition matrix. The following sufficient condition
isolates the target-probability question at higher confidence.

\Needspace{14\baselineskip}
\begin{theorem}\phantomsection\label{thm:one-block-learner}
Fix $A,D,c_R,\kappa>0$, let $R=\lceil c_R\log n\rceil$ (more
generally use an explicitly specified comparable logarithmic integer),
$K=Q^R$, and $a=N^{-1}\mathbf1^\top K$. There are a polynomial-time
terminal-label learner and labeled-target events of probability
$1-o(n^{-A})$ such that, on these events, it succeeds on every table
satisfying
\[
                    \|K-\mathbf1 a\|_{2\to2}\le n^{-\kappa}.
\]
It has error at most $\epsilon$ at any prescribed length $m\le n^D$
with sample probability $1-\gamma$. Resources are polynomial in
$n,1/\epsilon,\log(1/\gamma)$ with fixed exponents. The learner
receives only $n$ and the labeled words. The transition condition can
also be replaced by $\|K-(K\mathbf1)a/(a\mathbf1)\|\le n^{-\kappa}$
after a fixed reduction in $\kappa$. Here $a\mathbf1>0$ on the events
used below.
\end{theorem}

\begin{proof}
The mask event below is required only on tables satisfying the block
condition and the stated geometric estimates. Other tables automatically
pass this implication. The extra events consist of the suffix dictionary theorem,
the simultaneous collision bound of Lemma~\ref{lem:cross-birthday},
the finite-distance bound on page~\pageref{sec:coarse-distance-proof} with exponent $A+2$, a uniform short-word
fiber bound, and accepting-mask calibration. They are all chosen
independently of the examples, the accuracy and the training length.

\emph{Uniform fibers.} For every fixed $T=O(\log n)$ and every
fixed $B_0$, apart from probability $O(n^{-B_0})$, every fiber
of every word of length at most $T$ has size $M=(\log n)^{O(1)}$.
One proof is the ghost-root bound used in the absorber learner. Sample
$k=\lceil(\log n)^2\rceil$ distinct uniform roots. For any fixed
word and target, initially mark all roots and the target,
then expose the $k$ complete paths successively.
For all endpoints to equal the target, the undirected graph of the
revealed paths must connect the $k$ roots to it. If the target is
distinct from the roots this needs $k$ fresh returns to named
components, among $O(kT)$ probes/states. If one of the roots is the
target, the initial coincidence costs $k/n$ and $k-1$ returns are
necessary. This connected-component argument counts even previously
queried arc reuse; one fresh return can join at most two components.
The factorial moment for the
fiber $Z$ is thus
$\mathbb E (Z)_k/(n)_k\le [C(k(T+1))^2/n]^k$
after increasing the absolute constant to cover the root coincidence.
Markov at a sufficiently large fixed power of $\log n$ bounds the
tail by $e^{-\Omega((\log n)^2)}$. Union over states and at most
$2^{T+1}$ words proves the estimate. It holds for every fixed
multiple of $\log n$ needed by the schedule.

Consequently every entry of $a$ is at most $M^2/N$, and
$\|a\|_2\le M/\sqrt N$. To compare the two table conditions, put
$g=\mathbf1-Q\mathbf1$. Same-letter indegrees are at most $M$, so
$\|g\|_2^2\le nM$ by summing the numbers of same-target pairs.
Every deterministic killed pair-word map has row sums at most one
and column sums at most $M^2$ through time $R$. By the Schur bound
$\|Q^l\|_{2\to2}\le M$ for $l\le R$. Hence
$\|\mathbf1-K\mathbf1\|_2
\le\sum_{l<R}\|Q^lg\|_2\le R\sqrt{nM^3}$.
It follows that $\rho=a\mathbf1=1-o(1)$, and the rank-one
difference between the two centered matrices is
$(K\mathbf1-\rho\mathbf1)a/\rho$, of norm
$O(R M^{5/2}/\sqrt n)$. For either condition take a fixed
$0<\kappa'<\min(\kappa,1/2)$ as the new ordinary centered exponent.

\emph{History calibration.} The action on antisymmetric
differences $f(x)-f(y)$ gives a unique ordinary stationary law and
worst-row mixing error $2n^{1-k\kappa'}$ after $kR$ steps.
After a logarithmic burn to total variation error $n^{-3}$, the
uniform starting law has atoms at most $M/n$ by the fiber bound.
Thus $\pi_{\max}\le M/n+n^{-3}$. For every short history $v$ the
stationary row $\mu_v=\pi T_v$ has largest atom at most $M\pi_{\max}$.
The history collision mass $\operatorname{tr}S_t$ is at most
$M\pi_{\max}=o(1)$ for every required logarithmic $t$, so all
relevant survival denominators are at least $3/4$ for large $n$.
For $B_t=\xi^\top S_t\xi$, the fair-mask expectation is
$\operatorname{tr}S_t$. Flipping the bit at $y$ changes this statistic
by at most $4\mathbb E_v\mu_v(y)=4\pi_y$. Bounded differences
therefore give $B_t\le1/2$ except with superpolynomially small
conditional mask probability. The synchronized collision lemma now
gives $q=1-H_{t+1}/H_t\ge c/[n(1+t+G)^4]\ge n^{-2}$ for any
fixed logarithmic history coefficient, independently of that coefficient
in the exponent. It uses ordinary single-walk mixing and fresh final
slots only. The diameter estimate gives $\pi_{\min,+}\ge n^{-S}$
for a constant $S$ depending on $A$.

\emph{Arbitrarily accurate diagonal extraction.} The normalized
history update in Appendix~\ref{sec:near-inverse} has the same
difference formula with $\|a\|_2\le M/\sqrt N$. After a fixed
number of blocks its rows have norm $O(M/n)$; later differences
contract by at most $n^{-\kappa'/2}$. At $t=kR$, multiplying the
off-diagonal discrepancy by $n$ to change norm and dividing by
$q\ge n^{-2}$ costs at most $n^3$. Matching the diagonal uses
the two stationary marginals. Therefore for any fixed $K_0$ one
can choose $k=O_{K_0,\kappa'}(1)$ to approximate
$\operatorname{diag}\pi$ to $n^{-K_0}$ in absolute-entry norm.
Also $S_{t+1}\succeq c n^{-(S+2)} I$ on recurrent states once
$K_0>S+4$, by positive semidefiniteness of $S_t$.
The exponent in this floor is independent of the chosen $k$.

Use the dictionary of Theorem~\ref{thm:suffix-all}, and let $M_0$
be the membership matrix of predictive classes within the recurrent
support. Since $M_0^\top M_0\succeq I$, the array $H=FV$ with
$F$ as in the original estimator has least positive singular value
at least $n^{-C}$ for a fixed $C=C(A,S)$ independent of $K_0,k$.
Readouts and true word-product norms are then polynomially bounded.
Choose a fixed $P_*>20(C+D+1)$. The ridge and product expansion in
Appendix~\ref{sec:near-inverse} only need normalized array precision
$n^{-4P_*}$ to obtain a uniform signed margin $1/2$ through length
$n^D$. To achieve that precision in the fixed tensor dictionary,
use the balanced product recursion of Appendix~\ref{sec:lifted-suffix-all}.
Its fixed depth is at most $d_*=\lceil\log_2 k_A\rceil$.
At each level clipping keeps factors bounded, division uses $q^{-1}
\le n^2$, and a perturbed ratio costs $O(n^4)$ times its error.
Taking the primitive mean precision and the diagonal tolerance at
most $n^{-K_0}$ for $K_0>4P_*+8(d_*+2)+S$ suffices by induction.
The empirical guard can use $\widehat q\ge n^{-3}$; the true lower
bound passes it with this precision. Use a long enough logarithmic
padding burn and suffix bins of length $O_{A,D,c_R,\kappa}(\log n)$.
For blocked shorter words, learn the truth table. Normalized dictionary
rows may be scaled by a common dyadic factor within a factor two of
unit norm, keeping all array operations rational.

The quotient by predictive equivalence has deterministic transitions
and terminal outputs. The original second independent sample panel
and per-prefix linear feasibility problems therefore apply with the
same margin proof and VC analysis. The logarithmic ordinary mixing
bound also supplies the requested prefix length with nonrecurrent
mass at most $\epsilon/8$. The conditional success and resource
bounds follow. All target predicates in this proof use fixed logarithmic
coefficients after $A,D,c_R,\kappa$ are chosen. None requires the
correct first-hit density $1/n$ at each starting pair.
\end{proof}

\subsection{Concentration of the centered pair kernel under row edits}
\label{sec:centered-edits}
Changing one row of the transition table can affect many entries of the killed pair kernel. We bound the change in the squared Frobenius norm of its centered block by summing over the starting pairs whose trajectories visit the edited row. This lets us apply concentration for independent table rows.

\begin{lemma}\phantomsection\label{lem:row-edit}
Let $Q_B$ be the killed ordered-pair kernel of a binary table $B$,
$N=n(n-1)$, $J=I-\mathbf1\mathbf1^\top/N$, and
$F(B)=\|JQ_B^R\|_F^2$. Consider any class of tables satisfying
\[
 \max_{i,j} Q_B^R(i,j)\le L/n^2,\qquad
 \max_{|v|<R,z}|f_v^{-1}(z)|\le M.
\]
On this class, for any two tables $B,C$ at row Hamming distance $h$,
\[
                 |F(B)-F(C)|\le 8 RMLh/n.
\]
\end{lemma}

\begin{proof}
Run each starting pair with the same word on the two tables,
including a cemetery outcome. They agree until a live coordinate is
about to read one of the $h$ edited rows. For each fixed time $j<R$
and input prefix, at most $hM$ first-coordinate starts can then lie
on an edited row in $B$; there are at most $n-1$ ways to choose the
second start. The same bound holds with coordinates exchanged.
Dropping the survival condition only enlarges the count.
Writing $K_B=Q_B^R$, the ordinary coupling inequality and summation
over all starts give
\[
   \sum_{i,t}|K_B(i,t)-K_C(i,t)|\le 4R(n-1)hM.
\]
The two matrices $JK_B,JK_C$ have all entries in
$[-L/n^2,L/n^2]$. Since $J$ is a self-adjoint projection,
\[
 F(B)-F(C)=\langle JK_B+JK_C,K_B-K_C\rangle_F .
\]
The bounds yield the claim, directly for the two endpoints, whether
or not any intermediate table belongs to the class.
\end{proof}

\begin{theorem}\phantomsection\label{thm:large-component-confidence}
For each fixed $A,D>0$ there exist constants $K_A$, $c_R$ and
$\kappa>0$ and a polynomial terminal-label learner such that, apart
from a labeled-target event of probability $o(n^{-A})$ under the
original i.i.d. model, every table whose terminal strongly connected
components all have size greater than $K_A$ is learned to prescribed
error $\epsilon$ at any prescribed $m\le n^D$, with sample failure
at most $\gamma$. The polynomial is in $n,1/\epsilon,\log(2/\gamma)$.
In particular, for these tables the block predicate of
Theorem~\ref{thm:one-block-learner} holds outside $o(n^{-A})$.
\end{theorem}

\begin{proof}
Choose $K_A$ as in Theorem~\ref{thm:suffix-all}, increasing it
for an extra unit of slack. On its outward-search event, in the
absence of a terminal component of size at most $K_A$, the maximum
ordinary heat entry at $j=\lfloor(\log_2 n)/8\rfloor$ is
$O_A(n^{-b_A})$ with $b_A>0$. Thus, from every start, the probability
of staying inside any given polylogarithmic named set for $d$ steps is
at most $n^{-3}$ for a sufficiently large fixed multiple $d$ of $j$,
by the Markov property at successive block ends. This conclusion
also applies to following only the entries of any partial table on
that named set. The original probability of search failure is
$o(n^{-A})$.

Here is the shared-table moment step explicitly. Set $d=qj$ with
$q$ a fixed integer larger than $\max(100,8/b_A)$, $r=4d+1$,
$k=\lceil(\log n)^3\rceil$ and $H=2kr+2k+10$. On the search event
and the no-small-terminal stratum, every partial history with at
most $H$ named states is safe: the walk following only its known
entries for $d$ steps stays in this set at all $q$ block ends,
with probability at most $(H C_A n^{-b_A})^q\le n^{-3}$.
Call this intersection $E$. For fixed ordered pairs $i,t$, put
$Z=Q^r(i,t)$. Run $k$ independent words starting from $i$, retaining
all exposed cells across words and naming $t$ before the first trial.
A trial is truncated only by the safety tests of Lemma~\ref{lem:upper}
at the start of each attempted acquisition. Given the entire past
transcript, future letters and unused cells have their original laws.
The two failed acquisitions on survival cost at most $\eta^2$, where
$\eta=4dH/n+2n^{-3}+d^2 2^{-d}$. After acquisition, split by zero,
one, or at least two returns strictly before the final letter. The
conditional probability of a truncated arrival is therefore at most
\[
 b_n=\eta^2+2/n^2+(2rH/n)(2/n)+\tbinom{2r}{2}(H/n)^2
       =O_A((\log n)^{10}/n^2).
\]
These bounds also apply if the first test aborts, when success has
probability zero. Conditional on the full table the ordinary arrivals
have joint probability $Z^k$. On $E$ no trial aborts; iterating the
conditional bound on partial transcripts under the original law gives
\[
 \mathbb E[\mathbf1_E Z^k]\le
 \Pr(\text{all $k$ truncated trials arrive})\le b_n^k.
\]
Thus $\Pr(E,\max_{i,t}Q^r(i,t)>2b_n)\le n^4 2^{-k}$.
Choose $R=\max(r,R_0)$, where $R_0$ is the four-walk lag of
Lemma~\ref{lem:poly-centered}. Left multiplication by the
substochastic $Q^{R-r}$ preserves the entry bound.
The uniform fiber bound of Theorem~\ref{thm:one-block-learner}
holds for this $R$ outside superpolynomially small probability.
Consequently the class $\mathcal G$ in Lemma~\ref{lem:row-edit},
with $L,M$ fixed powers of $\log n$, contains all the desired
tables except on an original event of probability $o(n^{-A})$.
It also has probability $1-o(1)$: a union over $1\le s\le K_A$
bounds the probability of any such closed set by
$\sum_s\binom ns(s/n)^{2s}=O_A(1/n)$.

For typical tables the earlier four-walk estimate
(Lemma~\ref{lem:poly-centered}) gives
$\|JQ^{R_0}\|_F^2\le n^{-u}$ with probability $1-o(1)$ for
the squared-norm exponent $u=1/10$. This is a coarse median estimate only.
Uniform fibers bound the largest column sum of $Q^{R-R_0}$ by
$M^2$ and its largest row sum by one (include length $R$ in the
fiber event). Hence $F(B)\le M^2n^{-u}$ on the intersection,
by submultiplicativity with the operator norm.
Let $d_H$ denote row Hamming distance and $c_n=8RML/n$. Use the Lipschitz extension of McShane~\cite{McShane} for
the restriction of $F$ by
\[
     \widetilde F(B)=\min_{C\in\mathcal G}
               [F(C)+c_n d_H(B,C)] .
\]
It equals $F$ on $\mathcal G$ and changes by at most $c_n$ under
any one row edit. All $n$ rows are independent under the original
law. McDiarmid's inequality~\cite{McDiarmid} implies that the mean of $\widetilde F$
is within $O(c_n\sqrt n)$ of a median and
\[
 \Pr\{\widetilde F> n^{-2\kappa}\}
       \le \exp[-n^{1-4\kappa}/(\log n)^{O(1)}]
\]
for any fixed $0<2\kappa<\min(u,1/2)$. Indeed its median is at
most $M^2 n^{-u}$ by the preceding typical bound. This argument
needs no independence conditional on $\mathcal G$, and the
extension is used only in the proof. In particular the squared norm $F$
is the bounded-difference observable. If $S$ is the no-small-terminal
stratum, then
$\Pr(S,\|JQ^R\|_{2\to2}>n^{-\kappa})
\le\Pr(S\setminus\mathcal G)+\Pr(\widetilde F>n^{-2\kappa})$
because $\|JQ^R\|_{2\to2}^2\le F$ on $\mathcal G$. Apply
Theorem~\ref{thm:one-block-learner}.
\end{proof}

\subsection{Common suffixes determine the geometry of input histories}
\label{sec:history-tree}
Write $\lambda$ for the uniform row on states in an i.i.d. $r$-letter
table, $r\ge2$ fixed. For words $u,v$, let $s(u,v)$ be the length
of their longest common suffix and let $u\preceq v$ mean that $u$
is a suffix of $v$, including equality. For any fixed $C>0$ and
$c<1/2$, with superpolynomial confidence over the table,
simultaneously for all $|u|,|v|\le C\log n$,
\begin{equation}\label{eq:history-uniform-gram}
 n\langle\lambda T_u,\lambda T_v\rangle=1+s(u,v)+O(n^{-c}).
\end{equation}
Conditioning every row to have distinct destinations preserves this
claim, since that event has probability bounded away from zero.
On the no-small-terminal stratum of the centered-block theorem, for
any fixed confidence exponent $A$, apart from $o(n^{-A})$,
\begin{equation}\label{eq:history-stationary-gram}
 n\langle\pi T_u,\pi T_v\rangle=
 1+s(u,v)+\frac{\mathbf1\{u\preceq v\text{ or }v\preceq u\}}{r-1}
 +O(n^{-c}).
\end{equation}
Here the small-terminal cutoff may depend on $A$; the first formula
uses no ergodicity assumption.

\begin{proof}
For fixed $u,v,a,b$ define $D$ to be the probability over independent
uniform starts $i,j$ that the states $f_u(i),f_v(j)$ are distinct
and their images under letters $a,b$ respectively are equal. Taking
$T=1+\max(|u|,|v|)$, under the table prior
\[
                   \mathbb E[nD]=1+O(T^4/n).
\]
Here are details for histories of unequal lengths as well. Initially
condition on distinct starts and mark both. Expose the two full
histories, before the last letters, in a fixed chronological order
for each path. Except with probability $O(T^4/n^2)$ there is at
most one fresh answer to an already named vertex. If there is none,
both ends are distinct and fresh. With one return and distinct ends,
at least one end is still unqueried as a source. If the return is a
self-intersection in one traversed component, the other trajectory
is a fresh simple path. If it joins the two initial components the
underlying undirected multigraph is a forest. Each path is then
simple. Both distinct terminal vertices could previously have been
sources only if each had occurred strictly inside the other path,
which would give a directed cycle. These alternatives also cover a
contact with the second root before its trajectory is run. On such
histories at most one of the last cells may already be exposed; the
two cells at the distinct endpoints are distinct. The conditional
probability of equality is exactly $1/n$. The no-return case alone
gives a lower bound $(1-O(T^2/n))/n$. Equal starts have probability
$1/n$; conditional on equal starts, with no fresh return all
revealed paths lie in an outward tree. Distinct endpoints can both
be old sources only if each is internal to the other root path,
which is impossible. Hence their additional conditional success
probability is at most $1/n+O(T^2/n)$, including the one-return
exception. This proves the expectation estimate under the original
independent-cell prior.

Put $M=(\log n)^{O(1)}$ on the uniform fiber event through time
$T$, whose complement has superpolynomial probability by the
growing-moment proof of Theorem~\ref{thm:one-block-learner}. For
two such tables differing at $h$ rows, an event contributing to
the difference in $D$ must visit an edited row in at least one
trajectory by its final probe, and must have equal final endpoints
in one of the two tables. When counting positive changes use that
table for the entire count, and likewise for negative changes.
For each word there are at most $ThM$ initial states that use one
of the edited rows. For a fixed such start, the fiber of its final
endpoint among the opposite starts has size at most $M$. Thus
$|nD-nD'|\le4ThM^2/n$. This is global on the good fiber set.
Extend the statistic there by the row-Hamming Lipschitz extension
of McShane~\cite{McShane} and clip to $[0,n]$. Its expectation differs from the unextended
original one by at most $n$ times the exceptional probability.
McDiarmid's inequality~\cite{McDiarmid} and a union over the polynomially many
words, letters and lengths give $nD=1+O(n^{-c'})$ simultaneously
for any fixed $c<c'<1/2$ at superpolynomial confidence.

If the two final letters agree, the inner product for the two
extended histories is the inner product for their predecessors
plus exactly $D$. If the final letters differ, it is $D$ plus
the probability that the pre-endpoints equal some $z$ with
$f_a(z)=f_b(z)$. For each pair of different letters the number
of such states is $\operatorname{Bin}(n,1/n)$ and is at most
$\log^2 n$ outside superpolynomially small probability. Each state
has mass at most $M^2/n^2$ for the two independent starts using
their specified histories. Finally any inner product with an empty
history equals $1/n$ exactly. Recursing for at most $C\log n$
steps proves \eqref{eq:history-uniform-gram}.

Use a burn of $G=O_{A,c}(\log n)$ ordinary random letters, long
enough for total variation $n^{-4}$ from $\lambda$ and
$r^{-G}\le n^{-2}$. Insert two independent such burn words before
$u,v$ and average \eqref{eq:history-uniform-gram} at the enlarged
fixed horizon. For suffix-incomparable histories the common suffix
is unchanged. In the comparable case the extra length has mean
$\sum_{i=1}^G r^{-i}=(1-r^{-G})/(r-1)$: conditional on the longer
augmented word, the relevant shorter burn is independent uniform.
The two burned endpoint rows may be replaced by $\pi$ at cost
$O(n^{-4})$ in the inner product. This proves the stationary formula.
\end{proof}

For example at every logarithmic binary history length $t$,
$n\operatorname{tr} S_t=t+2+O(n^{-c})$ on the stationary stratum.
More explicitly, fix a unit vector $d\in\mathbb R^r$ of coefficients
with $\sum_a d_a=0$. Define
$H_w(d)=\sqrt{n(r-1)/r}\sum_a d_a\,\pi T_{aw}$.
For a fixed orthonormal basis of that coefficient space, all the
vectors $H_w(d)$ indexed by words $w$ of length at most $C\log n$
and basis vectors have Gram entries
$\mathbf1_{\{w=w'\}}\langle d,d'\rangle+O_r(n^{-c})$.
For $w=w'$ the diagonal and unequal-letter entries before scaling
differ by $1+1/(r-1)$. For incomparable nodes all child cross
entries are equal; for different comparable nodes the cross
entry depends only on the shorter node's added letter, and the
other coefficient sum is zero. Thus every subfamily of at most
$n^{1/2-\eta}$ such vectors has Gram matrix $I+o(1)$ in operator
norm, simultaneously for every fixed $\eta>0$. This is a statement
about ordinary stationary history distributions. The terminal-label
learner additionally needs the hidden-state product construction.

\subsubsection{A constant relative gap for every accepting direction}
\label{sec:history-absolute-gap}
On the large-terminal stratum of Appendix~\ref{sec:history-tree},
fix a logarithmic horizon $t\le C\log n$. For each suffix $w$ let
$\mu_w=\pi T_w$. There is an absolute $c>0$ in the binary model
such that, except on a table event of original probability
$o(n^{-A})$ with the cutoff chosen for each fixed $A$, simultaneously
for every pair $u,v$ at this horizon,
\[
  \sum_{x:\pi_x>0}\frac{\mu_u(x)\mu_v(x)}{\pi_x}
       \ge c(1+s(u,v)).
\]
Consequently uniformly for every real test $h$ on the recurrent
support, even a test selected from the table,
\begin{align}
  \mathbb E_{|w|=t}\operatorname{Var}_{\mu_w} h
       &\ge c\operatorname{Var}_\pi h, \label{eq:constant-history-gap}\\
  \frac12\sum_{i\ne j}\nu(i,j)(h_i-h_j)^2
       &\ge c'\operatorname{Var}_\pi h. \label{eq:constant-survivor-gap}
\end{align}
In the second inequality $\nu$ is the quasi-stationary row of the
killed common-letter pair kernel, and $c'$ is also absolute. These
give relative bounds on rare accepting labels and dense linear
combinations in the same table. They do not lower-bound the
smallest singular value of an arbitrary one-sided sign dictionary.

\paragraph{A collision bound for three prescribed histories.}
For equal-length words $a_1,a_2,a_3$ of length $b=O(\log n)$ put
$p_j=\lambda T_{a_j}$, where $\lambda$ is uniform on states.
Write $s_{ij}$ for the common suffix lengths and $s_-,s_+$ for
their minimum and maximum. With superpolynomial confidence,
uniformly over these words and all such lengths,
\[
 n^2\sum_x p_1(x)p_2(x)p_3(x)
       \le C_0(1+s_+)(1+s_-)+o(1).
\]
Start from three independent uniform states equipped with distinct
artificial sources, so that each starting draw is itself a fresh
uniform probe. Expose the three trajectories by simultaneous rounds,
beginning with these three probes. With $O(b+1)$ probes and states,
three or more fresh returns cost $O((b+1)^6/n^3)$. Successful
coincidence connects the three initial components, hence uses at
least two returns. With exactly two the joined multigraph is an
undirected tree: both returns join distinct components, while every
other exposed edge discovers a new state. Every trajectory is simple
and ends at the same vertex. Intersections must then be synchronous
and permanent, since the remaining directed simple path in this tree
to the common endpoint is unique. Shared edges after a join use the
same letter, including for duplicate physical heads: two distinct
labeled slots with the same endpoints would create a parallel-edge
cycle and are excluded here.

Choose the first pair to merge and the two numbers of remaining
steps at their join and at the final join. A simultaneous triple
join is included. These numbers are at most $s_+$ and $s_-$,
respectively. On a successful two-return history, every current
source at each of these rounds is being used for the first time:
an appearance at a previous depth would contradict its unique
remaining distance to the final vertex. A coincident birth at
round zero uses the artificial fresh probes. To upper-bound the
probability for two specified joins, impose the history-measurable
condition that the required source slots are fresh and the pair
components still distinct at those rounds. At the join, read the
distinct fresh probes in a fixed order; the two prescribed
equalities have conditional cost $1/n$ each, including at one
round from three slots. The joint upper bound is $1/n^2$.
Union bounds prove the claim in expectation with error
$O((b+1)^6/n)$ after scaling.

On the simultaneous maximum-fiber bound $M=(\log n)^{O(1)}$,
the statistic on the left has row-edit constant
$6(b+1)M^3/n$. Indeed a selected path must read an edited row in
order for the triple coincidence to change. Per edited row each
word has at most $bM$ affected starts; for each, at most $M^2$
choices of other starts give the required final endpoint. Count
positive and negative changes on the corresponding endpoint tables.
Extend and clip the normalized statistic to $[0,n^2]$, choosing a
superpolynomially good fiber event before taking expectations.
Bounded differences and the polynomial union over words now give
a uniform $o(1)$ deviation. This establishes the asserted bound.

\paragraph{From triples to a weighted overlap.}
Prefix $u,v$ by two independent uniform burn words of length
$G=O_{A,C}(\log n)$, obtaining $a_1,a_2$ of equal length $b=t+G$.
Take $a_3$ independently uniform of length $b$. The three averaged
rows are within $n^{-5}$ in total variation of $\mu_u,\mu_v,\pi$,
respectively. Conditional on the first two words, the overlap
lengths $s_{13},s_{23}$ have geometric upper tails and bounded
second moments. Using $s_+\le s_{12}+s_{13}+s_{23}$ and
$s_-\le s_{13}$ gives
$\mathbb E[(1+s_+)(1+s_-)\mid a_1,a_2]\le C(1+s_{12})$.
The remaining mean is at most $C'(1+s(u,v))$, including equality
of $u,v$ when the burn adds a geometrically bounded length.
The cost of replacing three averaged rows after normalization is
$O(n^{-3})$. Thus
\[
 n^2\sum_x\mu_u(x)\mu_v(x)\pi_x\le C''(1+s(u,v)).
\]
Together with the two-history formula, Cauchy--Schwarz gives the
claimed lower bound via
$(\sum_x\mu_u\mu_v)^2\le(\sum_x\mu_u\mu_v/\pi)(\sum_x\mu_u\mu_v\pi)$.
All rows are supported on the positive recurrent masses.

Choose a suffix uniformly and draw its endpoint with law $\mu_w$.
The conditional expectation operator $B:L^2(\pi)\to L^2(W)$
maps $h$ to $\mu_wh$. The stochastic self-adjoint operator $BB^*$
has density $\sum_x\mu_u(x)\mu_v(x)/\pi_x$ relative to uniform
suffixes, hence minorizes $c$ times the stationary law at every
pair. Subtracting this constant kernel and dividing by $1-c$ gives
another stationary Markov contraction. On centered tests,
$\|B\|^2\le1-c$, proving \eqref{eq:constant-history-gap}.

For \eqref{eq:constant-survivor-gap} run two independent $\pi$
starts on the same suffix until the logarithmic PF burn.
Their surviving row $u_t$ differs from $(u_t\mathbf1)\nu$ by at
most $n^{-E}$ in total variation for any fixed specified $E$,
on increasing the burn coefficient. Also
$\tfrac12\sum u_t(i,j)(h_i-h_j)^2=\mathbb E_w
\operatorname{Var}_{\mu_w}h$. Normalize a nonconstant $h$ to have
oscillation one. Then $\operatorname{Var}_\pi h\ge\pi_{\min,+}/2
\ge n^{-C_A}/2$, by the diameter and mixing estimates.
Choosing $E>C_A+2$ absorbs the additive error uniformly in $h$.
Since $u_t\mathbf1\le1$, this proves the second claim, e.g. with
$c'=c/2$. Constants in the limiting collision coefficients do
not depend on the fixed logarithmic window multiplier.

\subsection{Identifying the recurrent machine and predicting every long input}
\label{sec:proper-exact}
A transient state may never appear as the endpoint of a training word of the prescribed length. Even so, on a typical target we can recover a discrete automaton that predicts every sufficiently long word correctly, including words of exponentially small probability.

\begin{lemma}[Recovering the recurrent Moore machine]
\phantomsection\label{lem:exact-recurrent}
On the clean centered-block event with a fixed confidence exponent,
the recurrent automaton minimized by predictive equivalence, including
its two named letter transitions and its accepting signs, is determined
up to a permutation of its states by a polynomial-time terminal-label
algorithm. The algorithm uses independent uniform words of one
prescribed length $m\ge C\log n$, with a fixed sufficient $C$, and
polynomially many samples in $n,\log(2/\gamma)$ for conditional success
$1-\gamma$. The individual starting state in this machine is not
part of this identification.
\end{lemma}

\begin{proof}
Write $\mathcal C$ for the predictive classes of the true recurrent
machine, $V$ for the lifted dictionary on these classes, and $\omega$
for their positive stationary weights. By
Appendices~\ref{sec:lifted-suffix-all} and \ref{sec:one-block-learner},
$V$ has unit rows and $\sigma_{\min}(V)\ge n^{-C_0}$, while
$\min\omega\ge n^{-C_1}$. Distinct classes have different sign
signatures on the words of length at most $L=O(\log n)$. Fix the
full list of words of length at most $L+1$, including the empty word.
Let $s_j$ be the signs after the $j$th such word, as a vector on
classes, and define
\[
  D=V^\top\operatorname{diag}(\omega)V,\qquad
  D_j=V^\top\operatorname{diag}(\omega s_j)V.
\]
The product $\omega s_j$ here is entrywise. Each entry is a known
scalar multiple of a stationary expectation of the product of at most
$2k_A+1$ short sign tests. The recursive terminal-only product
estimator therefore estimates every required entry to any fixed
inverse power of $n$ by choosing a sufficiently long logarithmic
lag. The dictionary dimension and list length $d$ are fixed
polynomials in $n$. Taking a sufficiently fine entry precision gives
simultaneous symmetric rational estimates in operator norm with
error $o(\eta/d)$, where $\eta$ is a positive dyadic number between
$n^{-(C_1+2C_0+3)}/2$ and $n^{-(C_1+2C_0+3)}$.
Known real normalization factors can be rounded to this precision,
or common positive rescalings can be applied to their columns.

For a proposed string of $j$ signs $a_1,\ldots,a_j$, consider
\[
  K(a)=\sum_{\ell\le j}(D-a_\ell D_\ell)/2-D/2 .
\]
If the sign string is absent, this is positive semidefinite, since
the corresponding integer mismatch count is at least one on every
class. If it occurs at class $x$, choose $u=V^\dagger e_x$ so
$Vu=e_x$ and $\|u\|\le n^{C_0}$. Its Rayleigh quotient for $K(a)$
is at most $-\omega_x/(2\|u\|^2)\le-n^{-(C_1+2C_0)}/2$.
Consequently the empirical matrix $\widehat K(a)+\eta I$ is positive
definite exactly for absent patterns. Positive definiteness can be
tested by rational elimination or leading principal determinants.
Extend realized patterns by both signs, retaining the children for
which this test fails. At any depth at most $n$ patterns pass.
Cap the search at $n$ for an arbitrary empirical input.
The final list thus contains exactly one extended sign signature per
recurrent class. Its empty-word bit is its accepting sign. For each
letter $a$, the values at tests $aw$, $|w|\le L$, determine the unique
signature to which this state moves because its successor is recurrent.
All steps have polynomial bit complexity. The stationary quantities
are obtained using unobserved random padding from the prescribed
start, as in the product estimator, even when it lies outside the core.
\end{proof}

\begin{theorem}[A proper automaton correct on every long word]
\phantomsection\label{thm:typical-exact}
In the original i.i.d. binary model with independent fair accepting bits,
fix the start in advance and put $g=\lceil3\log_2 n\rceil$.
From independent uniform training words of length $n$ with their
single final labels, a polynomial-time algorithm returns a DFA with
at most $n$ states. Except on a target event of probability
$O(1/n)$, with conditional sample probability at least
$1-\gamma$, its predictions agree with the target on \emph{every}
word of \emph{every} length greater than $g$. The sample size is
polynomial in $n,\log(2/\gamma)$ and is independent of a desired
prediction error.
\end{theorem}

\begin{proof}
Fix the clean events of the row-edit, suffix dictionary, spectral
and incoming-mass estimates with exponent $A=3$. Bounded terminal
components have original probability $O(1/n)$ by the closed-set
bound. There is a closed strongly connected core $\mathcal K$ of
size greater than $3n/4$ outside probability $O(1/n)$.
Events specifying a particular such set are disjoint and depend only
on its internal rows. Conditional on that set, exterior slots are
independent uniform states. If the fixed start is exterior, the
probability it can reach a completely exterior cycle without entering
the core is $O(1/n)$. Indeed enumerate a simple initial path of length
$a\ge0$ and a simple cycle of length $b\ge1$, meeting just at the
entry into the cycle. There are at most $2^{a+b}$ letter choices
and at most $(n/4)^{a+b-1}$ choices of new states, the last vertex
being prescribed. The distinct source slots cost $n^{-(a+b)}$.
Summing over $a,b$ gives the bound. Without such a cycle a purely
exterior path of length $g$ must be simple, with probability at most
$2^g(1/4)^g$. Hence every word of length $g$ takes the start into
the core on the remaining event.

Let $x\sim y$ mean equality of the two ordered physical outgoing
rows. Fix a sufficiently large $t=jR=O(\log n)$. We claim that except on an $O(1/n)$ table event,
\[
     \inf_{x\not\sim y}\delta_{xy}Q^t\mathbf1\ge1/32 .
\]
For each prescribed pair expose both starts along the full binary
tree of four letters. At its 16 leaves attach independent uniform
continuations of length $t-4$, shared within each leaf pair.
Their conditional all-meeting probability equals the product of the
16 leaf meeting probabilities. There are only constantly many cell
proposals in the branching prefix, so the probability of three fresh
returns there is $O(n^{-3})$. Otherwise write $c_j$ for the numbers
of returns in rounds $1$ through $4$, with sum at most two. Let
$r_j$ be the rank of the independent prefix contrasts restricted
to newly discovered end states. The proof of
\eqref{eq:poly-rank-return} gives $r_1\ge2-c_1$,
$r_{j+1}\ge2r_j-c_{j+1}$: fork each contrast into two coordinates
in disjoint letter blocks before the next round. If $c_1=2$ and
$r_1=0$, the two newly discovered vertices must each receive
opposite first-letter proposals in the same coordinate. This requires
$x\sim y$, excluded here. Thus for $c_1=2$ we still have $r_1\ge1$.
These inequalities give $r_4\ge4$. For $c_1=0$ or $1$, the lower bound is smallest when all remaining returns occur in round two. For $c_1=2$, every later fork is fresh.
Arbitrary continuations can extinguish this rank only with at least
four further fresh returns. Their joint probability is $O(t^8/n^4)$
by sequential exposure of $O(t)$ cells with $O(t)$ named targets.
This bounds the expected indicator of unequal rows times the product
by $O(n^{-3}+t^8/n^4)$. If all 16 leaf meeting probabilities exceed
$1/2$, their product exceeds $2^{-16}$. Markov's inequality and
a union over the $n(n-1)$ possible starts prove the claim: otherwise
one specified four-letter branch has survival at least $1/2$.

Let $\nu$ be the unit-mass left Perron row of the killed pair
chain, and $\rho$ its eigenvalue. The rank-one burn from
Appendix~\ref{sec:labeled-synchronization} gives
$\|\delta_{xy}Q^t-(\delta_{xy}Q^t\mathbf1)\nu\|_1\le n^{-4}$
uniformly in starting pairs at a sufficiently large fixed $j$.
Moreover $1-\rho=(1+o(1))/n$.
The pair heat cap at $R$ gives $\max_{i,k}\nu(i,k)\le
\rho^{-R}(\log n)^{O(1)}/n^2$. Hence
$\nu\{(i,k):\xi_i\ne\xi_k\}=1/2+o(1)$ with
superpolynomial conditional probability: the quadratic form on
the off-diagonal has zero mean, Frobenius norm at most
$(\log n)^{O(1)}/n$, and one can apply Hanson--Wright.
At $\ell=n-g$ the conditional probability of a different
final label from $x,y$ is consequently uniformly bounded below
by an absolute constant for $x\not\sim y$, by applying $Q^{\ell-t}$
to the burned row and using $\rho^{\ell-t}\to e^{-1}$.
For $x\sim y$ it is zero for every positive length.

First recover the minimized recurrent Moore machine by
Lemma~\ref{lem:exact-recurrent}. Call each of its
vectors of two successor states a \emph{row class}: two recovered
states share a row class when their ordered successor pairs agree.
These abstract classes coincide on the core with physical $\sim$.
Indeed equal physical rows have identical successor signatures;
and different physical rows disagree on a positive fraction of
length-$\ell$ inputs, so cannot have identical successors
even after minimization.
Take an independent panel of examples. For each prefix $p$ of
length $g$, compare each core candidate on the sample's remaining
suffix with the observed sign. The true endpoint is in the core.
All incorrect row classes have constant mismatch mass, and members
of the correct class always agree for positive remaining lengths.
Thus $O(2^g(g+\log n+\log(2/\gamma)))$ additional independent
words identify the correct row class for all prefixes, by binomial
concentration and a union bound.

It remains to reduce the prefixed core to at most $n$ states.
For physical row classes let
$U([x])=([f_0(x)],[f_1(x)])$. This map is well-defined and injective
on the whole physical table on the stated event: equality of two such
tuples forces the trajectories to coincide after every two-letter
word, whereas different rows have positive survival at time $t$.
Start a lookup by these ordered tuples for all recurrent row classes.
Keep the full recovered Moore machine, and choose one representative
state of each recurrent row class for entries from the prefix tree.
Each leaf $p$ at depth $g$ has its identified recurrent row-class
token. In descending depth intern each pair of child tokens: if it
is in the lookup, reuse that row-class token; otherwise create one
token and attach it to that pair. Each new token has one new DFA
state with arbitrary sign and edges to the representative states of
its two child tokens. By induction on the construction, token
equality is exactly equality of the represented physical row class.
Indeed the child tokens represent the actual child classes, and
$U$ is injective, including when those tokens were interned at
different depths. Thus there are at most $n-|\mathcal K|$ new
tokens, while the full minimized core has at most $|\mathcal K|$
states. Use the root representative as initial state and keep
all true core edges in its recovered Moore automaton.
Cap all counts and use a one-state default if a guard fails.

For any length-$g$ input prefix the simulated row-class token is its
true token: every transition, including those within the full core,
has the appropriate child token. Its simulated state is then
recurrent, and the true state lies in the same physical row class.
Every nonempty continuation gives the same output, since equal-row
states have identical physical successors. This proves the assertion
simultaneously over arbitrary future word lengths.
\end{proof}

\begin{proposition}[A target-prior obstruction to exact confidence]
\phantomsection\label{prop:exact-confidence-barrier}
For every learning rule based on $N$ independent uniform length-$m$
words and their terminal bits, the joint probability of making
an error on at least one length-$m$ word under the original prior
is at least
\[
 {1\over n}(1-1/n)^{2n-1}(1-2^{-m})^N .
\]
This holds also if the learner knows the entire transition table.
In particular for $m=n$ and polynomial $N$ the lower bound is
$(e^{-2}+o(1))/n$.
\end{proposition}
\begin{proof}
Exactly one of the start's slots is a self-loop, its other slot
leaves, and no other state has a slot pointing to the start, with
probability $2n^{-1}(1-1/n)^{2n-1}$.
The repeated self-loop letter is the unique length-$m$ input ending
at the start. If no sample used it, the fair accepting bit there
is independent of every observation and the learner's random bits.
The test on that word fails with conditional probability $1/2$.
\end{proof}

\subsubsection{The exact confidence bound uniformly in the accepting set}
\label{sec:mask-uniform-exact}
On a set of i.i.d. binary \emph{transition tables} of probability
$1-O(1/n)$, the proper exact conclusion at training length $n$
of Theorem~\ref{thm:typical-exact} holds for every accepting set,
including one chosen after the table is inspected. All sample and
running time polynomials may be fixed independently of the accepting
set; sample confidence is conditional on the particular target.
An inverse-polylogarithmic constant in the cut estimate suffices for this uniform bound.

For the table events take the large-component centered block,
logarithmic fibers, prefix and rank-four fork from
Theorem~\ref{thm:typical-exact}, the heat and common-suffix estimates,
and \eqref{eq:constant-survivor-gap}. The giant, the prescribed-start exterior-prefix event, and the rank-four
fork together cost $O(1/n)$; all required no-small-terminal failures can
be prescribed higher inverse powers. Write $\psi$ for the nonnegative
right Perron column normalized so the rank-one projection is
$\psi\nu$ with $\nu$ of unit mass. For all fixed accuracy exponents
$E$ and sufficiently long fixed logarithmic burns $l$,
\[
 \|Q^l-\rho^l\psi\nu\|_{2\to2}\le n^{-E}.
\]
Powers between block times are allowed by the fiber bound and
commutation; in the following $l$ may always be a whole block.
The normalization and the $1/32$ survival of unequal rows give
$\psi_{xy}\ge c$ when $x\not\sim y$, uniformly. For $z_i=2b_i-1$,
write $\alpha=\nu(1-z_i z_j)$. If the core has both colors then
\eqref{eq:constant-survivor-gap} implies $\alpha\ge c\pi_{\min,+}
\ge n^{-C_0}$, since the variance of a nonconstant sign is at
least $2\pi_{\min,+}$. Otherwise the core is one output class;
at length $n$ every input reaches it before its final letter.
The nonconstant case has both terminal signs of probability at least
$\pi_{\min,+}/2$ at training length $n$, by single-walk mixing.
A polynomial sample panel distinguishes the two cases, with a
constant fallback when only one sign is seen.

\paragraph{Label-independent structure for a centered lift.}
Choose $t_0$ a sufficiently large fixed logarithmic block multiple.
Let $D_{t_0}(x,y)$ be meeting probability and $G_j(x,y)$ the mean
product of labels after a common $j$-word. Extend $\psi_{xx}=0$.
The pair Perron estimates, evaluated on the bounded functions $1$
and $1-z_i z_j$, yield uniformly over the mask
\[
 D_{t_0}(x,y)=1-\rho^{t_0}\psi_{xy}+O(n^{-E}),\qquad
 G_j(x,y)=1-\alpha\rho^j\psi_{xy}+O(n^{-E}),
\]
for $j,t_0$ sufficiently large chosen logarithmic multiples. Reserve
one additional inverse power when applying an operator bound to a
bounded vector on ordered pairs. Take $j=t_0$ with a geometric
precision exponent $E>C_0+3$; it is fixed before the feature
dimension and the later moment-extraction precision. Write $P_*=I-\mathbf1\pi$.
Let $l_w(x)=[z_{f_w(x)}-\pi T_w z]/2$ and use the base features
$l_{w_0}(f_0 x)l_{w_1}(f_1 x)$, with independent length-$j$
words and the usual unit normalization by word count.
The single-branch Gram after division by $\alpha/4$ equals
$P_*D_jP_*^\top+O(n^{-E}/\alpha)$ entrywise, since both centered
constant matrices vanish. Its diagonal is $1+O(\sqrt{j/n})$
and each off-diagonal differs from $D_j(x,y)$ by
$O(\sqrt{j/n})$. Indeed $(D_j\pi^\top)_x\le
\sqrt{\mathbb E\|\pi T_w\|_2^2}=O(\sqrt{j/n})$ by the history Gram.
Take $E$ large enough to absorb the approximation into this bound.

By the table-only star bound in Appendix~\ref{sec:lifted-suffix-all},
applied at this fixed logarithmic $j$ with empty collapsed set,
each state has only a fixed number $B$ of heavy neighbors, where
the threshold is $n^{-a}$ on $\prod_iD_j(f_i x,f_i y)$ for a
fixed small $a>0$. Its proof uses only $O(j)$ probes at a fixed
star degree and permits this new multiple of $\log n$.
Therefore the twice-centered base Gram, divided by
$(\alpha/4)^2$, has diagonal $1+o(1)$ and absolute entries at
most $O(n^{-a})$ off these heavy edges. These estimates hold for
every nonconstant mask simultaneously on the structural table event.

For distinct physical rows the mismatch probability of their two
signs after length $j$ is
$\tfrac12\alpha\rho^j\psi_{xy}+O(n^{-E})>0$. This supplies a
short distinguishing word for every such pair. Two members of
the same physical row differ in their current sign or are
predictively equivalent. Consequently short tests through length
$j$ distinguish all predictive classes, uniformly over the mask
without a union bound. Apply the dual-vector construction of
Appendix~\ref{sec:lifted-suffix-all}: tensor a fixed power of the
centered two-branch features and $B$ short-test factors for the
heavy neighbors. Dividing the base features by $(\alpha/4)$ in the
analysis gives a diagonal $1+o(1)$ and light overlaps of size
$O(n^{-a})$. On taking fixed tensor powers, the $O(\sqrt{j/n})$
diagonal error remains $o(1)$, so the dual evaluation matrix has
diagonal bounded below and absolute row and column sums of
off-diagonal entries $o(1)$. The dual rows have polynomial norm
because their base norms are bounded, each selector reads a test
of length $O(\log n)$ and their count per row is at most $B$.
Returning to the unscaled features costs only a fixed power of
$\alpha^{-1}$. Thus the learner's bounded, polynomial-dimensional
dictionary has least singular value at least $n^{-C_2}$ on the
distinct recurrent predictive classes, uniformly over these masks.

\paragraph{Observable products and exact routing.}
Choose the product extraction lag $l$ only after these fixed
precision exponents have been set. With stationary history moments
$S_l$ and $B_l=z^\top S_l z$, the identity
$1-B_l=u_l(1-z_i z_j)$ and the burned-row approximation give
$1-B_l\ge c\alpha$ on every nonconstant mask. Indeed the row mass
is $1-O(l/n)$ by the stationary history Gram. With
$x_l=(u_{l+1}\mathbf1)/(u_l\mathbf1)$ we have
$1-x_l\asymp n^{-1}$ by the pair Perron rate and an arbitrarily
accurate logarithmic burn. Then estimation of the calibration ratio
$(1-B_{l+1})/(1-B_l)$ and the diagonal multiplication rule
loses only polynomial precision as in Appendix~\ref{sec:one-block-learner}.
Centered features additionally use the scalar $\pi T_w z$, itself
a terminal mean estimated to the same precision. Expanding their
fixed-degree products or applying the balanced recursion costs
only a fixed power. No sampling step requires observing two
responses from one training trajectory.
In using centered features at a shifted state $f_a(x)$ keep the
same scalar $\pi T_w z$ as in $l_w$. Each such scalar and each
primitive terminal mean is bounded by one. Changing these scalars
by $\zeta$ in a fixed-degree bounded product changes its value
by $O_A(\zeta)$, by telescoping the factors. Allowing for the
polynomial number of matrix coordinates gives the operator
precision in Lemma~\ref{lem:exact-recurrent}. That lemma's
factorization and its witness $Vu=e_x$ use a lower singular-value
bound; bounded but unequal feature row norms are permitted here.
The PSD pattern test therefore recovers the minimized recurrent
machine. The length-$(n-g)$ disagreement between different rows
is at least $c\alpha$ by the same Perron formula; a factor
$O(n^{C_0})$ in the second sample panel suffices. The exterior and
row-interning proof is independent of the mask, proving the claim.

\subsection{Finite exceptional sets of pairs}
\label{sec:exception-pair-matching}
On the i.i.d. binary table, a prescribed pair need not have a uniform
positive survival coefficient. Both letters at the two states might
give identical successors, or just one repeatedly chosen letter might
keep a pair distinct. We count disjoint copies of the latter type
of obstruction on the same table.

\begin{lemma}\phantomsection\label{lem:small-bad-pair-set}
Fix $A,C>0$ and an integer $3\le t\le C\log n$. Outside an original
table event of probability $o(n^{-A})$, at most $B_A$ unordered
pairs with \emph{different ordered outgoing rows} have probability
less than $1/8$ of still being distinct after $t$ common letters.
Here $B_A$ can be fixed independently of the logarithmic multiplier
$C$ (for each fixed $C$, the required initial $n$ may vary).
\end{lemma}
\begin{proof}
Fix a matching of $k$ pairs (all $2k$ starts distinct). Run from
each pair the four two-letter prefixes, and append a separate
uniform length-$(t-2)$ continuation to each resulting pair,
sharing its letters within that pair. Let $E$ mean that the
original $k$ row comparisons were all unequal and that all $4k$
leaf pairs met. We claim that realizing $E$ requires at least
$3k$ fresh-cell returns. Mark all $2k$ starts before exposing
the table. A return means the first exposure of a cell whose
head was named before this exposure; rereading a cell is free.

First expose both letters from all starts, with $c_1$ returns.
There are $4k-c_1$ new states. There are $2k$ contrasts, one for
each pair and each letter. Restrict their endpoint difference
vectors to the fresh states. Their rank is $r_1$. At most $k$
raw contrasts have equal endpoints since the two ordered rows
differ in each pair. Each such contrast can account for at most
one new state. Every fresh state on any raw unequal contrast lies
in the support of the restricted contrast. Select a basis of
restricted contrasts from the actual columns; its $r_1$ columns
each have support of size at most two, and together they cover
the support of every restricted contrast. It follows that
\[
                   4k-c_1\le k+2r_1 .
\]
Fork each contrast once more with both fixed letters into two
disjoint coordinate blocks. Outgoing proposal vectors at the
fresh frontier now span dimension at least $2r_1$. In each subsequent round use one coordinate per leaf contrast,
with opposite signs for its two endpoints. At a fresh state the two
outgoing proposal vectors sum to the current coefficient vector, even
if distinct leaves choose different letters. Let $r$ be the rank on
fresh states before the round, $r'$ the rank on newly discovered
states at its end, and $b$ the number of fresh returns in the round.
An old cell cannot point to a newly discovered state. Each proposal
that discovers a state equals its final coefficient vector minus the
later return contributions to that state. Thus all fresh-source
proposals lie in the span of the final new-state vectors and the $b$
return vectors. This gives $r\le r'+b$; in the fork round the same
argument starts with proposal rank at least $2r_1$.
At final equality all coefficients vanish, so at least $2r_1$
returns occur after the first round. This proves the claim also
when different leaf continuations subsequently choose different
letters at the same state.

There are $O_k(t)$ proposed cells and states. Sequential
conditioning on the original i.i.d. cells and union over return
times give $\Pr(E)=O_k(t^{6k}n^{-3k})$. Conditional on a table
where every root in this matching has survival less than $1/8$,
each of its four equally likely two-letter branches has conditional
survival less than $1/2$. If all rows differ, the probability of
$E$ on that table is at least $2^{-4k}$ by the independent
continuations (the table has been fixed). There are at most $n^{2k}$
candidate matchings. The probability of any size-$k$ bad matching
is $O_k(t^{6k}n^{-k})$.

Also this bad-edge graph has bounded maximum degree with overwhelming
fixed-power confidence. If the root survives with probability under
$1/8$, the meeting probability after each specified first letter
exceeds $3/4$. The product of these two meeting probabilities is
greater than $9/16$. For a star of degree $h$ on distinct neighbors,
fork one contrast along each first letter and use independent
continuations for all $2h$ contrasts. The $2h$ singleton proposal
coordinates of the neighbors have full initial rank; final extinction
requires at least $2h$ fresh returns. Markov and union over centers
and neighbors therefore bound the probability of any such bad star
by $O_h(t^{4h}n^{1-h})$. Choose $h>A+2$ and $k>A+1$. Endpoints
of a maximal matching cover all bad edges. Fewer than $2k$ endpoints
of degree less than $h$ cover fewer than $2kh$ such edges.
\end{proof}

\begin{corollary}\phantomsection\label{cor:perron-gap-dichotomy}
On the clean centered-block event at confidence exponent $A$, outside
$o(n^{-A})$, write $\psi_{xy}$ for the right survivor coefficient,
normalized by a unit-mass left Perron row $\nu$:
$Q^\ell=\rho^\ell\psi\nu+O_{2\to2}(n^{C_A}e^{-\theta_A\ell})$.
Uniformly over all different $x,y$, either $\psi_{xy}\ge c_A>0$,
or $\psi_{xy}=0$. In the latter case their common-word live
pair never reaches a pair with logarithmic survival at least $1/8$,
and for all $\ell\ge0$ its survival is at most $C_Ae^{-c_A\ell}$.
Any pair with unequal rows in this second case stays, until
coalescence or a final equal-row pair, in a set of at most $B_A$
unordered pairs. The relation $\psi_{xy}=0$ (with $\psi_{xx}=0$)
is a right congruence.
\end{corollary}

\begin{proof}
Choose a sufficiently long fixed logarithmic block multiple $t$.
A good pair, defined by $(Q^t\mathbf1)_{xy}\ge1/8$, has
$\psi_{xy}\ge1/9$ for large $n$ by the burned approximation and
$\rho^t\le1$. If such a pair can be reached, a shortest live path
to the good set visits at most $B_A$ bad unequal-row pairs; a
pair of equal rows goes to the cemetery at its next step. A
shortest path does not repeat an unordered pair. The eigenvector
identity and the mass of this path give
$\psi_{xy}\ge2^{-B_A}/9$. If the good set cannot be reached,
the accessible live submatrix has $t$-step row sums below $1/8$.
Since $\rho^t\to1$, restriction of the nonnegative eigenvector
to these states is zero, by its maximum norm and the eigenvector
identity. Every such pair has a path to the cemetery: otherwise
its survival at all times would be one. A shortest path has length
at most $B_A+2$ (allow a final equal-row pair). Markov blocks of
this length have absorption probability at least $2^{-(B_A+2)}$,
giving the displayed exponential decay. Finally survival on a
common word satisfies $q_{xz}(\ell)\le q_{xy}(\ell)+q_{yz}(\ell)$
by pathwise transitivity of endpoint equality. Divide by $\rho^\ell$
and let $\ell$ tend to infinity on the given finite table to obtain
the triangle inequality for $\psi$. The right eigenvector identity
makes its zero relation closed under each letter.
\end{proof}

\paragraph{Exactly distinguishing all recurrent predictive classes.}
For any accepting signs $z$ nonconstant on the clean recurrent
class, write $g_z(i,j)=1-z_i z_j$ and $p=\pi\{z=1\}$.
Then \eqref{eq:constant-survivor-gap} gives
$\alpha=\nu g_z\ge4c'p(1-p)\ge 2c'\pi_{\min,+}\ge n^{-C_A}$,
since both colors occur. The disagreement probability from pair $i$
at time $t'$ is $(Q^{t'}g_z)_i/2$. In the Perron approximation the
absolute entry error for $Q^{t'}g_z$ is at most
$2\sqrt{n(n-1)}\,n^{-E}$, since $\|g_z\|_2\le2\sqrt{n(n-1)}$.
Choose $E>C_A+2$ and its logarithmic burn $t'$ before choosing $z$.
The main term $\rho^{t'}\psi_i\alpha$ dominates this error for every
good pair, since $\rho^{t'}=1-o(1)$ and $\psi_i\ge1/9$. The same
table event therefore supplies witnesses even for accepting sets
chosen after inspecting the table. From any
pair which can reach the good set, prefix the short route to this
test. If it cannot, every live path stays in the bounded bad set,
possibly ending at an equal-row pair which merges at the next
step. A shortest route to any unequal current sign is then at
most $B_A+1$. Thus every predictively distinct pair of recurrent
states differs on a word of length at most $t'+B_A+1$. The
core-constant case has a single predictive class.

The centered two-branch Gram and bounded-degree selector proof
of Appendix~\ref{sec:mask-uniform-exact} now uses these short
witnesses at the same arbitrary confidence exponent $A$. The
PSD signature test reconstructs the \emph{exact} minimized
recurrent Moore machine at one training length $m\ge C_A\log n$
with time and samples polynomial in $n,m,\log(2/\gamma)$.
For this inference the target event is uniform over all masks and
all prescribed starts on the clean large-terminal stratum.
When the core is constant, a separate sample sees its dominant
sign; if the start is exterior its atypical signs can be removed
by a frequency threshold below the nonconstant core mass floor
and above the exterior burn error. The inverse-polynomial
geometric thresholds are fixed before the sampling precision.

\section{Passive regression and hidden closed components}
\subsection{An unknown stochastic output at each state}
\label{sec:stochastic-terminal}
Fix a table and numbers $h_i\in[-1,1]$, one for each state.
Given a word ending at $i$, the observed response $Y\in\{-1,1\}$
has mean $h_i$. The emissions in different examples are independent
conditional on their words. Two neighboring states can have
arbitrarily close means, or the same mean. The task is to estimate
the conditional law on each word.

\begin{theorem}\phantomsection\label{thm:stochastic-routing}
Fix a designated start in advance and put $g=\lceil3\log_2 n\rceil$.
Outside $O(1/n)$ original \emph{table} probability, for every
assignment of the means, a learner using length-$n$ uniform words
and their single noisy terminal outputs returns a deterministic
transition automaton with at most $n$ states and an estimated mean
at each state. With sample and internal success probability at
least $1-\gamma$ its conditional mean is within $\eta$ of the
target on \emph{every} word longer than $g$. Its time and sample
size are polynomial in $n,1/\eta,\log(2/\gamma)$. The table event
depends on the designated start and is common to all mean assignments.
The assertion is asymptotic in $n$.
\end{theorem}

\begin{proof}
Use the transition-only proper event from
Appendix~\ref{sec:mask-uniform-exact}. In particular every $g$-word
from the start ends in the recurrent class, each recurrent mass
exceeds $n^{-C_0}$, the centered block has polynomial contraction,
and unequal physical ordered rows have right Perron coefficient
at least an absolute constant. Throughout this proof a row token
is an ordered pair of \emph{physical} successors. Real means of
two states sharing a row may differ. Their responses agree after
every nonempty continuation.

\paragraph{Finding a scale for the emissions.}
Write $D=\max_{\mathcal K}h_i-\min_{\mathcal K}h_i$, $\pi$ for
the recurrent stationary row, and $S_l$ for the second moment of
$\mu_w=\pi T_w$ over all length-$l$ histories. Put
\[
 B_l=h^\top S_l h,\quad M_2=\pi h^2,\quad
 \beta=\tfrac12\sum\nolimits_{i\ne j}\nu(i,j)(h_i-h_j)^2 .
\]
Here powers of scalar-valued state tests are pointwise.
Equation~\eqref{eq:constant-survivor-gap} gives
$\beta\ge c\operatorname{Var}_\pi h\ge c n^{-C_0}D^2/2$.
The lower bound for the variance uses the maximum and minimum
atoms, each of mass at least $\pi_{\min,+}$. With $u_l$ the
off-diagonal history row and $w_0=(\pi\otimes\pi)_{\rm off}\psi$
bounded above and away from zero, the exact variance identity is
\[
 M_2-B_l=\tfrac12\sum_{i\ne j}u_l(i,j)(h_i-h_j)^2
         =w_0\beta\rho^l+O(D^2 n^{C_1}e^{-\theta l}).
\]
It uses only recurrent pairs. At any sufficiently long fixed
logarithmic burn, with arbitrarily prescribed fixed relative
precision in the remainder, the increment
$\Delta_l=B_{l+1}-B_l$ is
$w_0\beta\rho^l(1-\rho)+O(D^2 n^{-E})$. In particular for a
fixed $j_0=C_2\log n$, $\Delta_{j_0}\ge D^2 n^{-C_3}$.
Centering $h$ at $\pi h$ does not change this increment, since
the mean of each history row is $\pi$.

Let $c_n$ be the actual length-$n$ mean from the prescribed start.
Every history of length $\le j_0+1$ leaves at least $g$ unconstrained
input letters before it for sufficiently large $n$. Its padding
state is in $\mathcal K$ and its distribution differs from $\pi$
by $n^{-E'}$ in total variation for every needed fixed $E'$.
Thus each true conditional terminal mean differs from $\mu_w h$
by at most $n^{-E'}D$, and $|c_n-\pi h|\le n^{-E'}D$.
The stationary centered squared means at length $j_0+1$ average
at least $\Delta_{j_0}$. For a fixed $C_4$ we conclude that the
largest absolute conditional deviation $R$ from $c_n$ over
these histories satisfies $n^{-C_4}D\le R\le D$.
This is also true for $D=0$ with $R=0$.

Use an independent panel to estimate all these conditional means
and $c_n$ to $e=\eta n^{-C_4-5}$. Compute $\widehat R$ using the
estimated center $\widehat c$. If $\widehat R\le
\eta n^{-C_4}/4$, the true range is less than $\eta/2$ for large
$n$; returning the estimated constant $\widehat c$ suffices
simultaneously on all the indicated words. Otherwise take
$S=4n^{C_4}(\widehat R+3e)$, with the power rounded upward
once to an integer exponent. The inequalities above give
$S\ge D+e$, $S\ge c\eta$, and $S/D\le C n^{C_4}$.
The normalized core means $z_i=(h_i-\widehat c)/S$ therefore
lie in $[-1,1]$ and have range at least $c n^{-C_4}$.
All subsequent geometrical and reconstruction precisions for these
normalized means can be fixed inverse powers of $n$, independent
of $\eta$. Normalized sample responses may have magnitude
$O(1+1/\eta)$; this costs a polynomial factor in sample size,
and their conditional means on the relevant suffix bins remain
bounded. The empirical center and scale are fixed on fresh panels.

\paragraph{A dictionary on physical rows.}
For $z$ the same increment formula gives
$\Delta_{l+1}/\Delta_l=\rho+O(n^{-E''})$ at an appropriately chosen
product lag $l=O(\log n)$. The denominator is now of order at least
$n^{-C}(1-\rho)$, since the normalized range and positive stationary
floor are inverse polynomial. Impose an empirical guard below this
floor, as well as $1-\widehat\rho\ge1/(4n)$. Products of independently
estimated terminal means give the $B_l$; the observed $Y^2$ is
not a measurement of $z_i^2$. The recovered $\rho$ and
$(S_{l+1}-\rho S_l)/(1-\rho)$ supply the same diagonal-product
recursion as in Appendix~\ref{sec:mask-uniform-clean}.
For fixed geometric parameters below choose $l$ and the raw bin
precisions only after fixing the final polynomial accuracy of the
required products. All windows are then fixed multiples of $\log n$
and fit within training length $n$ with enough unused padding.

To see the needed rank, write $D_j(x,y)$ for meeting probability at
a fixed large logarithmic $j$ and $G_j(x,y)$ for expected product
of the two normalized means on a common $j$-word. Uniformly on
the core,
\[
 G_j(x,y)=\tfrac12[(P^jz^2)_x+(P^jz^2)_y]
                 -\rho^j\psi_{xy}\beta_z+O(n^{-E}),\qquad
 D_j(x,y)=1-\rho^j\psi_{xy}+O(n^{-E}).
\]
Here $\beta_z$ is the displayed Dirichlet quantity for $z$ and
is bounded below by a fixed inverse power. Double centering by
$I-\mathbf1\pi$ annihilates the two one-coordinate terms.
For the bounded features $(z_{f_w(x)}-\pi T_w z)/2$, the centered
single-branch Gram divided by $\beta_z/4$ thus agrees with the
centered $D_j$ up to the needed geometric error. The two-branch
product Gram has diagonal $1+o(1)$ after the corresponding
normalization and has absolute entries $O(n^{-a})$ off a graph of
bounded degree. This is the table-only star proof in
Appendix~\ref{sec:mask-uniform-exact}. Choose the geometric
precision just to dominate the $\beta_z$ floor and this threshold.

For any two distinct recurrent row tokens $x,y$, the same pair
calculation gives a word $w$ of length $j$ with
$|z_{f_w(x)}-z_{f_w(y)}|\ge n^{-C_5}$. Indeed the mean squared
difference is $2\rho^j\psi_{xy}\beta_z+O(n^{-E})$. Choose one
such test for each heavy neighbor $y$ of $x$ and multiply the
linear selector factors
\[
   { z_{f_w(v)}-z_{f_w(y)}
       \over z_{f_w(x)}-z_{f_w(y)}} .
\]
They take values $1$ at $x$ and $0$ at their respective neighbors;
their product is bounded in magnitude by $n^{O(1)}$ on the whole
core. A sufficiently large fixed tensor power of the two-branch
features makes all its light evaluations smaller than $n^{-2}$
even after this factor. Each selector is a linear functional
of the constant and length-$j$ tests, with polynomial coefficient
norm. The same dual-vector calculation gives $\sigma_{\min}(V)\ge
n^{-C_6}$ on the distinct recurrent row tokens. Here $V$ lists
the bounded, unscaled dictionary; row norms need not be identical.
Only tests of \emph{positive} length enter $V$, so it is constant
on each physical row irrespective of that row's current emission.
No minimum separation of the individual emission values was used.

\paragraph{Naming the rows with one random scalar.}
For each of the $d=2^j$ tests $z_w(x)=z_{f_w(x)}$ independently
draw $u_w$ uniformly from $\{1,\ldots,N\}$, $N=16n^2$. Put
$g_*(x)=\sum_w u_w z_w(x)/(Nd)$. Conditional on all but a
distinguishing coefficient, two rows' scalar difference changes
by at least $n^{-C_5}/(Nd)$ per unit change of that coefficient.
With probability at least $3/4$ all row tokens are therefore
separated by at least $\delta_*=n^{-C_5}/(4Nd)$ in $g_*$.
The scalar is bounded by one.

We describe a test that recognizes sufficient separation and
recovers the values needed to read the transitions. For any
observable bounded linear combination $f$ of our positive tests,
the terminal recursion estimates, to any specified fixed inverse
power, the matrices
\[
  D_V=V^\top\operatorname{diag}(\omega)V,\quad
  D_f=V^\top\operatorname{diag}(\omega f)V,\quad
  D_{f^2}=V^\top\operatorname{diag}(\omega f^2)V .
\]
Here $\omega$ is physical row mass. Squaring a fixed polynomial-size
linear combination yields polynomially many products of fixed
degree. The positive spectrum of $D_V$ is bounded below by
$\lambda_0=n^{-C_7}$ from the preceding dictionary floor and the
stationary mass floor. Count eigenvalues of its symmetric estimate
above $\lambda_0/2$ to get the exact number $r$ of row tokens;
rational inertia or polynomial-precision eigensolvers with a gap
suffice.

For grid location $a$ and radius $s_0$, scan with the regularized
positive-definiteness test on
\[
          D_{f^2}-2aD_f+(a^2-s_0^2)D_V.
\]
If all actual $f$ values have distance at least $s_0$ the matrix
is positive semidefinite. If one value is within $s_0/4$ of $a$,
the witness $V^\dagger e_x$ gives a negative Rayleigh quotient
at most $-s_0^2\lambda_0/2$ (decrease $\lambda_0$ if necessary
to the common product floor $\min\omega\,\sigma_{\min}(V)^2$).
A small positive diagonal shift and still smaller estimation errors
therefore ensure that every accepted grid point is within $s_0$
of a true value and every true value has an accepted point within
$s_0/4$ on mesh $s_0/8$. Scan an interval slightly larger than
the known bounded range to include endpoints.

Take $s_0\ll\delta_*$, and greedily retain accepted points of $g_*$
separated by at least $\delta_*/2$. If there are $r$ such points,
each represents a different physical row, all rows are represented,
and their true centers are separated by $\delta_*/2-2s_0$.
If the stronger sampled separation holds there will be $r$ points
for any such greedy scan. Draw $O(\log(2/\gamma))$ independent
coefficient sets until this verifiable test succeeds. All underlying
primitive moment estimates may be made simultaneously on one event
before the draws. For $a\in\{0,1\}$ compute the same scans for
\[
     f=g_*+t_*(g_*\circ f_a)\quad\hbox{and}\quad
     f=g_*+t_*(z\circ f_a),\qquad t_*=\delta_*/16 .
\]
Each new center lies within $t_*$ of the corresponding old one and
different centers remain separated. All accepting points can thus
be matched to their unique old center. Subtract the locations and
divide by $t_*$, incurring error at most $2s_0/t_*$. For the first
perturbation take the nearest recovered row token; this is the
exact successor token if $2s_0/t_*\ll\delta_*$. The second gives
the provisional mean on this letter edge. Choose $s_0$ small enough
also to make its normalized mean error a fixed inverse power well
below $\sqrt{\beta_z}$. All these scales are inverse polynomial
in $n$ alone. In particular returning only approximate real
locations in the scans does not cause exponential branching.

\paragraph{Routing and estimating the final outputs.}
Fit the entry row token separately for each length-$g$ prefix
on an independent panel using squared loss on the terminal bit.
On any fixed suffix the expected noisy squared loss equals its
conditional variance plus squared error to its true mean. The correct
token has uniformly small normalized mean error. Two different
tokens have mean squared discrepancy at least $c\beta_z$ on a
length-$(n-g)$ continuation by the pair formula. Thus the population
loss gap is at least $c'S^2\beta_z$ after unnormalizing the provisional
edge means and clipping them to $[-1,1]$ (uniform error relative
to the true means of the corresponding token cannot increase).
The gap is inverse polynomial in $n,1/\eta$, as $S\ge c\eta$.
Hoeffding on all prefix bins selects the exact row tokens. Intern
successor tuples back to the root as in Appendix~\ref{sec:proper-exact}.
This gives the true row-token trajectory on every input from time
$g$ onward. New transient tokens have no effect on the readout
after the next recurrent letter.

Finally simulate this recovered routing on a fresh panel through
the penultimate state and condition on its row token and the last
letter. Every recurrent token has probability at least half its
stationary mass at time $n-1$ by mixing; the last letter is fair.
Estimate each recurrent edge mean to $\eta/16$ by ordinary
averaging of terminal responses in these polynomial-mass bins.
This last step uses the actual word distribution and works at
arbitrarily fine requested precision, since it has no stationary
approximation bias. Every recurrent edge's true output belongs
to the physical head of that edge.

Convert the row presentation to an ordinary machine with means
on states. For each recurrent token group all its incoming
recurrent edges. Among their estimated means greedily retain a
maximal subset with mutual distances greater than $\eta/2$.
Each retained estimate must come from a different state identifier
in this row, since estimates of the same state differ by at most
$\eta/8$. Create one state per retained value. Direct each recurrent
edge to a nearest retained state of its true successor token and
give every state of a token its token's two outgoing edges. The
mean error on such an edge is at most $\eta/2+\eta/16$.
All recurrent tokens have some incoming recurrent edge by closure
and irreducibility. When the prefix routine first enters a recurrent
token it may choose any of its retained representatives, as its
next letter already has the correct edge output. Add at most one
state per new transient token. The resulting state count is at
most $n$, and every length-$>g$ word takes a correct recurrent edge
on its last letter. Each branch stops within its polynomial time bound and returns the default predictor if a cap is reached. The failure bounds above use independent batches.
\end{proof}

\subsection{Learning a real terminal mean on a clean table}
\label{sec:stochastic-clean-stratum}
This section extends the clean-table construction to squared loss for
an arbitrary function $h:[n]\to[-1,1]$. The observations are independent,
with conditional mean $h(f_W(s))$ and conditional second moment at most
one. The transformed responses in Section~\ref{sec:suffix-detail-regression} satisfy these moment conditions. All guarantees concern
the same length $m$ in training and testing.

Fix a target exception exponent $A$. Use the clean event at this
exponent for the single-walk mixing and stationary-mass bounds, the
centered ordered-pair block and its Perron projection, the relative gap
\eqref{eq:constant-survivor-gap}, the heavy-neighbor graph of
Section~\ref{sec:mask-uniform-exact} and the dichotomy of
Corollary~\ref{cor:perron-gap-dichotomy}. A zero right Perron
coefficient $\psi_{xy}$ means that $x,y$ merge on a uniform
$j$-word except with probability $C_Ae^{-c_Aj}$. A positive
coefficient is at least $c'_A$. The zero relation, with the diagonal
included, is an equivalence relation: use the union bound on
disagreement through a third state and the positive survival
coefficient in the other case. It is a right congruence by the
nonnegative eigenvector equation. A quotient class may contain two
different present means. We recover its transitions and fit a
separate readout at the observed length.

\paragraph{Choosing a measurable scale.}
On the recurrent class put $D=\max h-\min h$.
The derivation in Section~\ref{sec:stochastic-terminal} gives
$\beta_h=\nu((h_x-h_y)^2)/2\ge n^{-C_1}D^2$.
The stationary increment $B_{j+1}-B_j$ for a sufficiently large
fixed $j=C_2\log n$ is at least $D^2n^{-C_3}$. Here $B_j$ uses
the product of two independent estimates of each history mean.
The largest absolute centered mean $R$ through length $j+1$ thus
satisfies $n^{-C_4}D\le R\le D$. Use ordinary padding of length
$C\log(n/\epsilon)$ and estimate all means and the center with
absolute error $e=\epsilon n^{-C_4-10}$, including mixing bias.
If $\widehat R\le\epsilon n^{-C_4}/16$, the recurrent range is at
most $\epsilon/8$ for large $n$ and the constant center suffices.
Otherwise $S=8n^{C_4}(\widehat R+8e)$ satisfies
$S\ge\max(D+2e,c\epsilon)$ and $S/D\le Cn^{C_4}$.
For $z=(h-\widehat c)/S$ the recurrent tests have absolute value
at most one and $\beta_z\ge n^{-B_0}$ for a fixed $B_0$.
The ratios of successive observable increments estimate the pair
rate: their denominator is bounded below by a fixed inverse power
using $1-\rho\ge n^{-C}$ and the positive stationary projection
coefficient. The two-history multiplication rule now estimates any
required fixed-degree moment. Choose its lag after fixing its required
precision. The normalized response has conditional second moment
$O(\epsilon^{-2})$. Median-of-means on independent panels supplies
the needed additive precision and confidence. Extra unseen padding
$O(\log(n/\epsilon))$ controls transient contributions on this scale.

\paragraph{Conditioning on quotient representatives.}
Let $t_w(x)=z(f_w(x))$ and $l_w=(t_w-\pi t_w)/2$ for $|w|=j$,
where the constant in $j=C_j\log n$ will be chosen last.
Sample $d_1=n^{K_1}+O(1)$ independent tests and form the vector
$t(x)$ of their values and the constant, divided by $\sqrt{d_1+1}$.
Independently sample $d_0=n^{K_0}+O(1)$ word pairs for the vector
$v(x)$ of products $l_{w_0}(f_0x)l_{w_1}(f_1x)/\sqrt{d_0}$.
Round $d_1+1,d_0$ to powers of four of the same polynomial order.
For distinct classes,
$\mathbb E_w(t_w(x)-t_w(y))^2\ge c_A\beta_z$ at sufficient $j$,
by the pair projection. Boundedness gives a sampled gap
$n^{-B_2}$ for every representative pair with probability at least $.99$,
for fixed $K_1,B_2$ depending only on $B_0,A$.

For the centered Gram use
\[
 \mathbb E_w t_w(x)t_w(y)
 =\tfrac12(P^jz^2(x)+P^jz^2(y))-\beta_z\rho^j\psi_{xy}
       +O(n^{C}e^{-cj}).
\]
The two one-coordinate terms disappear under double centering. The
remaining centered two-branch estimate in
Section~\ref{sec:mask-uniform-exact} therefore applies after dividing
the product Gram by $(\beta_z/4)^2$. Its diagonal is $1+o(1)$ and its
entries are $O(n^{-a})$ off a graph of fixed degree $B$. The centering
correction uses $\max_w\|\pi T_w\|_\infty\le n^{-1+o(1)}$ at
logarithmic length. Entrywise sampling accuracy follows by choosing a
fixed $K_0$ using the floor on $\beta_z$. The exponents of both
sample counts are independent of $C_j$. One can fix the exceptional
graph at an initial $j_*=C_*\log n$ with a negligible projection
error. For any later fixed multiple $j$,
$\|\psi\|_\infty\le2$ and $1-\rho\le n^{-1+o(1)}$ make the
change in the single-branch survival kernel at most
$2|j-j_*|(1-\rho)+O(n^{-E})=n^{-1+o(1)}$. Each previously light
two-branch entry remains $O(n^{-a})$ for $a<1$. Thus choosing
$C_j$ after the tolerances invokes the same bounded-degree graph.

Use $V(x)=v(x)^{\otimes p}\otimes t(x)^{\otimes B}$ and take one
representative per class, giving a row matrix $\bar V$. At a heavy
neighbor $y$ of $x$ multiply the dual test for $x$ by
$(t_w-t_w(y))/(t_w(x)-t_w(y))$ using its sampled witness.
These factors have polynomial coefficient norms. A sufficiently
large fixed $p$ makes all light cross-evaluations at most $n^{-3}$
after normalizing the diagonal. The resulting evaluation matrix is
within $o(1)$ of the identity in operator norm. It gives
$\sigma_{\min}(\bar V)\ge n^{-C_V}$, with $C_V$ fixed before $C_j$.
On the other hand, within a class the expected squared distances in
the normalized $V$ and $t$ vectors are at most $C_{p,B}e^{-c_A j}$,
by coupling, right congruence and telescoping products of bounded
tests. For any subsequently required $\delta_0=n^{-K}$, Markov's
inequality and a union bound give distances at most $\delta_0$ on
all recurrent pairs of the same class with probability at least $.99$
by putting $C_j$ sufficiently large. This order of choices avoids
a dependence of the scalar separation on $2^j$.

\paragraph{Naming classes and their successors.}
Let $\omega$ be the class masses and
$\bar D=\bar V^\top\operatorname{diag}(\omega)\bar V$.
Its positive eigenvalues exceed a known floor
$\lambda_0=n^{-C_\lambda}$ by the stationary floor and the preceding
singular value bound. The true moment $D_V=\mathbb E_\pi V^\top V$
is within $2\delta_0$ of $\bar D$. Form a bounded scalar
$g=\sum_i u_i t_{w_i}/(N d_1)$ with the $u_i$ independently
uniform on $[N]$, $N=64n^2$. Its coefficient norm in $t$ is at
most two. A sampled witness separates a pair of representatives in
steps of at least $n^{-B_2}/(N d_1)$ as its coefficient varies.
Conditioning on the other coefficients and taking a union bound
gives a simultaneous gap $\Delta=n^{-C_\Delta}$ with probability
at least $.98$. Within a class $g$ is constant to $2\delta_0$.

For $f=g$ and $f=g+u(g\circ f_a)$, $u=\Delta/16$, compute moments
$D_f=\mathbb E_\pi f V^\top V$ and $D_{f^2}$. Their estimates are
within $O(\delta_0+\eta)$ of the ideal class matrices, with coded
values $g_C+u g_{f_a C}$ in the second case. Count the eigenvalues
above $\lambda_0/2$ to find the class count. On a rational grid
consider
\[
 D_{f^2}-2bD_f+(b^2-s_0^2)D_V.
\]
If every class value lies at least $s_0$ from $b$, its ideal matrix
is positive semidefinite. A class within $s_0/4$ yields a negative
Rayleigh quotient at most $-15\lambda_0 s_0^2/16$ by its dual
coordinate. Add the shift $\lambda_0s_0^2/10$ and keep grid points
with a negative direction. With mesh $s_0/8$ each true value has
a kept point within $s_0/4$, and each kept point is within $s_0$
of a true value. The ideal coded values remain separated by
$\Delta-2u$. Clustering the kept points at distance $2s_0$ recovers
one center per class. Match its coded center to the original center,
subtract and divide by $u$, then read the nearest successor center.
Choose $s_0\ll\Delta^2$, then $\delta_0,\eta\ll\lambda_0s_0^2$.
The $\delta_0$ tolerance fixes $C_j$; $\eta$ fixes the multiplication
lag. These choices give exact routing with trial success at least
one half. Use fresh independent trials $O(\log(2/\gamma))$ times
at the same fixed dimensions, cap all scans at their polynomial
limits and validate the resulting predictors.

\paragraph{The readout.}
Burn $g_0=C_A\log(2nm/\epsilon)$ leading letters. The mass of bins
ending outside the recurrent class is $O(\epsilon)$ with an
arbitrarily small fixed multiplier. In each bin try every initial
class of each fitted machine. Track that class until a last window
of $L=C'_A\log(2nm/\epsilon)$ letters. Fit a bounded regression
table on its class at the window boundary and the last $L$ letters.
For a correct routing trial and initial class, the comparator
$h(f_w(x_C))$ for a fixed physical representative of the boundary
class has squared risk at most $4C_A e^{-c_A L}$, uniformly over
the physical mixture there. Each cell table has at most $n2^L$
entries. An inverse-polynomial grid of possible values has log
covering number $O(n2^L\log(2/\epsilon))$. Fitting can use
cellwise truncated empirical means; truncate the raw response at
$O(1/\epsilon)$ so the uniform conditional bias is $O(\epsilon)$
by the second moment assumption. Median-of-means comparisons
on held-out samples in each initial bin select a candidate: the
centered score below has bounded conditional second moment for
every bounded predictor. The
expected loss $\widehat h^2-2Y\widehat h$ differs from the
mean-squared risk by a common term. Adjust the constants in the
burn, fitting and validation errors to reach risk $\epsilon$.
If $m$ is too short for any required window and its padding,
$2^m$ is polynomial in $n,m,1/\epsilon$ and direct full-word
regression suffices. The number of histories and prefix bins,
the bit precisions and all trial sizes obey the stated polynomial.

\subsection{Regression with a closed component of constant mean}
\label{sec:stochastic-singleton}
A stochastic terminal observation is a bounded random variable $Y\in[-1,1]$
with $\mathbb E[Y\mid W]=h(f_W(s))$, where $h:[n]\to[-1,1]$.
Given independent training examples, we estimate the conditional mean under squared loss, allowing the variance to depend on the whole word. A trap with one physical state has
constant mean automatically; the same argument covers any bounded closed set
on which the means agree.

\paragraph{Spectral inputs and notation.}
Use the clean repair of a closed component in the finite-basin reduction.
Collapse the closed set to one distinguished state $x$. Repair its outgoing
arcs as in that reduction. The original exterior chain is recovered by
killing the repaired process at $x$. For $b=1,2$ let $P_b$ be the clean
transition kernel on distinct $b$-tuples, killed on coincidences between
coordinates. Write $P_1=P$ and $P_2=Q$. The clean heat and Perron estimates
at $D=C\log n$ give $P_b^D(i,j)\le M/n^b$, a rank-one approximation
$P_b^D=\rho_b^Dh_b\nu_b+E_b$ with $\|E_b\|_2\le n^{-B}$ for any prescribed
fixed $B$ by increasing $C$, and $\nu_b\mathbf1=\nu_bh_b=1$,
$\|h_b\|_\infty=O(1)$, $\|\nu_b\|_\infty\le M/n^b$. Here and below $M$
can be a fixed polylogarithm or $n^{o(1)}$. The uniform upper bounds on
short-word fibers apply to every substring of a word of length $O(D)$.
The clean pair first-meeting probability at step $D+1$ from uniform
distinct states is at least $n^{-1-o(1)}$ on the event used below. All these
statements concern the clean repair, including its choice; independence of
the marker from that repair is unnecessary.

\paragraph{Removing a state from the clean walk.}
Write $U_b$ for the same kernels killed when any coordinate visits $x$,
padded by zero on tuples containing $x$, and $A_b=U_b^D$. Then
\[
 \|U_b^{3D}-A_b P_b^D A_b\|_2\le n^{-1+o(1)}.
\]
Indeed in the two outer blocks every fixed endpoint has probability at most
$M/n^b$. After the first burn the killed trajectory has mass at most
$O(M/n)$ on the forbidden slice at any subsequent middle-block time
(use domination by $P_b^j$ for $j\ge D$ and its row-substochasticity).
After hitting the slice the final endpoint has probability at most $M/n^b$.
A union over $D+1$ times bounds every entry of the difference by
$n^{-b-1+o(1)}$. The operator norm is at most the number of rows times
the entry bound. The leading rank-one operator has nonzero eigenvalue
$\rho_b^D\nu_b U_b^{2D}h_b=1-n^{-1+o(1)}$, since $\nu_b$ is a left
eigenmeasure of $P_b$ and the mass lost by touching the marker in $2D$
steps under $\nu_b$ is at most $n^{-1+o(1)}$. In this estimate the clean
pair killing rate is $n^{-1+o(1)}$. The rank-one operator has norm $n^{o(1)}$
by the heat bounds on its outer factors. Its error, including the middle
clean approximation, is $n^{-1+o(1)}$.

We use the resolvent directly. For a rank-one
$J$ with eigenvalue $\lambda$ near one, $(z-J)^{-1}=z^{-1}I+
J/[z(z-\lambda)]$. The Neumann bound on the contours $|z|=n^{-1/4}$ and
$|z-\lambda|=n^{-1/4}$, followed by continuity of the projector rank,
gives a simple top Perron root and a complement whose $k$th power has
norm at most $n^{O(1)}n^{-k/4}$. Apply this to $U_b^{3D}$; intermediate
powers have norm at most $n^{b/2}$ by row-substochasticity. In particular,
for some fixed $a>0$ and $C_0$,
\[
 U_b^j=\zeta_b^j H_b\mu_b + F_{b,j},\qquad
 \|F_{b,j}\|_{\infty\to\infty}\le n^{C_0}e^{-aj}.
 \tag{S1}
\]
Normalize $\mu_b\mathbf1=\mu_b H_b=1$. The contour around $\lambda$
may also be taken with radius $n^{-1+o(1)}$ by the same rank-one formula
with a slightly larger $o(1)$ exponent. Thus
$1-\zeta_b\le n^{-1+o(1)}$. For a sufficiently large fixed
$D'=C'\log n$, (S1) and row-substochasticity imply $\|H_b\|_\infty\le2$;
uniformly outside $n^{o(1)}$ exceptional single states
$H_1(y)\ge 1/3$. For the last statement the sum over initial states of
the probability of touching $x$ in $D'$ steps is at most $(D'+1)M$,
by the fiber bound for each fixed initial segment and averaging over words.

\paragraph{Quantitative separation of the two roots.}
Put $\sigma=\zeta_1$, $\tau=\zeta_2$. Write $C$ for the kernel from a
safe distinct pair to a safe common destination in one letter. From
uniform distinct states, a first meeting at $D'+1$ survives the deletion
and has $H_1$ at its endpoint at least $1/3$ with probability
$n^{-1-o(1)}$. To see this, expose a fixed full word. If one coordinate
visits the mark at time $j$, its initial state lies in a fiber of size $M$.
For each such state, the other coordinate must map to its final endpoint,
again a fiber of size at most $M$. A union over $j$ therefore removes at
most $O(D'M^2/n^2)$. A final join at one of $n^{o(1)}$ exceptional states
costs at most $n^{-2+o(1)}$. These bounds hold uniformly under the
original conditioning on distinct starts. Thus (S1) for $b=2$ and a
long enough burn imply $\mu_2 C H_1\ge n^{-1-o(1)}$. Set
$\Phi(i,j)=H_1(i)+H_1(j)$. Coordinatewise the identity for the sum of
the two individual eigenfunctions gives
$\sigma\Phi\ge U_2\Phi+2CH_1$, with the excess from steps on which
only one coordinate visits the marker. Left multiplication by $\mu_2$
and $\mu_2\Phi\le4$ yield $\sigma-\tau\ge n^{-1-o(1)}$.

\paragraph{A one-sided observable energy.}
Subtract the common trap mean $\theta$, extend $z=h-\theta$ by zero at
the mark, and write $U_w$ for the single killed transition along a
specific word. Let $B_l=\mathbb E_{|w|=l}(\mu_1 U_w z)^2$.
For two independent $\mu_1$ states driven by a common word, the mass
on distinct safe pairs and the mass on safe singletons evolve by
$u_{l+1}=u_lU_2$, $v_{l+1}=v_l U_1+u_l C$, starting at the off-diagonal
and diagonal pieces of $\mu_1\otimes\mu_1$. Consequently
\[
 B_l=A\sigma^l\mu_1(z^2)+L(z)\tau^l+O(n^{C_1}e^{-a'l}),\qquad
 A=v_0 H_1+\sum_{j\ge0}\sigma^{-j-1}u_0U_2^jCH_1 .
 \tag{S2}
\]
One derivation expands the triangular recurrence using (S1). The
convolution of the two slow terms is
$(\sigma^l-\tau^l)/(\sigma-\tau)$; a single fast term can be extended
to its infinite weighted sum with an exponentially small tail.
The fast--fast convolution is $O(n^{2C_0}l e^{-al})$. These estimates
are uniform for bounded $z$ and $\sigma,\tau\ge 1-o(1)$.
All terms in $A$ are nonnegative, and
$A\ge\sum_i\mu_1(i)^2 H_1(i)\ge1/(2n)$ by Cauchy--Schwarz.
In particular
\[
 B_{l+1}\ge B_{l+1}-\tau B_l
 \ge n^{-3}\sigma^l\mu_1(z^2)-O(n^{C_1}e^{-a'l}).
 \tag{S3}
\]
The learner never conditions on avoiding the mark.

\paragraph{Using only observed words and responses.}
At an exterior origin $y$, for a remaining task length $r$ and visible
suffix $w$ of length $l+1$, the actual conditional residual mean is
$e_y U_1^{r-l-1}U_w z$. The unconditioned target mean-square residual
equals $e_y U_1^r z^2$. For a detection threshold $d\in(0,1)$ set
$l=\lceil C_2\log(2nm/d)\rceil$. If $r<2l+2$, enumerate the entire
remaining word. Otherwise (S1) and (S3) show that whenever the mean-square
residual exceeds $d$, some suffix bin has absolute mean residual at
least $(2nm/d)^{-C_3}$. Indeed, with $s_y=\sigma^{r-l-1}H_1(y)$, we
have $s_y\sigma^{l+1}\mu_1 z^2\ge d/2$ and $s_y\ge d/8$.
The mean over suffixes of squared observable residuals is at least
$n^{-3}s_y^2\sigma^l\mu_1 z^2-O(n^{C_4}e^{-a''l})$.
The main term is $\Omega(n^{-3}d^2)$; the error can be made smaller
by the choice of $C_2$. Every suffix bin has inverse-polynomial
probability. Inside the trap every residual is exactly zero.

Use the original finite-basin construction's safe recurrent class $S$,
safe diameter $h_0$, and exact-length safe routing from any $y\in S$
to any $z\in S$ at every length at least $G$. All safe letters remain
in $S$. If the root has small risk relative to the constant $\theta$,
include the constant candidate. Otherwise fix a safe prefix $p$ with
$|p|\le W$ and let $L=W+G$. If $m'\le L$, use the full-word dictionary.
If $m'>L$, average over the first $L$ letters of the root task:
some such prefix $v$ retains conditional mean-square residual
at least $\epsilon/16$. It stays safe since any entrance makes
the remaining residual identically zero. From the endpoint of $p$
route safely in exactly $L-|p|$ steps to the endpoint of $v$.
Thus, given $p$, the residual risk over the remaining letters is
at least $d=2^{-L}\epsilon/16$. No transformation clock is needed.
For a prefix inside the trap the true continuation is the constant.
The suffix-bin test just proved therefore supplies the membership
decisions on all the required prefixes. Estimate every required
conditional mean by independent examples. The prefix and suffix
lengths are logarithmic, and the union bound costs a polynomial.
Guess $\theta$ on a grid finer than one tenth of the detection margin
and include the true branch among candidates. Final validation compares
squared losses; differences are unbiased for differences of mean-square
risks despite arbitrary conditional variances. The repaired clean-core
branch uses its separate bounded-regression learner. This proof uses
the high-confidence finite-basin geometry and clean heat and first-hit
estimates as inputs. A bounded closed component with nonconstant means
has additional internal modes that (S2) does not include.

\subsection{Learning arbitrary terminal means on the exceptional components}
\label{sec:stochastic-finite-action}

\begin{theorem}\phantomsection\label{thm:stochastic-finite-action}
Fix $A>0$. On a set of i.i.d. binary transition tables of probability
$1-o(n^{-A})$, the following holds for every start, every $m\ge1$,
and every vector $h\in[-1,1]^n$. Examples are independent pairs
$(W,Y)$ with $W$ uniform of length $m$, $|Y|\le1$, and
$\mathbb E[Y\mid W]=h(f_W(s))$. Given these pairs, a learner returns
a predictor with mean squared error at most $\epsilon$ for the
terminal mean on an independent word of the same length. Conditional
sample failure is at most $\gamma$. Its example count and arithmetic
cost are polynomial in $n,m,1/\epsilon,\log(2/\gamma)$, with degree
depending on $A$. For Bernoulli data the same holds for bit operations. For real-valued data
it holds in the usual bit model with access to successive binary digits
of a bounded response, counting the cost of those digits.
\end{theorem}

The geometry and zero-mean closed-component learner of the preceding
section give the only table-specific inputs. The reduction below
works for any deterministic finite action on the exceptional states.
It also works with input-dependent response variances. In particular
it applies to one Bernoulli observation per word with state-dependent
success probability. Squared error is relative to the conditional
mean; fresh observed squared loss includes the irreducible response
variance.

\paragraph{Conditioning on the action of a suffix.}
Let $T$ be the bounded saturated closed set in the table reduction.
If it is empty, use the clean bounded-regression result. Otherwise
guess its abstract transition action. There are at most $s^{2s}$
possibilities after numbering its $s=|T|$ states, a constant for each
fixed $A$. No internal state means are supplied or guessed. For the
correct action, write $G(v):T\to T$ for the transformation induced
by $v$ and $G_a$ for a single letter. Consider the following
increasing sequence of information sets:
\[
 \mathcal F_j=\sigma\big(W_{1:j},G(W_{j+1:m})\big),
                 \qquad 0\le j\le m.
\]
Indeed $G(av)=G(v)\circ G_a$, so revealing the next letter and
the remaining transformation retains the previous information.
Initially one sees only $G(W)$; at the end one sees the whole word.

At a split with $r=m-j\ge1$ letters remaining, write
$B=(a,G(v))$ and $C=G(av)$ for $av=W_{j+1:m}$. Define
$p_b=\Pr(B=b)$, $p_c=\Pr(C=c)$ under a fresh uniform suffix.
For each $b$ of positive probability there is a unique $c=c(b)$,
and $0<p_b\le p_c$. The probabilities are known from the guessed
action: multiply the probability vector on its at most $s^s$
transformations by the two-letter transition rule for lengths
$0,\ldots,m$. Their binary numerators and denominators have
$O(m+1)$ bits. Let
\[
 \psi_b(av)=\mathbf1_{B=b}-(p_b/p_c)\mathbf1_{C=c},\qquad
 g_{j,b}(x)=\mathbb E_{a,v}\,[\psi_b(av)h(f_{av}(x))].
\]
The contrast has absolute value at most one and mean zero on every
coarse cell. If $x\in T$ then $h(f_{av}(x))=h(C(x))$ depends on
$C$ alone. It follows that $g_{j,b}$ vanishes on the \emph{entire}
saturated closed set, including its transient states.

From each training word retain its first $j$ letters and use
$Y\psi_b(W_{j+1:m})$ as their response. Its conditional
mean is precisely $g_{j,b}(f_{W_{1:j}}(s))$. The suffix is random
given the visible prefix, and the defining conditional-mean
assumption on $Y$ is all that is used. For $j>0$, the preceding
constant-mean-on-$T$ learner applies with known internal mean zero.
For $j=0$, estimate the single number by its sample average.
No training operation tests the physical state, decides whether
the suffix entered $T$, or requests a second response to a word.

\paragraph{Reconstructing the prediction.}
Let $M_j=\mathbb E[Y\mid\mathcal F_j]$. For the actual prefix
ending at physical $x$, on the cell $B=b$ the conditional
increment is exactly
\[
                  M_{j+1}-M_j=g_{j,b}(x)/p_b.              \tag{F1}
\]
This follows by subtracting the two conditional cell means of
$h(f_{av}(x))$. The starting term $M_0$ is a function of at most
$s^s$ observable transformation cells. Fit it by ordinary cell means.
Finally $M_m=h(f_W(s))$. Thus the unidentified emission vector
appears only inside auxiliary tasks that have zero mean on $T$.

To bound the error in the sum of the increments, let $K=2s^s$, $L=m+1$,
$\eta=\epsilon/(64 K L^2)$, and
$e=\epsilon\eta/(64 K L^2)$. Skip a refined cell of probability
less than $\eta$ and predict a zero increment there. Its contribution,
summed over all skipped cells at one time, is at most $4K\eta$ because
$|M_{j+1}-M_j|\le2$. On other cells learn each $g_{j,b}$ to
squared error $e$ on uniform prefixes, divide by $p_b$ and clip to
$[-2,2]$. The prefix is independent of $B$. Hence the total
mean squared error of the predicted increment at a given time is
at most
\[
                 4K\eta+\sum_{b:p_b\ge\eta} e/p_b
                 \le 4K\eta+Ke/\eta.
\]
For $M_0$ use a bounded fit of squared error $\epsilon/(8L^2)$
on the finite cells. One may discard cells of mass
$\epsilon/(64KL^2)$ and apply conditional Hoeffding on the rest.
Adding the $L$ terms, Cauchy--Schwarz bounds total squared error
by $L$ times the sum of their squared errors, below $\epsilon/2$.
Clip the final prediction to $[-1,1]$. All conditional training
guarantees can hold together by assigning failure
$O(\gamma/(LK))$ to each fit. Independent sample batches are
sufficient. Sharing examples across fits also preserves this union
bound when each individual batch has i.i.d. rows.

Run the construction for every allowed finite action and size,
including the clean-table candidate. Select on fresh examples by
bounded empirical squared loss, to additive $\epsilon/8$ in risk.
Differences of the expected losses equal differences in error to
the true conditional mean. All good-table quantifiers precede the
choice of the auxiliary means and lengths. Resource polynomials
therefore remain uniform in $h,m,\epsilon,\gamma$ for fixed $A$.
For real input, first replace each $Y$ by the sign of $Y-U$, using
an independent uniform $U$ in $[-1,1]$. The new response has conditional
mean $Y$ and takes two values. Given a bound $N$ on the total sample
count, compare using $O(\log(2N/\gamma))$ bits per response and abort
if the intervals overlap. A union bound over draws of $U$, conditional
on their responses, makes the probability of any abort at most $\gamma/4$.
On all other executions the comparison is identical to exact rounding.
Use the remaining failure allowance for the fits and validation. All
subsequent observation weights are rational with polynomial bit length.

\subsection{A quadratic number of suffix tests for a supplied finite action}
\label{sec:suffix-detail-regression}
The preceding reduction admits a version whose cost is polynomial in the number of states in the supplied exceptional action, even if its transformation
monoid has exponentially many elements. We reduce estimation of an arbitrary terminal mean to estimation of means that vanish on the exceptional set. The reduction produces independent observations with the bounded conditional second moment required by the latter learner.

For a given action of size $s$, let $E_r$ be the space spanned in
uniform $L^2(\{0,1\}^r)$ by the indicators
$e_{xz}(v)=\mathbf1\{f_v(x)=z\}$ for $x,z\in T$, together with
the constant. Its dimension is at most $s^2+1$. The spaces
$V_j=L^2(\{0,1\}^j)\otimes E_{m-j}$ on the whole input are nested:
each one-letter restriction of $e_{xz}(av)$ is the indicator
$e_{f_a(x),z}(v)$. In particular $V_m$ is the whole function space.
Putting $r=m-j$, define
\[
 D_r=\big(L^2(\{0,1\})\otimes E_{r-1}\big)\ominus E_r .
\]
Then $V_{j+1}\ominus V_j=L^2(\{0,1\}^j)\otimes D_r$.
Choose mutually orthogonal rational bases for $E_m$ and each $D_r$,
normalizing their elements $d$ to have squared norm
$1/4\le v_d=\mathbb E d^2\le1$.

For every $d\in D_r$ define $g_d(y)=\mathbb E_{|v|=r}d(v)h(f_v(y))$.
For $y\in T$ this is exactly zero because $h(f_v(y))\in E_r$.
For any $y$, Cauchy--Schwarz gives $|g_d(y)|\le1$.
Use the actual training prefix and the response $Yd(W_{j+1:m})$
to learn $g_d$ at length $j$. If $\mathbb E[Y^2\mid W]\le1$,
the transformed response has conditional second moment at most one
given the entire prefix, since its independent suffix is uniform.
The resulting coefficient in the projection onto $V_{j+1}\ominus V_j$
is $g_d(f_{W_{1:j}}(s))/v_d$. The starting coefficients on $E_m$
are ordinary scalar response correlations divided by the same
normalizations. The final orthogonal sum gives the terminal mean.
For any fitted coefficient functions computed from training data,
the squared error on a fresh word adds \emph{exactly} across times
and detail directions before final clipping, by the displayed
orthogonality of the spaces. Each error $e$ in an estimated $g_d$
therefore contributes at most $4e$. Use $e=\epsilon/(32(m+1)(s^2+1))$
and at most $2(m+1)(s^2+1)$ fits. This bound is independent of the size of the transformation monoid and of the probability of its rarest suffix transformation.

All computation uses the supplied abstract transitions only. Compute
inner products of indicator generators using the two-coordinate chain
on $T^2$ driven by common letters. For the first-letter split multiply
the corresponding inner products by $1/2$ on a matching letter and
by zero otherwise. Process the coarse generators first and then the
fine ones, omitting zero pivots. Project each new generator off the
span of previously selected independent generators using the exact
rational Gram system on those original generators. The new orthogonal
vectors obtained after the coarse ones span $D_r$. Gram entries have
common denominator dividing $2^r$. Determinant bounds in dimension
$O(s^2)$ give $O(s^2(r+\log(2s)))$ bits for the numerator and
denominator of a projected coefficient up to a fixed constant
factor. Use an integral power of two to give squared norm in
$[1/4,1]$. Thus even exponentially small nonzero eigenvalues need
only a polynomial number of bits. There is no inverse-eigenvalue
factor in the variance bound or sample count. Each feature can be
evaluated by propagating the $s$ states along the observed suffix and
combining the rational coefficients.

\paragraph{Response estimation with bounded conditional variance.}
The zero-on-$T$ proof reads stochastic responses in conditional
mean estimates, in fitting a finite terminal readout, and in
fresh validation. The moment construction multiplies separately
estimated means; it does not request high powers of the same
observed response. Estimate each required mean by medians of
independent group averages. Inside a prescribed word bin the
conditional second moment remains at most one. For a readout with
$k$ cells estimate their probabilities and their unnormalized
response means; cells of mass $O(\eta/k)$ in total cost $O(\eta)$
when their predictions and true means are bounded by one. On the
others inverse-polynomial absolute precision suffices and the
outputs are clipped. For validation among bounded candidates $f$,
use $f(W)^2-2Y f(W)$ or paired differences of these scores. For the single score and $|f|\le1$, its conditional $L^2$ norm given
the fitted predictor is at most $1+2\|Y\|_2\le3$ on a fresh panel.
Its expectation equals the mean-square risk up to a term common
to all candidates. No fourth response moment is needed.
The same modifications apply on the clean component: all responses
enter there through primitive word-bin means, terminal readouts and
validation. The finite-dimensional random dictionaries and their
concentration use bounded \emph{mean functions} on the latent
states. If a bounded statistical-query implementation is desired,
for a linear query $X$ of second moment at most $V$, clipping it at
$T=2V/\eta$ incurs mean bias at most $\eta/2$, and the resulting
function divided by $T$ is bounded. Apply this only to individual
primitive estimates; clipping the response once could change its
wordwise mean on the exceptional set. For bit-input data, even the
largest normalized suffix feature on a length-$r$ atom is at most
$2^{r/2}$, and the rational bounds above cover its arithmetic cost.

\subsection{Resilience to a hidden closed modification}
\label{sec:closed-edit-resilience}
The size of a closed exceptional region can grow with $n$.
Let $A$ be an i.i.d. transition table on the ordinary clean event
for the one- and two-coordinate heat and projection estimates, the
short-word fiber bounds, the ordinary stationary floor, the
one-step uniform-start first-meeting estimate, and the clean
bounded-regression learner. The resilience assertion below is deterministic given these inputs.
For its binary application, Appendix~\ref{sec:iid-clean-completion}
lists the transition predicates used on a clean completion and bounds
their joint failure on the no-small-terminal stratum at each fixed
confidence exponent.
Fix $a>0$. After seeing $A$, choose $T\subset[n]$ of size
$s\le n^{1-a}$ and replace every transition out of $T$ by any
transition into $T$. Outside $T$ leave the original letters unchanged.
The accepting means can depend on both tables. The subsequent
machine, including its fixed initial state, is hidden.

\Needspace{12\baselineskip}
\begin{theorem}\phantomsection\label{thm:closed-edit-resilience}
Simultaneously for all such modifications of a clean table, a
passive learner supplied only the abstract internal action on $T$
in addition to $n$ learns the resulting terminal mean at its
training length $m$. For bounded responses with the specified
state-dependent conditional mean, risk $\epsilon$ and conditional
training failure $\gamma$, the sample count and bit complexity
are polynomial in $n,m,1/\epsilon,\log(2/\gamma)$ with degree
depending on $a$ and the clean-event constants. The state identifiers
in $T$, the original or exterior table, and all terminal means
remain hidden. Without the abstract action, an upper bound $s$ on
its size suffices with a multiplicative computational overhead
$s^{O(s)}$. More precisely the number of fixed-geometry sublearners
per action is $O((m+1)(s^2+1))$.
\end{theorem}

By Section~\ref{sec:suffix-detail-regression} it suffices to learn
a bounded mean $h$ which vanishes on all of $T$ using observations
with uniformly bounded conditional second moment. We may assume
$s\ge1$. The case of no modification is the clean learner.
All constants below may depend on $a$. For every set $B$ of size
at most $n^{1-a/2}$ let $U_1^B$ be the clean single walk killed on
visiting $B$, and $U_2^B$ the clean pair killed on a merger or
either visit to $B$. These are zero-padded at forbidden starts.
The middle-block argument in Section~\ref{sec:stochastic-singleton}
holds simultaneously over all these sets. In fact the forbidden
slice has $O(|B|n^{b-1})$ tuples for $b=1,2$, so the operator error
from an interior visit is at most $|B|n^{-1+o(1)}\le n^{-a/2+o(1)}$.
To separate the clean error from the deletion cost, take the clean
burn $D$ of Appendix~\ref{sec:stochastic-singleton}, with
$P_b^D=\rho_b^D h_b\nu_b+E_b$. Put $V=(U_b^B)^D$ and
$L_B=\rho_b^D Vh_b\nu_b V$. Then
\[
 \|(U_b^B)^{3D}-L_B\|_2
 \le \|V E_b V\|_2+|B|n^{-1+o(1)}.
\]
Indeed, the interior-visit estimate compares the first operator to
$VP_b^D V$. Short-word fibers give $\|V\|_2\le(\log n)^{O(1)}$
by the row and column sum bounds. Increasing the clean burn makes
the clean remainder polynomially small to any fixed order.
The rank-one eigenvalue is $\lambda_B=\rho_b^D\nu_b V^2h_b$.
Under $\nu_b$, a union bound over $2D+1$ visit times costs
$|B|n^{-1+o(1)}$, since $\nu_bP_b^j=\rho_b^j\nu_b$,
$\|\nu_b\|_\infty\le n^{-b+o(1)}$ and $\|h_b\|_\infty=O(1)$.
Thus $0\le\rho_b^{3D}-\lambda_B\le |B|n^{-1+o(1)}$.
When $B$ is empty, the two clean terms are still present:
the middle-block remainder and, for pairs, loss by meeting.
Applying the
rank-one resolvent on a contour of polynomially small radius and
taking block powers as there gives, uniformly,
\[
 (U_b^B)^k=(\zeta_b^B)^k H_b^B\mu_b^B+F_{b,k}^B,\qquad
 \|F_{b,k}^B\|_{\infty\to\infty}\le n^{C_a}e^{-c_a k},
 \quad 1-\zeta_b^B\le n^{-c_a}.
 \tag{R1}
\]
The normalized nonnegative factors satisfy
$\mu_b^B\mathbf1=\mu_b^B H_b^B=1$, $\|H_b^B\|_\infty\le2$.
Here a single fixed $O_a(\log n)$ burn already gives this last bound.
For the actual $T$ write $H=H_1^T$, $\sigma=\zeta_1^T$. At
most $n^{o(1)}s$ states have $H<1/3$: average the chance to visit
$T$ in a sufficiently long fixed logarithmic burn from each state
and sum over states, using the uniform fiber bound at each time.
Then apply (R1). This estimate will place a data-tested enlargement
of $T$ within the same spectral regime.

\begin{lemma}[Spectral separation after deleting states]\phantomsection\label{lem:deletion-root-gap}
Fix $0<a\le1$. On the clean table event for (R1) and the uniform
history estimate of Appendix~\ref{sec:history-tree}, the following
holds simultaneously for all $B\subseteq[n]$ of size at most $n^{1-a}$.
Write $\sigma_B=\zeta_1^B$ and $\tau_B=\zeta_2^B$.
Let $C_B$ send a live ordered pair to its common endpoint if the
next letter merges it outside $B$, with zero mass otherwise.
With the normalization in (R1),
\[
 \mu_2^B C_B H_1^B=\frac{1+o(1)}{n},
 \qquad \sigma_B-\tau_B\ge\frac{1-o(1)}{n}.
\]
Both errors are uniform in $B$. At $B=\varnothing$,
$n(\sigma_B-\tau_B)\to1$ on this event, so the leading constant
in the uniform lower bound is sharp.
\end{lemma}

\begin{proof}
Fix a table in the event and write $s=|B|$, $H=H_1^B$, extending
$H$ by zero on $B$. Take $D=\lceil c\log n\rceil$ so that the row
errors in (R1) at $D$ are at most $e=n^{-5}$ for both kernels.
The same $D$ works for every $B$ by the forbidden-slice estimate
preceding (R1). Let $M=(\log n)^{O(1)}$ bound all word fibers
through length $D+1$ on this table. By (R1),
$\sigma_B^D=1-o(1)$ uniformly, and
$\|(U_1^B)^D\mathbf1-\sigma_B^D H\|_\infty\le e$.
Thus $\|H\|_\infty\le(1+e)/\sigma_B^D=1+o(1)$.
Summing the probability of a visit to $B$ by time $D$ over all
single starts gives
\[
 \sum_y(1-H(y))_+\le (D+1)sM+ne .
\]
Indeed, at each time and for each fixed word, at most $sM$ starts
can be in $B$, including visits at time zero.

Let $u$ be uniform on the $N=n(n-1)$ ordered distinct starts.
On the undeleted table, two independent uniform starts first meet at $D+1$
with probability
$(1+o(1))/n$: apply the distinct-to-equal estimate of
Appendix~\ref{sec:history-tree} with the first $D$ letters as the
history and the $(D+1)$st as the next letter, then average over words. Conditioning the two
uniform starts to be distinct multiplies this mass by $n/(n-1)$.
For a fixed word, coordinate and visit time, at most $sM$ starts
of that coordinate can visit $B$. Each can meet at most $M$ opposite
starts at the terminal step. Removing meetings with any visit to
$B$ therefore costs at most $2(D+2)sM^2/N=o(1/n)$. At a prescribed
terminal state at most $M^2$ ordered starts can meet. Weighting each
remaining meeting by $H$ costs at most
$M^2[(D+1)sM+ne]/N=o(1/n)$ in the lower bound. The upper bound
uses $\|H\|_\infty\le1+o(1)$. We obtain
\[
             u(U_2^B)^D C_BH=(1+o(1))/n .
\]
The uniform-history Gram estimate gives probability $O(D/n)$ of
meeting by $D$ before deletion. The probability of either coordinate
visiting $B$ by $D$ is at most $2(D+1)sM/n$. Hence
$u(U_2^B)^D\mathbf1=1-o(1)$ and (R1) gives
$\tau_B^D uH_2^B=1+o(1)$. Applying (R1) again, now to the bounded
test $C_BH$, proves the flux formula. All losses above use only the
size of $B$ and the same table-wide fiber and kernel bounds. They
therefore apply also to a set selected after inspecting the table.

Put $\Phi(i,j)=H(i)+H(j)$ on live pairs. Apply the single-walk
eigenfunction equation to both coordinates with the same next letter:
\[
 \sigma_B\Phi=U_2^B\Phi+2C_BH+V_B .
\]
Here $V_B\ge0$ is the expected $H$-value of the remaining coordinate
when just one enters $B$. Multiplication by $\mu_2^B$ gives
$(\sigma_B-\tau_B)\mu_2^B\Phi\ge 2\mu_2^B C_BH>0$.
Since $0<\mu_2^B\Phi\le2+o(1)$, the lower bound follows. With $B=\varnothing$,
the ordinary walk has $\sigma_B=1$ and $H=\mathbf1$, while $V_B=0$.
The identity then gives $1-\tau_B=\mu_2^B C_B\mathbf1=(1+o(1))/n$.
\end{proof}

The accompanying Lean module \texttt{DeletionRootGap} verifies the
finite counting inequalities, the one-letter balance, and the scalar
gap and sharpness deductions given their flux and projection inputs.

Apply Lemma~\ref{lem:deletion-root-gap} at the actual modified set $T$.
The convolution in (S2) has denominator $\sigma_T-\tau_T$; the
lemma bounds its inverse by $(1+o(1))n$ and gives the energy
estimate (S3) used below.

\paragraph{Making a low-survival set observable.}
Set $u=\epsilon/[10^4(m+1)^2]$.
If the mean squared signal at the designated start is at most $\epsilon/8$,
the zero predictor already achieves this risk. Otherwise use the logarithmic
suffix proof (S1)--(S3) at the root to obtain some word $w$ of
length $l=O_a(\log(2nm/\epsilon))$ such that
\[
 |K|=|\sigma^{m-l}\mu_1^T U_w^T h|\ge(2nm/\epsilon)^{-C_a}.
 \tag{R2}
\]
Indeed the suffix conditional mean from the root is $K H(s_0)$
up to arbitrarily small polynomial error, where $s_0$ is the
prescribed start, and $H(s_0)\le2$. A short $m$ is handled by
a full-word dictionary whenever the required windows overlap.
For each prefix $p$ of length at most the required access and
testing horizon, ordinary labeled words with prefix $p$ and suffix
$w$ give the conditional mean $b_p$. By (R1),
\[
       \sigma^{|p|} b_p=K H(f_p(s_0))+
                 O(n^{C_a}e^{-c_a(m-|p|-l)}).
 \tag{R3}
\]
If $p$ enters the closed set its true endpoint lies in $T$ and
both sides have zero mean. Products of word-bin probabilities
remain inverse polynomial since the visible prefix and suffix
lengths below are logarithmic. If $m$ is less than their sum
plus the necessary logarithmic mixing margin, the full-word
dictionary is polynomial.

Among the numbers $H(y)$, choose a threshold
$\alpha\in[u,2u]$ at distance at least $u/[3(n+1)]$ from
all of them (split the interval at these at most $n$ points).
Put $D=\{y:H(y)<\alpha\}$. It contains $T$, has size
$s n^{o(1)}\le n^{1-a/2}$, and has this separation
simultaneously at all tested prefixes. Enumerate $w$, a sign,
a fine rational grid for $\sigma\in[0,1]$, and a threshold for
$|K|\alpha$ in $[0,2]$. Mesh sizes polynomial in
$\epsilon/(2nm)$ suffice by (R2), (R3), and
$|z^k-z'^k|\le k|z-z'|$ on the unit interval.
Together with empirical errors smaller than a fixed fraction of
$|K|u/(n+1)$, one branch recognizes exactly whether every
queried endpoint belongs to $D$. Values $K$, $H$ and $D$ themselves
are never supplied. Once a trajectory reaches $D$, ignoring it
with at least $t$ letters remaining has small cost:
for $y\in D$ and $r\ge t$, $\mathbb E h(f_V(y))^2\le
2u+n^{C_a}e^{-c_at}$, by (R1) for $T$ and $h|_T=0$.

\paragraph{One-sided entrance from a surviving pair.}
Suppose first that $T$ intersects the clean recurrent class.
Its stationary law $\pi$ has $\pi(D)\ge n^{-C}$ by the stationary
floor. There is a fixed logarithmic $d_0$ and a function $b$ on
distinct safe pairs, $0\le b\le1$, giving the chance that their
first visits to $D$ differ before or at step $2d_0$, before merger.
We claim $\mu_2^D b\ge n^{-C'}$. Start two independent uniform
states with the same random word of length $3d_0$. The first
ends in $D$ with probability at least $\pi(D)/2$ by ordinary
mixing. The chance it does so after already visiting $D$ in the
first $d_0$ steps is $o(\pi(D))$: the latter visit has probability
$n^{-a/2+o(1)}$ by the fixed-length fiber bound, and from that
visit at least $2d_0$ ordinary steps remain, giving conditional
final probability at most $2\pi(D)$. Conditional on the whole
word and the first state, the independent second uniform state
visits $D$ in any of the $3d_0$ steps or at the start with
probability at most $(3d_0+1)M|D|/n=o(1)$, uniformly in this
word. If it never visits and the first ends in $D$, they could
not have merged before the first visit. This supplies probability
$\Omega(\pi(D))$ from uniform distinct states of survival for
$d_0$ steps and then a one-sided entrance within $2d_0$.
Apply (R1) with $d_0$ large enough relative to the stationary
floor. If two bright states have identical first-entrance behavior
on every word of length at most $t$, their killed pair after
$t-2d_0$ steps assigns zero mass to $b$ in this test. The row
projection and $\mu_2^D b\ge n^{-C'}$ show that the chance of
remaining distinct and both bright through $t$ steps is at most
$n^{C''}e^{-c_at}$. This uses the killed pair for $D$ itself;
the inclinations of paths after a simultaneous entrance into two
different dark states play no role.

For any remaining length $r\ge2t$ it follows that replacing
one bright state by another of the same entrance signature changes
the final response on the shared future by expected squared
difference $O(u+n^{C''}e^{-c_at})$. If both enter within $t$,
condition on this stopping prefix. Each mean has expected square
at most $2u+o(u)$ because at least $t$ letters remain. If both
stay bright and physically coalesce there is zero difference.
The remaining distinct probability costs at most four times
the preceding bound.

\paragraph{Covering the relevant states without a safe diameter.}
Use a block of length $g=C_a\log(2nm/\epsilon)$ long enough for
(R1) to arbitrary required polynomial precision. Choose $t$ of
the same order with the previous coupling error at most $u$.
The constants in $l,g,t$ can all be chosen from fixed table
exponents, with the extra separation precision for (R3) paid
by the remaining free middle letters. If necessary enlarge the
dictionary cutoff for $m$ once after fixing these three lengths.
Write $\eta=u/n$. Call $z$ heavy when $\mu_1^D(z)\ge\eta$.
If root risk exceeds $\epsilon/8$, its probability of staying
bright through $g$ steps is at least $c\epsilon$ when $m-g\ge t$:
on any earlier entrance, the remaining mean-square response is
at most $2u+o(u)$. Thus (R1) gives
$(\zeta_1^D)^gH_1^D(s_0)\ge c'\epsilon$. For every heavy $z$
the corresponding entry of $(U_1^D)^g(s_0,z)$ is positive, by
choosing its absolute error below $c'\epsilon\eta/2$.
Every heavy state therefore has an accessible all-bright prefix of
length exactly $g$ from the actual root. From \emph{any} bright
state, the chance to be bright throughout a fresh $g$-block and
end at a nonheavy state is at most $2n\eta+o(u)$ by (R1).
Both assertions hold for sets selected using the table. No access
word or positivity of stationary mass for every exterior state
is required.

Compute first-entrance signatures through $t$ for the root and
all all-bright $g$-letter access prefixes. Store one representative
per signature. Given a representative and a fresh $g$-letter
block, the tests along its prefixes tell whether darkness was
hit. If safe, the tests after the block give its endpoint signature.
Replace by the stored representative when available; otherwise
terminate with output zero. All required prefix lengths are at
most $2g+t$, and all enumeration and data bins have polynomial
size. A heavy endpoint always has a stored representative.
Process such blocks while leaving $2t$ to $2t+g-1$ final letters.
A terminated path outputs zero. Otherwise the final representative
and the remaining short suffix form a key. For the comparison
predictor assign each key the true response obtained by reading
that suffix from the actual physical representative. The key has
a fixed well-defined comparator even when the algorithm has no
knowledge of its value.

Compare successive hybrids that perform the first $j$ and the
first $j-1$ block operations and then use the real table for
the rest. Conditional on an active representative before a fresh
block, a dark hit costs squared difference at most $2u+o(u)$;
every possible hit leaves at least $2t$ future letters. A
missing signature costs probability $2n\eta+o(u)$ from this
very representative, regardless of how it was selected.
A represented safe endpoint has the same signature as its replacement
and incurs squared difference $O(u+n^{C''}e^{-c_at})$ on the
fresh remaining word. Already terminated histories coincide.
The triangle inequality in $L^2$ over at most $m$ replacements
gives a final mean-square comparison error $O(m^2u)$.
Take the fixed numerical denominator in $u$ sufficiently large
for these universal bounded-output costs. Fit the bounded terminal
lookup from fresh observations and validate all branches. There
are at most $2^g+1$ representatives even for incorrect branches,
so the runtime bound holds on every sample.

If $T$ is disjoint from the clean recurrent class, split instead
on a sufficiently long logarithmic prefix. Except on total word
mass $O(\epsilon)$ the path has either already hit $T$ (and has
zero terminal mean), or reached the clean recurrent class before
hitting $T$ and will never touch it. This follows by coupling the
clean process before first entrance and ordinary logarithmic
mixing, with its stationary mass on the core. Within each bin, fit both a constant and the clean all-start regression learner. Validation selects between candidates without knowing which geometry holds. The construction for a zero-on-$T$ task uses no
state identifiers, internal action, or dependence on $s$ in its
polynomial degree. The quadratic-size detail reduction and fresh
validation over the optional action guesses now prove the theorem.

\subsection{Completing the exterior under the original i.i.d. prior}
\label{sec:iid-clean-completion}
We now apply the row-edit theorem to the original random table. For each fixed confidence exponent $A$, we allow arbitrary state-dependent terminal means and keep the action on the exceptional closed set hidden from the learner.

Use the clean estimates at exponent $A'=2A+10$. There is a
constant $k$ such that an i.i.d. table with no closed component of
size at most $k$ satisfies those estimates outside an original
exception of probability $o(n^{-A'})$. The clean inputs include
all the uniformly bounded-mean estimates listed in the proof of
clean regression and the uniform-start first-meeting floor.
Let $T$ be the union of all closed sets of size at most $k$ in
the original table. For any fixed $d> A+1$, if $|T|>d$
then successively adding such sets produces a closed set of size
between $d$ and $d+k$. The chance of any one of these sizes $j$
is at most $\binom{n}{j}(j/n)^{2j}=O_j(n^{-j})$.
Thus outside $o(n^{-A})$ the union has bounded size, say $s\le d$.
No prescribed output values are involved. The maximum in-degree
of the original graph is at most $\log^2 n$ except on
superpolynomially small probability, by the binomial tail and a
union bound.

For a specified nonempty $T$ of size $s\le d$, resample its $2s$
letter cells independently uniformly in $[n]$, keeping the
exterior rows. Conditional on the specified closure of $T$ alone,
this completed table has exactly the ordinary i.i.d. law.
We also need a lower bound on the chance that such a completion
has no small closed class on the event that this $T$ is the actual
union above. Every state outside $T$ has original forward reachable
set of size more than $k$. Except for at most
$s(1+\log^2 n)^{k+1}=o(n)$ states, its first $k$ forward layers
avoid $T$, by the original reverse-neighborhood bound.
For any remaining state these layers already contain more than
$k$ distinct vertices: if they stabilized at size at most $k$,
the entire forward orbit would be a small closed set.
With probability $1-o(1)$ all resampled heads land at such states.
A small closed set disjoint from $T$ would have been closed in the
original table and hence included in $T$. A closed set meeting
$T$ now contains one of these heads and its more than $k$ distinct
exterior descendants. Thus every sufficiently large $n$ gives at
least $1/2$ probability of a completion with no small closed class
on each retained original exterior and actual union event.

If such an original table admits no good completion, at least
half its completion draws have no small closed class but fail the
clean estimates. In the numerator one may discard the actual-union
and reverse-degree restrictions and condition only on specified
closure. Summing over choices and sizes bounds this part of the
exception by
\[
        2\sum_{s=1}^d {n\choose s}(s/n)^{2s}o(n^{-A'})
                           =o(n^{-A}).
\]
When $T$ is empty the original clean guarantee applies. Otherwise
witness one clean completion, with all exterior rows unchanged.
The actual table is obtained from that completion by replacing
the rows in $T$ with its true closed action. Apply
Theorem~\ref{thm:closed-edit-resilience}, enumerating every action
of size at most $d$. This gives the full same-length terminal-mean
learning result at joint failure $o(n^{-A})+\gamma$ in polynomial
resources for each fixed $A$. Deterministic final bits are the
special case with a noiseless response; threshold a mean predictor
at $1/2$, using squared error $\epsilon/4$ for classification
risk $\epsilon$. Bernoulli responses also give excess Bayes
classification error at most $2\sqrt{\epsilon}$ at squared error
$\epsilon$ when approximating success probabilities. With a
state-dependent positive margin, the error against the optimal
terminal sign can be bounded by the squared error divided by
the squared margin.

\paragraph{Choosing the classifier using terminal labels.}
For the noiseless classification statement in
Theorem~\ref{thm:poly-learning}, condition on all examples and coins
used to construct the candidates. Threshold and clip their predicted
means as above. Run the preceding fits at squared-error tolerance
$\epsilon/8$ and conditional failure at most $\gamma/2$ for the
correct branch. Its classification risk is then at most $\epsilon/2$.
If there are $J$ candidates, draw a fresh batch of
$\lceil8\epsilon^{-2}\log(4J/\gamma)\rceil$ complete labeled words.
Hoeffding's inequality and a union bound put every empirical
classification risk within $\epsilon/4$ of its conditional population
risk with probability at least $1-\gamma/2$. Choosing an empirical
minimizer gives risk at most $\epsilon$. The event defining the
correct table completion depends only on the transitions. The
conditional training failure and that table exception thus give the
two probabilities stated in Theorem~\ref{thm:poly-learning}.

\paragraph{A sample and precision schedule for Theorem~\ref{thm:poly-learning}.}\label{sec:binary-schedule}
Here the observation is a complete uniform $m$-letter word and one
bit $Y$. Set $A'=2\beta+10$, take the associated clean cutoff $k$,
and put $d=\lceil\beta+2\rceil$. Enumerate every binary action on
at most $d$ states, at most $1+\sum_{s=1}^d s^{2s}$ choices including
the empty action. A clean completion is only used to prove that one
of the enumerated candidates has low risk. The algorithm samples
from the original labeled machine throughout.

For binary data take $a=1/2$ in the row-edit estimates; eventually $d\le n^{1/2}$. The required clean event is transition-only.
At exponent $A'$ with one extra unit of slack, take the intersection
of the centered pair block and the uniform pair heat bound from
Appendix~\ref{sec:centered-edits}, the short-word fiber bound from
Appendix~\ref{sec:one-block-learner}, and the finite-distance bound
below. The antisymmetric part of the centered block gives ordinary
mixing, and the fiber bound from uniform roots then bounds the largest
stationary atom. A path from a heaviest stationary state supplies the
stationary floor. The distinct-to-equal estimate for a prescribed
common history in Appendix~\ref{sec:history-tree}, averaged over
histories, supplies the uniform-root one-step meeting floor. The
power estimate in Appendix~\ref{sec:approx-sink} gives the
logarithmic-burn row projection. In particular, if $H_i$ is surviving
mass from two independent stationary starts, the two stationary
history-Gram formulas at $i=l,l+1$ give
$H_l-H_{l+1}=(1+o(1))/n$ and $H_l=1-O(l/n)$. The row projection at
a sufficiently long fixed logarithmic burn gives
$|H_{l+1}-\rho H_l|=O(n^{-3})$. Hence $n(1-\rho)=1+o(1)$ on this
same stratum, providing the denominator guard. Finally intersect the table events for the
history overlap (Appendix~\ref{sec:history-absolute-gap}), the bounded
heavy-neighbor degree (Appendix~\ref{sec:lifted-suffix-all}) and
the exceptional-pair matching
(Appendix~\ref{sec:exception-pair-matching}). Use only the structural
bounds in those arguments. Their consequences for a sign mask then
hold simultaneously on the recurrent class. There are finitely many
fixed logarithmic multipliers after $\beta$ is fixed. The resulting
table exception on the no-small-component stratum is $o(n^{-A'})$;
the resampling calculation above therefore applies to this intersection.

Let $S_k$ denote the event that the i.i.d. table has no terminal component of size at most $k$ and let $G$ be the intersection just described. The
following budgets are for the original independent rows. Each
fixed logarithmic window is chosen before the table is drawn.
\begin{center}\small
\begin{tabular}{@{}p{.29\linewidth}p{.66\linewidth}@{}}
\hline
Event or estimate & Where its confidence is used \\ \hline
Four-source moment &
Lemma~\ref{lem:poly-centered} gives the typical median on the original law.
Its one-surplus search need only succeed with probability $1-o(1)$.\\
Clean pair heat and centered block &
Appendix~\ref{sec:centered-edits} allows a cut-off depending on $A'$,
controls the heat on $S_k$, and applies bounded differences to the
row-Hamming extension under the i.i.d. law. The intersection of $S_k$
with failure costs $o(n^{-A'})$.\\
Histories and output tests &
Uniform fibers and the centered block give ordinary mixing; the
finite-distance estimate below gives the stationary floor.
Appendices~\ref{sec:history-tree}, \ref{sec:history-absolute-gap} and
\ref{sec:exception-pair-matching} give the meeting, variance and pair
separation estimates on that stratum at the same budget with slack.\\
Completion and data &
$\Pr(S_k\cap G^c)=o(n^{-A'})$ enters the specified-closure resampling
sum above. The resulting original-table exception is $o(n^{-\beta})$.
Conditional on the resulting target, the candidate fit and held-out
selection together cost at most $\gamma$.\\ \hline
\end{tabular}
\end{center}

There are two possibilities for a witnessing completion. Write
$\mathcal K$ for its recurrent class. If $T\cap\mathcal K$ is empty,
burn a prefix of length $g_0=C_\beta\log(2nm/\epsilon)$.
Couple the original and clean walks until the first visit to $T$.
The probability of finishing this prefix outside $T\cup\mathcal K$
is at most the clean mixing error, chosen below $\epsilon/16$.
A prefix ending in $\mathcal K$ gives an ordinary clean binary
experiment on its remaining letters. A prefix ending in $T$ gives
a DFA on at most $d$ states. In each prefix bin fit the clean binary
learner and enumerate all such small machines with their masks and
starts. Select within each bin on fresh labeled words. Conditional on
a prefix, the remaining letters are independent uniform bits. With
conditional risk at most $\epsilon/8$ for a correct candidate and
selection overhead at most $\epsilon/8$ in each good bin, the
combined predictor has risk less than $\epsilon/2$. Each prefix bin has probability $2^{-g_0}$, so collecting examples within bins costs a factor $2^{g_0}$ in sample size. Divide the failure allowance among the bins. For $T=\varnothing$ the same argument uses the clean
learner alone. If the prefix or clean windows do not fit, use the
short-word rule below. 

For $T\cap\mathcal K\ne\varnothing$, apply
Lemma~\ref{lem:deletion-root-gap} on this same witnessing completion
with $B=T$ and $a=1/2$. Since $|T|\le d$, the two killed roots
are separated by at least $(1-o(1))/n$. The pre-entrance kernels
use only exterior rows, which agree with the original table.
This supplies the denominator in (S2) for the zero-mean detail
fits. Use the orthogonal detail tests of
Appendix~\ref{sec:suffix-detail-regression}. Bound their total number,
including the coarse tests, by $J=3(m+1)(d^2+1)$ and let
$r=\epsilon/(32J)$, $U=2n(m+1)/r$. For a detail test at prefix length $j$, transform a fresh training example $(W,Y)$ into
$(W_{1:j},Y d_*(W_{j+1:m}))$. The learner computes $d_*$ from the candidate action. For the true action, the transformed response has mean zero on $T$. Its second moment conditional on the full $j$-letter prefix is at most one. The same bound holds after conditioning on any bin of
that input. There is no rejection on the basis of a hidden state.
Coarse means (including $j=0$ when appropriate) need only scalar
estimation. Table~\ref{tab:fit-parameters} lists the remaining choices in dependency order. All constants in logarithmic windows are fixed once the clean
exponents at $A'$ have been specified.

\begin{table}[!ht]
\caption{Parameter choices for the clean decoder and residual fits.}
\label{tab:fit-parameters}
\centering\small
\begin{tabular}{@{}>{\raggedright\arraybackslash}p{0.19\linewidth}p{0.78\linewidth}@{}}
\hline
Quantity & Choice and use \\ \hline
Clean binary decoder & Let $v$ be the dictionary dimension, $k_s$ the
number of sign tests, $h_*$ the fixed product-recursion depth, and
$\lambda_*=w_*v_*^2$, where $w_*,v_*$ are deterministic clean lower bounds on $\pi_{\min,+},\sigma_{\min}(V)$, respectively. Choose signature depth $L_s=C_\beta\log n$
by Appendix~\ref{sec:exception-pair-matching}. Use entry tolerance
$e_* =\lambda_*/[32v(k_s+1)]$ in all decoder matrices and shift
$\lambda_*/4$ in the PSD test.\\[2pt]
Clean bin precision & Let $b_*$ be the clean inverse-polynomial lower
bound on $1-B_l$ for a nonconstant core. Use primitive accuracy and
population diagonal error at most $b_*e_* /[C_*(16n)^{2h_*+6}]$.
Here $C_*$ bounds the coefficient sums in the fixed-degree expansions
of the normalized features (a polynomial bound suffices). Choose the
whole-block product lag and prefix mixing burn \emph{after} this
accuracy, both $O_\beta(\log n)$. Use the word-bin sample row below
with the clean bin probability and tolerance.\\[2pt]
Residual budget & $u=r/[10^4(m+1)^2]$, heavy mass $u/n$.
Always include the zero predictor.\\[2pt]
Suffix and signal & Take visible length $l=1+\lceil C_l\log U\rceil$
and apply (S3) at $l-1$. If the mean squared signal at the designated start exceeds $r/8$, some suffix in (R2) has
$|K|\ge k_0=c r/n^2$ for an absolute sufficiently small $c>0$ and
large $n$.\\[2pt]
Entrance threshold & Put $z=k_0u/[48(n+1)]$. Grid spacing $z$
for the scaled threshold, and $z/(P+1)$ for $\sigma$, where $P$
is the longest tested prefix below.\\[2pt]
Replacement windows & Fix $d_0=C_\beta\lceil\log n\rceil$ for
the one-sided entrance test. Take $t\ge 2d_0$ with the survival
remainder after $t-2d_0$ divided by the inverse-polynomial witness
mass at most $u$. Take $g$ with (R1) error at most
$c' r u/n$ for a sufficiently small $c'>0$. Both are
$O_\beta(\log U)$. Put $P=2g+t$.\\[2pt]
Unused middle & Require $n^{C_a}\exp[-c_a(j-P-l)]\le z$ for
(R3), $j\ge2l$ for the signal estimate, and $j\ge g+2t$.
If a condition fails, $j=O_\beta(\log U)$
and all $2^j$ values can be fitted directly.\\[2pt]
Word bins & Use prefixes of length at most $P$ and suffixes of length
$l$. Each probability $p\ge 2^{-(P+l)}$. For $B$ queried
bins, $O(2^{P+l} z^{-2}\log(2B/\delta))$ records suffice for
simultaneous absolute error $z$.\\[2pt]
Terminal readout & At most $H=1+(1+2^g)2^{2t+g}$ keys. Estimate
each mass and unnormalized response mean to $r/(256H)$ on fresh
records. Fit values in $[-1,1]$.\\[2pt]
Selection & Compare $L$ bounded residual predictors by independent
estimates of $f^2-2Y' f$ with absolute error $r/8$, where $Y'$
is the transformed response. This takes
$O(r^{-2}\log(2L/\delta))$ records per shared panel.\\ \hline
\end{tabular}
\end{table}

To read the exponents in the residual sample rows, write
$P+l\le c_w\log_2 U+O_\beta(1)$ and
$g+t\le c_h\log_2 U+O_\beta(1)$. The displayed threshold gives
$z^{-1}=O((n+1)U^2)$, so one word-bin panel uses
$O_\beta((n+1)^2 U^{c_w+4}\log(2B/\delta))$ records. The number of
terminal keys satisfies $H=O_\beta(U^{2c_h})$, giving a readout
panel of $O_\beta(U^{4c_h}r^{-2}\log(2HL/\delta))$ records.
The window coefficients $c_w,c_h$ depend on the structural constants
chosen at the target exponent; these are per-panel bounds before
candidate enumeration and validation.

In the clean rows $v,k_s$ are polynomial in $n$ and $h_*$ is fixed
by $\beta$. The pattern matrix sums at most $2k_s+1$ matrices with
coefficients counted in absolute value. Its operator error is at most
$(2k_s+1)v e_*<\lambda_*/8$, whereas a present pattern has a negative
direction of size at least $\lambda_*/2$. In calibrating the history
ratio, an error $e$ in the two $B$ estimates costs $O(e/b_*)$; each
subsequent diagonal multiplication costs at most $O(n)$ on its two
arguments and $O(n^2)$ on the ratio. The displayed slack covers the
fixed recursion (increase $C_*$ for numerical constants), including
errors in the observable centering scalars. The population bias is
set by the burned-row bound before estimating word bins; no dimension
of the signature dictionary depends on this later burn. The clean
rows use the binary core alone; include both constant classifiers.
For the disjoint branch all these estimates take place inside the
corresponding prefix bin with its own padding.

Here and below each batch is independent of the quantities defining its
bins and candidates, or covers a fixed list of them simultaneously.
Specifically, for an observable bin of probability $p$ use
$Y'\mathbf1_{\mathrm{bin}}/p$. Its variance is at most $1/p$,
so a median of independent group averages gives the displayed
bin cost. The groups need only constant success probability before
amplification. For the terminal readout, conditional on the
learned key map, the mass $p_i$ and unnormalized response $b_i$
enter its risk as $\sum_i(p_i f_i^2-2 b_i f_i)$ plus a constant.
Absolute errors $a$ in both quantities cost at most $6Ha$ in
excess risk for a minimizer of the estimated score. This fit uses
$O((H/r)^2\log(2HL/\delta))$ records and allows keys of arbitrarily
small mass. The validation score has second moment at most ten.
Bounds for direct full-word regression follow from the same key
calculation with $H=2^j$.

For example, a designated start $s$ and another state $x$ can satisfy
$f_0(s)=f_0(x)=s$ and $f_1(s)=f_1(x)=x$. Distinct choices of
$x$ give disjoint events, of total probability $(n-1)/n^4$.
With opposite accepting bits the label of every nonempty word is
determined by its last letter. Every constant classifier has error
$1/2$ on this event. This action is among those enumerated; its
entire response already lies in the coarse suffix space.

For the signal row, let $s_*=\sigma^{j-l}H_1^T(s_0)\le2$ for
the $j$-letter detail task. If $\mathbb E_W[h(f_W(s_0))^2]>r/8$ for a uniform $j$-letter word $W$,
then (S1) gives $s_*\sigma^l\mu_1^T(h^2)\ge r/16$, hence $s_*\ge r/16$.
Apply (S3) at $l-1$ and multiply by $s_*^2$.
As $0<\sigma\le1$, the leading lower bound on the mean of the squared
suffix-bin means is at least $n^{-3}(r/16)^2$. Errors from (S3) and
the prefix projection in (S1) are smaller than half this bound by
the choice of $C_l$ and $j\ge2l$. Some length-$l$ suffix therefore has
absolute population mean at least $c_1n^{-3/2}r$. At the root (R3)
approximates this mean by $K H_1^T(s_0)$, with $H_1^T(s_0)\le2$;
this gives the weaker floor $k_0=cr/n^2$ used by the grid. Choose a true boundary in $[u,2u]$
at distance $u/[3(n+1)]$ from every $H_1^T(y)$. Mixing bias, bin
estimation, rounding the threshold, and rounding $\sigma^{|p|}$
each cost at most $z$ in (R3) (clip bin means to $[-1,1]$).
Consequently one enumerated branch recognizes all tested entrance
signatures. The block replacement coupling of
Appendix~\ref{sec:closed-edit-resilience} charges at most $32u$ per
block in squared loss: at most $3u$ for an early hit, $3u$ for a
missing heavy representative, $12u$ after simultaneous entrance and
$4u$ for distinct bright survivors, with room for the spectral
remainders. The $L^2$ triangle inequality over at most $m$ blocks
then gives comparator risk at most $32m^2u<r/16$.
For a low-signal task the zero comparator has risk at most $r/8$.
Readout fitting and selection give risk at most $r$ in either case.
Fit and validate each detail separately before summing them for the
trial action; no Cartesian product of their candidate lists is needed.
The number of thresholds, rates, suffixes, keys and their bit lengths
is polynomial in $U$. Enumerate the positive-signal branch even when zero would suffice and
try both geometry branches on every table. Cap every fitted list
and numerical scan at its prescribed polynomial budget, including
on wrong action guesses. No sampled quantity uses the unknown
threshold or a physical representative.

The detail vectors are orthogonal and have squared norm at least $1/4$.
On simultaneous success the sum has squared error at most
$4Jr=\epsilon/8$, and clipping can only improve it. Thresholding at
$1/2$ gives a binary candidate with classification risk at most
$\epsilon/2$. Allocate at most $\gamma/2$ to construction by a union
bound over the polynomial lists of fits and actions, using independent
batches where a fitted choice is reused. Only logarithms of these list
sizes enter confidence amplification. The fresh final classification
panel in Appendix~\ref{sec:iid-clean-completion} costs a further
$\gamma/2$ in failure. Since the table exception is $o(n^{-\beta})$,
taking $\gamma=n^{-\beta}/2$ gives total failure at most $n^{-\beta}$
for all sufficiently large $n$.

\paragraph{A finite-distance bound at fixed confidence exponent.}\label{sec:coarse-distance-proof}

For every fixed $A>0$, there is a constant $C_A$ such that the maximum
finite directed distance in the i.i.d. binary table is at most $C_A\log n$
apart from probability $O_A(n^{-A})$.

\begin{proof}
In a reverse search at a target, let
$m$ be the number of processed targets and $z$ the current frontier.
Every unseen source has two independent cells uniform outside the $m$
processed targets. Thus its next frontier is binomial with mean
$z(1-u)(2-u)$, $u=z/(n-m)$. Until a frontier reaches
$K_0=\lceil k\log n\rceil$, and during $J_0=\lceil J\log n\rceil$
rounds, its mean is at least $1.8z$ for large $n$.
For $\theta=1/5$ and
$a_0=1.8(1-e^{-\theta})-\theta>0$, killing at zero or $K_0$ gives
probability at most $e^{\theta K_0-a_0J_0}$ of remaining positive
and small. Choose $J$ sufficiently large for any desired polynomial
exponent. Extinction exhausts the in-component.

After reaching $K_0$, require growth by at least $1.4$ until a frontier
has size $n/100$. On previous success, processed size is at most
$J_0K_0+2.5z$, so its mean remains above $1.8z$. Chernoff bounds a
growth failure by $e^{-z/25}$; choosing $k\ge25(A+6)$ makes their
union, and then the union over targets, sufficiently small.

Stop before querying the large frontier $F$ as reverse targets.
Conditional on this transcript, each unseen source's cells are still
independent uniform outside the processed targets. From a prescribed
outside start, every fresh outward probe before a hit has probability
at least $1/100$ of hitting $F$. Exhaustion without hitting $F$ implies
the target is unreachable. Otherwise $O_A(\log n)$ misses have probability
$O(n^{-A-5})$. A hit within that many fresh probes yields a path of
that length plus the reverse depth. Union over starts and targets proves
the result. All conditioning consists of explicit cellwise exclusions;
there is no conditioning on a global graph event inside a product-law
claim.
\end{proof}
In a unique recurrent class, a path of this length from a stationary
atom of mass at least $1/n$ gives $\pi_{\min,+}\ge n^{-1}2^{-C_A\log n}$.

\section{Supporting estimates}
\subsection{The rate of a surviving pair}\label{sec:labeled-synchronization}
For every fixed $A>0$ there exists an integer $b(A,r)$ such that,
except on an original event of probability $o(n^{-A})$, every
table without a terminal component on at most $b(A,r)$ states
satisfies, for each fixed $c<1/2$,
\begin{equation}\label{eq:global-rho}
                 n(1-\rho(Q))=1+O_{A,r,c}(n^{-c}).
\end{equation}
Here $Q$ acts on ordered distinct pairs killed at their first common
endpoint and $\rho(Q)$ is its spectral radius. Row-injective tables
inherit this estimate by conditioning.

For clarity, the centered proof for fixed $r$ uses
$\ell=\lfloor(\log_r n)/8\rfloor$ in the bounded-surplus search.
There are $O_r(r^\ell)$ slots and the same surplus tail.
In the local part reaching a specified point, all but that point
have at least one represented slot and at most $K$ vertices have
all $r$ slots represented, using the excess count of
Appendix~\ref{sec:lifted-suffix-all}. The exit chance per $K+1$
letters is at least $r^{-(K+1)}$. At logarithmic burn this gives
the same safe-history heat protocol. In the four-source anchor
calculation replace $2^{-\max(k,h)}$ by $r^{-\max(k,h)}$, since the
two independent words must share their last $\max(k,h)$ letters.
Its leading constant for both moments is
$\sum_{i\ge0}(2i+1)r^{-i}=r(r+1)/(r-1)^2$. There are at most two
distinct letters requested at each frontier vertex, one per
replica; all new-cell probabilities and the forest subtraction
are unchanged. The one-surplus typical anchor uses
$\ell_0=\lfloor(\log_r n)/5\rfloor$, $O_r(r^{\ell_0})$ slots,
and sufficiently long $d=O_r(\log n)$; repeated offset words cost
$r^{-d}$. Thus the same polynomially small typical Frobenius
norm supplies the median. The Hamming extension, fiber bounds
and concentration of Appendix~\ref{sec:centered-edits} apply.
Only table geometry, with no sign or mask condition, is needed.

Consequently $Q^{jR}=\lambda^j\Pi+E^j$ with $\lambda=1-o(1)$,
$\|\!E\!\|\le n^{-\kappa/2}$, and a nonnegative rank-one
projection with normalized common left row $\nu$, by the resolvent
argument in Appendix~\ref{sec:approx-sink}. This leading eigenvalue
of $Q^R$ is simple and isolated, so $\nu Q=\rho\nu$ with
$\rho^R=\lambda$: commutation preserves the eigenspace, positivity
gives the positive root, and all eigenvalues of $Q$ have modulus
at most $\lambda^{1/R}$. With $g=\mathbf1-Q\mathbf1$,
$\nu g=1-\rho$. Give two independent uniform starts a common
random word of length $t=jR$ for a sufficiently large fixed $j$.
Their unnormalized surviving row $u_t$ then obeys
$\|u_t-(u_t\mathbf1)\nu\|_1\le n^{-4}$ by the rank-one power estimate in Appendix~\ref{sec:approx-sink}.
The uniform-history estimate in
Appendix~\ref{sec:history-tree} gives
$u_t\mathbf1=1-O(t/n)$ and
$n u_tg=1+O(n^{-c})$, by using its one-step distinct-to-equal
bound for the common history and averaging. Dividing yields
\eqref{eq:global-rho}. Choosing a slightly larger exponent than
$c$ in the input estimate absorbs all logarithmic factors.

\subsection{Short suffix tests on all predictive classes}
\label{sec:lifted-suffix-all}
On a fixed labeled automaton, $x\equiv y$ if their accepting responses
agree after every finite word, including the empty word. This is a right
congruence. In this section the dictionary is specified before either
the transition table or the accepting mask is drawn.
\begin{theorem}\phantomsection\label{thm:suffix-all}
For every fixed $A>0$, there exists an explicit polynomial-dimensional
dictionary of products of at most $k_A$ signs at the ends of words of
length at most $J_A\log n$. On the entire original binary model,
outside probability $o(n^{-A})$, the unit-row dictionary on the
distinct predictive classes has least singular value at least $n^{-C_A}$.
The dictionary is specified before drawing the table and its one
shared fair mask. The bounds are for fixed $A$.
\end{theorem}

\begin{proof}
Take $j=\lfloor(\log_2 n)/8\rfloor$ and a fixed integer $K$ with
$3(K+1)/4>A+2$. Except with probability
$O_K(n^{1-3(K+1)/4})$, every depth-$j$ outward search has at most
$K$ surplus arcs. Indeed a search uses $O(2^j)$ probes and states.
At least $K+1$ nondiscoveries require as many fresh named-target
hits; a union over probes and all roots gives the bound.

For $j$-letter continuations on both branches, let $v_x$ have coordinates
$2^{-j}\xi_{f_{0w_0}(x)}\xi_{f_{1w_1}(x)}$, so $\|v_x\|=1$.
Its Gram entry is $K_{xy}=\prod_{a=0}^1 G_j(f_a(x),f_a(y))$,
where $G_j(u,v)$ is the common-word expected product of the two signs.

Let $T$ be the union of all terminal strongly connected components
of size at most $K$ in the full table. For any given start, endpoint
$z\notin T$ and its depth-$j$ search, keep only the vertices with
a path to $z$ in this finite search graph, and all the search slots
between them. If $z$ was never discovered the assertion below is
immediate. Otherwise the remaining subgraph is connected: its
vertices can all reach $z$. It has $E-V\le K-1$ by the cycle rank
bound for subgraphs of the connected search. All its vertices except
possibly $z$ have at least one slot staying inside. Therefore at most
$K$ vertices have both slots inside, since the number with two such
slots is $E-V$ plus the number with none. If there were no way to
take an unrepresented slot from a given vertex, its whole forward
orbit would be closed and consist of fully represented rows. It
would contain a terminal component on at most $K$ vertices. Every
one of these vertices can reach $z$ inside the subgraph, so the
closed component contains $z$, a contradiction. A shortest route to
an unrepresented slot has length at most $K+1$ including that slot,
because all earlier vertices are distinct fully represented rows.
A random $(K+1)$-block thus leaves the represented slots with
probability at least $2^{-(K+1)}$. Every actual length-$j$ path
from the start ending at $z$ stays in the represented slots throughout:
all its sources appear before depth $j$ and their slots were explored.
Consequently, uniformly over all starts and all $z\notin T$,
\[
  P^j(x,z)\le (1-2^{-(K+1)})^{\lfloor j/(K+1)\rfloor}
       \le C_K n^{-b_K},\qquad
  b_K=\frac{-\log_2(1-2^{-(K+1)})}{8(K+1)}>0.
\]

Choose an integer $M>A+2$. If $|T|>M$, a union of its components
has size in $[M+1,M+K]$. Each specified size-$t$ set is closed with
probability $(t/n)^{2t}$. A union over these fixed sizes gives
$\Pr(|T|>M)=O_{M,K}(n^{-M-1})$.

For a closed set $U$ with $|U|\le M$, collapse its states to a single
absorber (keep the original table if $U$ is empty). On ordinary
sources the independent slots, conditional on closure, hit each
specified ordinary state with probability $1/n$ and the absorber
with probability $|U|/n$. Write $\bar D_j$ for the equality
probability of two collapsed endpoints and
$\bar m_{xy}=\prod_a\bar D_j(\bar f_a(x),\bar f_a(y))$, evaluating
inside states at the absorber. For each center and $d$ distinct
ordinary neighbors, all different from the center, the conditional
expectation of the product of these $d$ weights is
$O_{M,d}(j^{4d}n^{-2d})$. To see this, run the $2d$ branch contrasts
used in the star calculation. Mark the absorber and all starting
states at time zero. The two initial outgoing slots of each ordinary
neighbor supply singleton coefficient vectors in its two respective
contrast coordinates, hence the complete initial proposal rank is
$2d$, even when the center is already at the absorber. Track rank on
newly discovered ordinary frontier states. Previously exposed cells
and the absorber cannot lead to a newly discovered state. As in
\eqref{eq:poly-rank-return}, each real fresh return can reduce this
rank by at most one. At final collapsed equality all coefficients
are zero, so at least $2d$ fresh hits are required. With $O_d(j)$
slots and states, each hit has conditional probability
$O_{d,M}(j/n)$, including the possibly nonuniform absorber.
No condition on the outside table beyond closure was imposed.

Choose $0<a<\min(1/8,b_K/8)$ and $d$ with $d(1-a)>A+2$.
A union bound over centers and $d$ ordinary neighbors gives
$O_{M,d}(j^{4d}n^{1-d(1-a)})$ conditional probability of a
star with all edges $\bar m_{xy}>n^{-a}$.
To justify using the table-selected $U=T$, sum this bound times
$(t/n)^{2t}$ over all candidate sets with $t\le M$, including
the empty set. Their total expected number is $O_M(1)$.
On the resulting event, every physical state has at most $M+d-1$
heavy neighbors relative to the actual collapsed $T$.

For a fixed table and two starting states, separate the signed
correlation $G_j$ into outcomes where the collapsed endpoints agree
and all remaining outcomes. The first term has absolute value at most
$\bar D_j$, even for different inner signs. The remaining coefficient
matrix is supported on distinct physical endpoints, at least one
of which is outside $T$. Each such ordered coefficient is bounded
by $C_Kn^{-b_K}$ by the marginal estimate above. Their total is at
most one. Its symmetrization thus has squared Frobenius norm at most
$C_Kn^{-b_K}$ and zero diagonal. Conditional Hanson--Wright for the
original one fair mask, followed by a union over starting pairs,
bounds the remainder everywhere by $n^{-2a}$ with superpolynomially
small failure. Consequently every light base Gram entry has absolute
value at most $2n^{-a}$ for large $n$.

Set $B=M+d-1$. The class representatives have degree at most $B$
in this heavy-edge graph. Their shortest distinguishing words have
a separate estimate which holds conditional on every transition table.

\emph{Short witnesses.} For any fixed table with a fair independent
mask, except with probability $n^2 2^{-L}$ every inequivalent pair
is distinguished by a word of length at most
$L=\lceil(A+4)\log_2 n\rceil$. For a fixed pair grow the span of
$e_{f_w(x)}-e_{f_w(y)}$ for $|w|\le l$. If it stops growing it is
invariant under both letters, hence contains all longer differences.
Otherwise its rank at $L$ is at least $L$ (for a distinct pair).
Orthogonality of all these incidence differences to a fair mask has
probability exactly $2^{-\mathrm{rank}}$, since signs must be constant
on each connected component of their endpoint graph. This is the
random-final-state argument underlying the average-case analysis of
Moore minimization by Bassino, David and Nicaud~\cite[Lemmas~3.2--3.3]{BDN}.

\emph{Dual vectors for all predictive directions.} Let $t_x$ be the normalized vector consisting of one constant
coordinate $1$ and the coordinates $2^{-|w|}\xi_{f_w(x)}$ for
$|w|\le L$. Its squared norm before normalization is
$S_L=1+\sum_{l=0}^L 2^{-l}=3-2^{-L}$. This choice keeps substantial
weight on the few shortest tests.
Take the expanded unit row
\[
 V_x=v_x^{\otimes p}\otimes t_x^{\otimes B},
 \qquad p>2/a.
\]
All rows are identical on a predictive class. Consider any set of
representatives and a representative $x$. For each heavy neighbor
$y$ among the representatives, select one short sign distinguishing
it from $x$. The product of the corresponding factors
$(1+\xi_{f_w(x)}\xi_{f_w(z)})/2$, evaluated at $z$, is a selector
$g_x(z)\in\{0,1\}$. It equals one at $x$ and zero on the heavy
neighbors. Realize it as a linear functional of $t_z^{\otimes B}$:
each factor uses the constant and one sign coordinate, and unused
slots read the constant. The coefficient norm is at most
$(\sqrt{3}\,2^L)^B$, since reading a normalized sign at length
$l$ costs at most $\sqrt{3}\,2^l$.
Tensoring with $v_x^{\otimes p}$ gives a dual vector of the same norm
having evaluations $K_{xz}^p g_x(z)$. The resulting square matrix
has unit diagonal and its off-diagonal entries have magnitude at
most $(2n^{-a})^p$, so its operator distance from identity is
at most $n(2n^{-a})^p=o(1)$. The matrix of dual vectors has norm at
most $\sqrt n(\sqrt{3}\,2^L)^B$. Applying these evaluations to the
transpose of the feature matrix therefore gives a least singular value
at least $[2\sqrt n(\sqrt{3}\,2^L)^B]^{-1}$. The selectors and physical identities are
needed only to prove this inequality. The learner can list the
dictionary without knowing any target-dependent choices.
\end{proof}

\paragraph{Interface to terminal-only observations.}
Suppose an adjacent-history statistic approximates the diagonal in
entrywise norm to $n^{-K}$, its loss $q$ is at least $n^{-C_1}$
and its observable calibration and primitive stationary terminal
means can be estimated to $n^{-K}$ at logarithmic histories.
Here $K$ may be chosen as a sufficiently large fixed constant for
the desired final precision. A coordinate of the new dictionary is
a product of at most $k=2p+B$ ordinary sign tests, counting
constant factors for padding. Split these factors into two balanced
groups and use \eqref{eq:poly-product} recursively.
At a node with prefix $v$ this calls the two child estimates for
each $wv$ with $|w|=t$ or $t+1$. At a leaf one terminal mean
or a known constant suffices. With $k$ fixed, the stack has
depth $\lceil\log_2 k\rceil$ and all required suffix lengths are
$O_{A,K}(\log n)$. Clip each estimate to $[-1,1]$. Each level
amplifies child errors by at most $O(q^{-1})$ and calibration error
by at most $O(q^{-2})$. Thus fixed inverse-polynomial precision at
the root costs only fixed inverse-polynomial precision at the leaves.
All coordinates and all needed word bins number polynomially many;
unit row normalization may be replaced by a dyadic factor between
one half and one times the unit normalization. The true predictive
quotient has well-defined deterministic letter transitions and terminal
readout. This constructs the tests given the stated diagonal approximation. A bound on the probability of obtaining that approximation is still needed to control failure over the random target.

\subsubsection{A simultaneous lower bound on meeting mass after short prescribed histories}
\label{sec:log-history-loss}
To normalize the passive estimator, we need an inverse-polynomial lower bound on the observable diagonal fraction. We obtain it by counting meetings after two prescribed histories, which may differ.

\Needspace{9\baselineskip}
\begin{lemma}\phantomsection\label{lem:cross-birthday}
Fix $C>0$ and $T=\lceil C\log n\rceil$. With failure
$\exp[-\Omega_C(n/\log^4 n)]$, simultaneously for any words
$u,v$ of lengths at most $T$ and either two last letters $a,b$,
there are at least $c_0 n/(T+1)^4$ distinct ordered pairs $(i,j)$
satisfying
\[
       f_u(i)\ne f_v(j),\qquad
       f_a(f_u(i))=f_b(f_v(j)).
\]
Here $c_0>0$ is absolute. Each group of starting states can additionally
be chosen without repetitions among the trials constructing the witnesses.
\end{lemma}

\begin{proof}
Set $k=\lfloor n/[100(T+1)^2]\rfloor$. Make $k$ attempts with
the first word, then $k$ with the second. In each attempt select an
unnamed start by a fixed rule and follow the prescribed word. Name
each fresh answer immediately and abort at the first already named
vertex, including one visited on this attempt. Every still-active
source is newly named and has unqueried outgoing cells. At all times
there are at most $2k(T+1)$ named vertices. Thus conditional on any
past, the failure probability of a trial is at most
$2kT(T+1)/n\le1/50$. Conditional Bernoulli domination gives at
least $k/2$ successes in each group except with probability
$2e^{-\Omega(k)}$. Successful paths are vertex-disjoint, and the
terminal states of all such paths have no outgoing cell exposed.
Even a later failed trial that visits an earlier endpoint stops
before querying it. The argument includes empty words.

Expose letter $a$ on the first group's successful endpoints.
The probability of any one head duplicating a previous head
on this list is at most $k/n$. Except with probability $e^{-\Omega(k)}$
there are at least $k/4$ distinct heads. At all second-group
endpoints the requested $b$-cells remain independent uniform,
also when $a=b$. Each hits a first-group head with probability
at least $k/(4n)$. Chernoff gives at least $k^2/(16n)$ successful
hits except with probability $e^{-\Omega(k^2/n)}$. These yield
distinct ordered witness pairs since their second starts are
distinct; a first start can be reused. A union bound over fewer
than $4^{T+2}$ choices of words and letters completes the proof.
\end{proof}

Let $\pi$ be a stationary law and assume the uniform row $\lambda$
satisfies $\|\lambda P^G-\pi\|_{\rm TV}\le n^{-3}$ at a logarithmic
time $G$. For every prescribed common history $v$ of length
$t\le T-G$, two i.i.d. $\pi$ starts have probability at least
$c_1/[n(T+1)^4]$ of distinct endpoints after $v$ and a meeting
on the next prescribed letter. Indeed first give two independent
uniform starts their independent $G$-letter burns. For each pair
of burn words $u_1,u_2$ apply Lemma~\ref{lem:cross-birthday} to
$u_1v,u_2v$, then average. Replacing the burned initial product
law by $\pi\otimes\pi$ costs at most $2n^{-3}$. Thus any subsequent
normalization by stationary survival probability only increases the
bound; in particular $1-H_{t+1}/H_t\ge c_1/[n(T+1)^4]$ after
averaging the history and next letter. Neither a table-selected
initial pair nor independence of different meeting events is assumed.
The exponent of $n$ here is independent of the fixed constant in
the logarithmic history length.

On the same log-word scale the uniform short-word fiber bound also
gives a polylogarithmic upper bound times $1/n$ on the unconditional
one-step meeting mass. If every relevant fiber has size at most $M$,
uniform independent starts assign at most $M^2/n^2$ per distinct
endpoint pair. The number of pairs sharing a target under two fixed
letters is at most $nM$ by the single-letter indegree bound. Burn,
average, and transfer as above. A survival denominator bounded below
by a constant gives the same upper bound for the conditional loss.

For perspective, the one-group sampling step also gives rank
$\Omega(n/(T+1)^2)$ simultaneously for all $T$-letter words when
$T\le c n^{1/3}$ and $c>0$ is sufficiently small. Take a union
over $2^T$ words against $e^{-\Omega(n/T^2)}$.
For every fixed $\delta>0$, Chapuy and Perarnau~\cite[Thm.~1.4]{CP}
show that, with high probability, no reset word has length below
$n^{1/3-\delta}$. For calibration we use the simultaneous meeting bound of Lemma~\ref{lem:cross-birthday} over all logarithmic histories.

\subsubsection{Learning an arbitrary accepting set on the clean table stratum}
\label{sec:mask-uniform-clean}
Fix a confidence exponent $A$. On the large-terminal centered-block
stratum outside an original \emph{table} exception $o(n^{-A})$, one
terminal-label learner works for every binary output assignment, every
training length $m\ge1$, and every error $\epsilon>0$. Its sample
budget and bit running time are polynomial in $n,m,1/\epsilon$ and
$\log(2/\gamma)$ for conditional sample confidence $1-\gamma$.
The same table event is used for all these choices.

Set $U=2nm/\epsilon$. For any two states, let $d_l(x,y)$ be their
disagreement probability on a uniform $l$-letter continuation.
The clean spectral block gives a right Perron coefficient
$\psi_{xy}\ge0$ and a common mismatch coefficient $\alpha/2$,
where $\alpha=\nu(1-z_i z_j)$ for the accepting signs $z$. Uniformly
over all masks, pairs and $l$, for absolute polynomial prefactor
and constants $\theta>0$ depending on $A$,
\[
 |d_l(x,y)-\rho^l\psi_{xy}\alpha/2|
             \le n^{O_A(1)} e^{-\theta l}.
\]
For $x=y$ all terms vanish. This follows from the block decomposition
with remainder at most $n^{-\kappa/2}$ per block $R=O_A(\log n)$;
the shorter remaining block is bounded in operator norm by uniform
fibers. If the core has both signs then
\eqref{eq:constant-survivor-gap} gives
$\alpha\ge c\pi_{\min,+}\ge n^{-C_A}$.
Choose $L=C_A'\log U$ sufficiently long that the remainder at $L$
is at most $\eta/2$, $\eta=\epsilon/[32(m+1)]$. If two states have
the same sign signatures for \emph{all} words of length through $L$,
then
\[
                    d_l(x,y)\le\eta\qquad(l\ge0).
\]
For $l\le L$ their responses are identical. For larger $l$ use
$d_L=0$ to control the nonnegative leading term there; its magnitude
can only decrease at future times. This avoids requiring every
physically distinct ordered row to have positive long-word signal.

\paragraph{Identifying short signatures.}
Use the centered two-branch features and heavy-star estimate in
Appendix~\ref{sec:mask-uniform-exact} at a fixed $j=C_A''\log n$.
Only the table-only Gram estimate and star count enter here. At
threshold $n^{-a}$ the failure probability for a star of degree
$d$ is $O_d(j^{4d}n^{1-d(1-a)})$ by
Appendix~\ref{sec:lifted-suffix-all}; choose $d(1-a)>A+2$.
The signature argument below uses each pair's own short tests
and needs no fork-survival event.
The dual selector now ranges over tests of length through $L+1$,
increasing $L$ to exceed $j+2$ if needed. Features are constant
on a signature through $L+1$. Distinct such signatures differ on
one of these tests by definition, even if their long continuations
have tiny discrepancy. The threshold on light centered Gram entries,
the bounded heavy degree and the number of tensor factors require
only the fixed $j$ inputs. A chosen selector has norm
$O(2^{L+1})^B$, so the dictionary on distinct extended signatures
has least singular value $U^{-O_A(1)}$. Its coordinates, computed
from bounded centered short tests, have known normalizations; all
unknown centering constants are ordinary stationary terminal means.
The class weights are at least $\pi_{\min,+}$.

The PSD signature test of Lemma~\ref{lem:exact-recurrent} now
enumerates precisely the extended signatures through $L+1$ in the
support of the recurrent stationary law; it tests only signed
coordinates \emph{constant} on these extended classes.
It uses products of a fixed number of bounded terminal tests, whose
lengths may now reach $O_A(\log U)$. For completeness the adjacent
moment calibration remains accurate on these longer windows.
From two independent stationary starts, the off-diagonal history
row at $l$ is $u_l=u_0Q^l=\rho^l w\nu+e_l$ where $w=u_0\psi$ is
bounded above and below by absolute constants on this stratum:
at a fixed logarithmic burn its mass is $1-O((\log n)/n)$ by
Appendix~\ref{sec:history-tree}. The absolute-entry norm of $e_l$ is
at most $n^{O_A(1)}e^{-\theta l}$. Choose $l$ a whole block of
length $C\log U$ after fixing the desired inverse-polynomial
precision. Then the relative error $\|e_l\|_1/\rho^l$ can be made
smaller than any prescribed fixed power of $U^{-1}$, since
$\rho\ge1-2/n$. In particular for large enough $n$, $\rho^{-l}$
costs at most $U$ after $C$ is fixed. With $B_l=z^\top S_l z$,
\[
  1-B_l=u_l(1-z_i z_j),\qquad
  1-B_l\ge c\rho^l\alpha .
\]
The observable ratio $(1-B_{l+1})/(1-B_l)$ approximates $\rho$ to
the desired primitive precision. Since $1-\rho\asymp n^{-1}$,
$(S_{l+1}-\rho S_l)/(1-\rho)$ approximates the stationary diagonal
in absolute-entry norm with the desired precision, by cancellation
of the leading off-diagonal row and equality of both marginals.
In detail put $D_l=1-B_l$. On this event
$D_l\ge c\rho^l\alpha\ge U^{-C_A}$, and
$|D_{l+1}/D_l-\rho|\le C\|e_l\|_1/(\rho^l\alpha)$
with a constant enlarged for the next row. Raw absolute errors
$\zeta<D_l/4$ perturb their ratio by $O(\zeta/D_l)$.
Choose the exponent of the desired ratio precision exceeding the
final diagonal precision by at least three powers of $U$ to allow
division by $1-\rho$ and perturbation of its denominator. Use
capped empirical guards on both $D_l\ge U^{-C'_A}$ and
$1-D_{l+1}/D_l\ge1/(4n)$, at exponents and constants with slack.
Fix this desired precision first, then fix the lag coefficient $C$.
The balanced recursion, including estimation of the scalar centering
means, loses only fixed powers in $U$. Compute raw means from suffix
bins and unseen random padding at the one training length. Every
required suffix and padding has length $O_A(\log U)$ after precision
choices; if there are too few letters use the whole-word dictionary,
which then has $2^m=U^{O_A(1)}$ entries. A finite initial range of
$n$ can instead use bounded-state enumeration.

\paragraph{An approximate transition system.}
For each distinct signature through $L$ choose any one of its
realized extended signatures. For letter $a$, use its entries at
$aw$ for $|w|\le L$ to move to the resulting short signature.
Use its empty-word sign as output. Every step of this machine can
be regarded as following the true successor of the current physical
representative and then replacing it by a representative with
the same short signature. For a uniform unconsumed continuation of
any length, this replacement changes the final sign with probability
at most $\eta$. A telescoping hybrid over at most $m$ replacements
therefore costs at most $m\eta$ in risk, with one additional replacement for an initial
representative choice, giving at most $(m+1)\eta$ in total. This comparison uses independent uniform
remaining letters conditional on earlier letters and the first
independent fitting panel.

For the prescribed start use $g=O_A(\log U)$ random prefix bits
so its chance to end outside the core is at most $\epsilon/16$,
by ordinary mixing. A true core bin is approximated by one of the
at most $n$ reconstructed states. In every bin select the state of
least empirical loss on an independent panel; using sufficiently
many samples per bin gives excess risk at most $\epsilon/8$ per
bin with joint failure at most $\gamma/2$ by Hoeffding and a
union bound. If $m\le g$, use the dictionary. Include both constant
signs as candidates under fresh validation; these handle a
core-constant mask even when rare exterior words with different
labels occur at the training length. Every numerical algorithm is
capped on failed data and target predicates use only fixed table
windows independent of $\epsilon,m$. This proves the claim.

\subsubsection{Terminal-label learning with joint failure below every fixed
power \texorpdfstring{$n^{-\beta}$, $\beta<1$}{n to the power -beta, beta < 1}}\label{sec:near-inverse}
Fix $\beta<b<1$ with $b$ rational and set $c=\kappa=(1-b)/32$.
We retain the observable arrays, rational ridge inversion and
independent prefix fit in Appendix~\ref{sec:poly-learning-proof}.
Only their windows and precision depend on $\beta$.
Let $\ell=\lfloor c\log_2 n\rfloor$, $d=\lceil12\log_2 n\rceil$,
$r=4d+1$, $w=2d+1$, $R=3w+d$, $j=r$, $N=n(n-1)$ and
$\Lambda=\log^{12} n$. With $K=Q^R$, put
$a=N^{-1}\mathbf1^\top K$ and $B=K-\mathbf1a$.
Write $G$ for the event that each depth-$\ell$ outward breadth-first
tree has at most one extra arc. A union bound over the root and two
hits among $O(2^\ell)$ named vertices gives $\Pr(G^c)=O(n^{-1+4c})$.
On $G$, at most $(\ell+1)2^{-\ell}$ of the probability of an
$\ell$-letter walk can end at a prescribed vertex: specifying the
number of uses of the extra arc fixes the intervening tree paths.
For any $O(\log^4 n)$ exposed states, remaining inside for $d$
steps thus has probability below $n^{-3}$ for large $n$, by applying
the endpoint bound in disjoint blocks.

\paragraph{The two centered estimates.}
Write $f_r(i)$ for the probability of a first meeting at step $r$ from
ordered pair $i$. The exposure estimates used for the pair kernel give
\[
 \mathbb E\left[\ind_G\left(\|B\|_F^2+
       N^{-1}\sum_i(nf_r(i)-1)^2\right)\right]
       =O(n^{-1+4c}\log^{20} n).
\]
Here are the changes needed when $c$ is small. After a $2d$-step
attempt to obtain fresh sources in a surviving pair, the probability
of failed acquisition is $O(d\log^4n/n)+2n^{-3}+d^22^{-d}$ on
observable safe histories. Two attempts and final exposure give a
first-meeting probability at most $1/n+O(\log^{10} n/n^2)$, or
$O(\log^{10} n/n^2)$ for a specified distinct endpoint pair. Replicate
on the same table, checking safety on the partially exposed cells
as in the pair-kernel proof. Two replicas bound the truncated second
moment of $f_r(i)$ by the square of the first upper bound. For the
lower first moment plant the two paths joining only on their last
step; their probability is $(n-2)_{2r-1}/n^{2r}$. Conditional on
the entire planted tree, an outward search with two extra arcs still
requires two fresh hits to named vertices, now including the $O(r)$
planted states. Its all-root failure is still $O(n^{-1+4c})$.
Subtract on the same event $G$ to get the displayed first-hit estimate.
For $B$, the four-source enumeration gives upper row second moment
$6/n^2+O(\log^6 n/n^3)$. Its lower mean-row moment plants both joins
as a forest and applies precisely the same conditional bound,
now with $O(R)$ extra states; it is at least
$[6-O(n^{-1+4c}+R^2/n+R2^{-R})]/n^2$. Subtracting $N$ times
this lower bound yields the displayed estimate for $B$.

Using $\lceil\log^3n\rceil$ replicas for the endpoint bound and a union
over starting and ending pairs gives $\max_{i,j}Q^r(i,j)\le\Lambda/n^2$
on $G$ outside a superpolynomially small exceptional set. The bound
propagates to every later time by prepending transitions. Markov's
inequality now gives, outside $O(n^{-1+4c+2\kappa}\log^{20} n)$,
\[
 \|B\|\le n^{-\kappa},\quad
 N^{-1}\sum_i(nf_r(i)-1)^2\le n^{-2\kappa},\quad
 \|a\|_2\le\sqrt\Lambda/n.
\]
The last inequality uses the entry bound and $a\mathbf1\le1$.
On antisymmetric tests $Q^{kR}$ acts as $B^k$, so the ordinary
walk has a unique stationary law $\pi$ and maximum total variation
distance from it at time $kR$ at most $2n^{1-k\kappa}$.

\paragraph{Stationary mass and diagonal calibration.}
For each fixed $A>0$, all finite directed distances are at most
$C_A\log n$ outside probability $O_A(n^{-A})$. In a reverse search
with $m$ targets processed and current frontier $z$, the next
frontier is binomial on $n-m-z$ unseen rows, with parameter $2u-u^2$,
$u=z/(n-m)$. Up to $J\log n$ rounds with $0<z<k\log n$, its mean
is at least $1.8z$. For $\theta=1/5$, exponential Markov iteration
killed at either threshold makes the probability of staying small
and positive at most $\exp(\theta\lceil k\log n\rceil-
[1.8(1-e^{-\theta})-\theta]\lfloor J\log n\rfloor)$. Choose $J$
large after $k$. From size $k\log n$, require growth by a factor
$1.4$ up to $n/100$. On previous success $m\le O(\log^2n)+2.5z$,
and the conditional failure is $\le e^{-z/25}$; take $k\ge25(A+6)$.
If the reverse search exhausts, all predecessors have been found.
Otherwise, before its large frontier is probed, every outward
fresh slot at an unseen source has probability at least $1/100$ of
hitting that frontier. After $O_A(\log n)$ misses the remaining
failure is $O(n^{-A-5})$, and exhaustion without a hit implies
unreachability. Union over starts and targets. Taking $A=2$ and
following a shortest path from a heaviest stationary state gives
$\min_{x\in\operatorname{supp}\pi}\pi_x\ge n^{-S}$ for a fixed
integer $S$.

Let $S_t=\mathbb E_v[(\pi T_v)^\top(\pi T_v)]$ for $|v|=t$,
$H_t=1-\operatorname{tr}S_t$, and $B_t=\xi^\top S_t\xi$ for the
independent fair signs. For every fixed $T=O_{\beta,D}(\log n)$,
on the preceding mixing event the truncated expectation of $1-H_T$
is $O_{\beta,D}(\log^2n/n)$. Indeed independently burn two uniform
starts for $kR$ letters until their joint law is within $2n^{-10}$
of $\pi\otimes\pi$, then use a common word for $T$ letters. Initially
distinct endpoints can collide only after a fresh named-target hit in
$O(\log n)$ probes. Monotonicity and Markov give $H_l\ge3/4$ up to
$T$ with this same failure order. Since
$\mathbb E_\xi B_t=1-H_t$ and $B_t\ge0$, calibration $B_t\le1/2$
at the chosen $t$ also costs only $O(\log^2n/n)$. For a sublaw
$\mu$ of mass at most one and $l\ge2r$, the heat bound gives
$\|\mu Q^{l-r}\|_2\le\sqrt\Lambda/n$. Its pairing with
$(nf_r(i)-1)_i$ has absolute value $\le\sqrt\Lambda n^{-\kappa}$.
Applied to the stationary off-diagonal sublaw, this gives the next
conditional loss $q=1-H_{t+1}/H_t\in[1/(2n),2/n]$ at our chosen $t$:
the mass at $t+1-r$ lies between $H_t\ge3/4$ and one.

\paragraph{Two-branch separation.}
Use the weights $m_{xy}$ of Lemma~\ref{lem:poly-feature}. On the
physical states greedily retain a matching in decreasing order of
weight. Each omitted edge has a distinct charge to the product of
its own weight and that of an incident retained edge of at least
equal weight. Thus its squared weight is covered by the sum of all
incident-edge products. Equation~\eqref{eq:poly-four-return} bounds
the expected charge by $O(j^8/n)$. Except with this failure order the
omitted symmetric matrix has Frobenius norm at most $1/20$. By
\eqref{eq:poly-three-return}, outside probability $O(j^6/n)$, on
each unequal-row pair at least one of the four grandchildren has
collision probability at most $1/10$. The retained edge weight is
then at most $0.55$ on row representatives. The unit-diagonal
structural matrix restricted to representatives has floor at least
$0.45-1/20$. For the signed perturbation
\eqref{eq:poly-sign-concentration} holds with failure $O(n^{-2})$
by increasing its constant. The expected sum of the squared branch
collision probabilities over distinct roots and letters is $O(nj^2)$
by the one-return bound. At threshold $n^{2-\kappa}$ its failure is
$O(j^2 n^{-1+\kappa})$. With $\eta=O(\log^7 n/n)$ the squared
Frobenius perturbation is now $O(\eta^2n^{2-\kappa}+n^2\eta^4)=o(1)$,
giving $VV^\top\succeq I/4$.

\paragraph{Numerical and sample parameters.}
Set $C=S+3$, $P_*=20(C+D+1)$, $k_0=\lceil4/\kappa\rceil+1$,
$K_*=k_0+\lceil2(8P_*+6)/\kappa\rceil$, and
$t=K_*R$, $h=t+1$, $L=2t+j+5$. A normalized surviving history
row $y$ updates to $z=yK/(yK\mathbf1)$. Denominators on the two
stationary histories are at least $1/2$. After $k_0$ blocks their norms
are at most $5\sqrt\Lambda/n$ by the recursion
$\|z\|_2\le2\sqrt\Lambda/n+2n^{-\kappa}\|y\|_2$.
For a second normalized row $x$ the update difference is
$(x-y)B(I-\mathbf1 z)/(xK\mathbf1)$, giving subsequent contraction
at most $n^{-\kappa/2}$. Comparing histories shifted by one step,
conversion to entrywise norm costs at most $n$ and division by $q$
costs at most $2n$. The diagonal error is covered by the off-diagonal
error because the marginals agree. Consequently
\[
 \left\|q^{-1}(S_{t+1}-(1-q)S_t)-\operatorname{diag}\pi
 \right\|_{\rm ent}\le n^{-8P_*}.
\]
It follows that $S_h\succeq c' n^{-S-1} I$ on the recurrent states
and the observable matrix $H=FV$ has $\sigma_{\min,+}(H)\ge n^{-C}$.
Calibration using $1-B_t\ge1/2$ has deterministic ratio bias at most
$2q n^{-8P_*}$.

Use padding $R\lceil(6P_*+4)/\kappa\rceil$ and primitive terminal-mean
precision $n^{-6P_*}$. Shorter lengths use their full truth table.
For longer lengths the first batch consists of
$O(2^L n^{12P_*}(L+\log(8/\delta)))$ words. Simultaneous bin
concentration, padding and division by $q$ give normalized array
error $\tau\le n^{-4P_*}$. With ridge penalty $n^{-P_*}$
the letter-operator error is $O(n^{-P_*+3C})$. Exact word products
have norm at most $n^{C+1}$ and true readouts norm at most $n^C$;
expanding between error positions as in the original proof bounds
readout error at lengths up to $n^D$ by
$O(n^{D+6C+2-P_*}+n^{2C+1-4P_*})=o(1)$. The uniform margin
therefore exceeds $1/2$. At requested accuracy $\epsilon$, take
a prefix of length $b_\epsilon=R\lceil(2+\log_n(64/\epsilon))/
\kappa\rceil$. Its nonrecurrent mass is at most $\epsilon/8$ and
$2^{b_\epsilon}\le(n/\epsilon)^{O_\beta(1)}$. The independent
prefix fit uses the realizable VC bound with dimension
$(2^h+1)2^{b_\epsilon}$ on the fixed recurrent bins. Its error is at
most $\epsilon/2$ there. If $b_\epsilon\ge m$, use the full
truth table. All guards, array normalizations and rational rounding
use the same bit-complexity bounds as the original algorithm.

The table and mask exceptions sum to
$O_{\beta,D}((n^{-1+4c+2\kappa}+n^{-1+\kappa})\log^{20} n+n^{-2})
=o(n^{-\beta})$. These exceptional events depend only on the target, $D$ and $\beta$. This proves the claimed rate for the fair-label, polynomial-length case with $\beta<1$. The extension in Theorem~\ref{thm:poly-learning} uses Appendix~\ref{sec:iid-clean-completion}.

\subsection{Approximate entrance signatures and second-order confidence}
\label{sec:approx-sink}
Entrance tests identify the quotient exactly when the entrance languages distinguish its different ordered rows. This condition can fail on tables that must be covered at higher confidence. Instead, we only require that replacing
one hidden state by another with the same short signature makes little
difference on an independent continuation. This permits an ordinary
supervised fit on the last few letters.

\begin{theorem}\phantomsection\label{thm:second-order-learning}
Fix $D\ge1$. For each $A>2$ there are a fixed integer $K$ and a
polynomial-time passive learner such that, outside an event of original
labeled-target probability $o(n^{-A})$, it achieves error $\epsilon$
with sample probability $1-\gamma$ at every prescribed $m\le n^D$
whenever the table has a unique absorbing row and has no other
nonempty closed set of size at most $K$. Here ``other'' excludes
its singleton. Its samples and time are polynomial in
$n,1/\epsilon,\log(2/\gamma)$, with constants depending only on
$D,A$. Combining with Theorem~\ref{thm:large-component-confidence}
gives an unconditional target exception at most $(8+o(1))/n^2$.
\end{theorem}

\begin{proof}
Take $K$ greater than $A+4$ and all local search cutoffs for a clean
repair at exponent $A+3$. Except on an original event of probability
$o(n^{-A})$, all finite directed distances in the given table are at
most $h=C_A\log n$, all logarithmic-word fibers for the fixed horizons
below are polylogarithmic, and no closed set has size in
$[K+1,n/4]$. For the last assertion sum
$\binom ns(s/n)^{2s}\le(es/n)^s$. For fixed $K<s\le2K+10$ the
sum is $O(n^{-K-1})$, for larger $s\le\sqrt n$ use
$(e/\sqrt n)^s$, and for the remaining range use $(e/4)^s$.
On the target stratum every exterior state can therefore reach more
than $n/4$ states, since its reachable set is closed.

\paragraph{Repair witness.}
Let $x$ be the absorbing state. Replace only its row by independent
uniform cells, obtaining a table $B$; give both tables the same
accepting signs. Conditional on a specified absorbing row in the
i.i.d. model, this repair has exactly the original clean law. For a
fixed target on the stratum with the fiber bounds, at most
$O_K((\log n)^{O(1)})$ states can reach $x$ in at most $K$ steps:
sum its fibers over the bounded set of such words. A new head outside
this set can reach more than $K$ states within $K$ steps in $B$.
Its first $K$ search levels agree with those in the original graph, where the reachable set has more than $K$ states.
No closed set of at most $K$ states avoiding $x$ can be created.
Consequently with conditional repair probability $1-o(1)$ the clean
table has no small terminal component.

Call a repair good if it has this property and passes the high-confidence
centered block and clean learner conditions at exponent $A+3$, the
ordinary diameter and fixed-horizon fiber events, and the fair-mask
balance below. Each of the additional failures intersected with
the no-small-terminal condition has original probability
$o(n^{-A-2})$. The mask has stationary mass at least $1/3$ on each
sign with superpolynomial conditional confidence because the clean
maximum stationary mass is at most $n^{-1}(\log n)^{O(1)}$.
The probability, on the specified stratum, of having no good witness
is at most twice the joint repair probability of a failure with no
small terminal component. Multiply by $n^{-2}$ for the prescribed
absorbing row and sum over $x$. This bounds the probability that no good repair exists. The learner uses only observations from the original table. In the sequel
$P,Q,\pi$ refer to a good clean repair, and $\mathcal K$ to its
recurrent support. Its bounds hold for every prescribed initial state.
All unmarked transitions agree with the observed target.

\paragraph{Exterior label gap.}
Let $L=P|_{[n]\setminus\{x\}}$ be the killed ordinary walk and
$S_j=L^j\mathbf1$. Choose $G=O(\log n)$ so that every clean row
at any time $\ge G$ is within $d=n^{-8}$ of $\pi$ in total
variation. In this paragraph extend $S_j$ by zero at $x$, including
$j=0$. Put $U=\{z\ne x:S_G(z)\ge1/2\}$. The complement has
polylogarithmic size: summing over initial states, the expected number
of visits to $x$ by time $G$ in the clean chain is at most $(G+1)M$
by the short-word fiber bound $M$. Thus $\pi(U)=1-o(1)$.
For any $z\in U$, condition on the first $G$ steps surviving. In
the next $G$ steps the probability of hitting $x$ is at most
$e=2G(\pi_x+d)=o(1)$, since the unconditioned clean endpoint at
each of these times has mass $\le\pi_x+d$. Accordingly the row
$L^{2G}(z,\cdot)$ has mass at least $1/3$ and after normalization
is within $e+d$ of $\pi$. The same statement applies to any
initial distribution with $G$-step survival at least $1/2$.
A law within $e+d$ of $\pi$ has this survival by a union bound
$e+d+(G+1)\pi_x$. Iteration, and a terminal block of fewer than
$2G$ steps, show that from $z\in U$, at every $j\ge2G$ the normalized
surviving endpoint law is $o(1)$ from $\pi$, uniformly in $j$.
For the terminal block one may couple the input law to stationarity
with cost $e+d$, and stationary hits cost at most $(2G+1)\pi_x$.

We record the accompanying survival comparison. For
$M_0=L^{2G}$, $f\ge0$, $H=\max f$, and $v=\pi f$, domination
by $P^{2G}$ gives $\max M_0 f\le v+dH$. Safe rows give
$\min_U M_0 f\ge(v-(e+d)H)/3$ whenever the latter is positive.
Hence $\pi f\ge H/8$ implies both
$\pi M_0 f\ge (\max M_0 f)/8$ and
$\min_U M_0 f\ge (\max M_0 f)/8$, for all large $n$.
The initial columns $S_j$, $0\le j<2G$, have
$\pi S_j\ge1-(2G+1)\pi_x$. Iteration gives
$\min_U S_j\ge\max S_j/8$ for every $j\ge2G$.

From any exterior $y$ at least one state in $U$ is reachable in
at most $h$ steps. A path to it avoids the absorbing row. Following
this chosen path, then using survival comparison and mask balance,
shows for every $r\ge h+2G$ and either sign $b$ that
\[
 \Pr_y\{\text{survive to }r,\ \xi_{f_{W_r}(y)}=b\}
       \ \ge\ 2^{-h}(\max_z S_r(z))/32 .
\]
If the constant prediction $\xi_x$ has error greater than
$\epsilon/8$ from the designated start $s$ at length $m$, this
is at least $2^{-h}\epsilon/256$ for every such $r\le m$.

\paragraph{A suffix distinguishes a physically persistent pair.}
First suppose $x\in\mathcal K$. Put $N=n(n-1)$. On the good
repair $K_0=Q^R=\mathbf1 a+E$, $\|E\|\le n^{-\kappa}$,
$\sqrt N\|a\|_2\le(\log n)^{O(1)}$, and
$a\mathbf1=1-o(1)$. There are a nonnegative row $\nu$ of mass
one and an absolute constant in the exponent such that for every $j\ge1$,
\[
 \max_i\|Q^{jR}(i,\cdot)-(Q^{jR}\mathbf1)_i\nu\|_1
      \le 2n(n^{-\kappa/2})^j .
 \tag{*}
\]
Here are details of the elementary rank-one argument. Set
$u=\mathbf1/\sqrt N$, $v=\sqrt N a$, $\alpha=vu$.
The scalar map $z\mapsto\alpha+v E(zI-E)^{-1}u$ maps
$[\alpha-1/8,\alpha+1/8]$ into itself and is a contraction,
by the Neumann series and $\|v\|=n^{o(1)}$. Its fixed point
$\lambda=1-o(1)$ has right and left eigenvectors
$(I-E/\lambda)^{-1}u$ and $v(I-E/\lambda)^{-1}$.
Their product, divided by their scalar pairing, gives the rank-one
projection $\Pi$. The same series yield
$\|\lambda\Pi-uv\|\le n^{-\kappa+o(1)}$.
Then $K_0^j=\lambda^j\Pi+(K_0-\lambda\Pi)^j$, with
$\|K_0-\lambda\Pi\|\le n^{-\kappa/2}$. Taking the limit of
$\lambda^{-j}K_0^j$ proves $\Pi\ge0$. All its nonzero rows
have the same normalization $\nu$. Passing to row absolute mass
costs at most $\sqrt N$; replacing its scalar coefficient by the
actual survival mass at most doubles the error. This proves (*).

Write $g_x(i,j)=\mathbf1\{\text{exactly one of }i,j=x\}$.
Use two independent $\pi$ starts with a common length-$t$ history,
$t=j_0 R$ for a sufficiently large fixed $j_0$. Their off-diagonal
mass at $g_x$ is $2(\pi_x-S_t(x,x))$, where here $S_t(x,x)$
denotes their joint diagonal entry. For every deterministic such
history its row $\pi T_v$ has largest atom $o(1)$ by the clean
fiber bound. Its mean over histories is $\pi$. Thus
$S_t(x,x)\le o(1)\pi_x$. Since $\pi_x\ge n^{-C}$ by diameter
and (*) applies to the off-diagonal subprobability, we obtain
$\nu g_x\ge\pi_x$ for all large $n$.

Let $H_y(v)$ indicate entrance into the absorbing state by the end
of $v$ on the observed table, including time zero. If $y,z$ have
identical $H$ signatures on all words through length $t=jR$, then
their clean trajectories cannot have exactly one endpoint at $x$
at time $t$: the earlier first visit would distinguish them and a
simultaneous visit makes the clean paths equal forever. Consequently
(*) bounds their clean survival at time $t$ by
$2n^{1+C}(n^{-\kappa/2})^j$. Their observed endpoints at time $t$
can differ only if the clean pair survives, since any first visits on
that word are simultaneous. Choose
$j=O_{D,A}(1+\log(2m/\epsilon)/\log n)$ so the bound is
$\le\epsilon/(16m)$. Thus $t=O_{D,A}(\log n+\log(2/\epsilon))$.
The proof only needs fixed-horizon fibers at the initial $j_0R$
and block $R$; (*) is powered for all $j$.

\paragraph{Learning with approximate signatures.}
If $m<2h+t+2G+2$, the whole-word dictionary has polynomial domain
size. Otherwise, by collecting both signs or one sign in each prefix
bin through length $h+1+t$, infer all those entrance indicators in
the nonconstant branch. Indeed absorbed bins are pure and every
other bin has both signs with probability at least
$2^{-h}\epsilon/256$, as shown above. Each bin has input probability
at least $2^{-(h+1+t)}$. A union bound therefore needs only
polynomially many examples, including $\log(2/\gamma)$.
Compute the $t$-signatures of all prefixes through length $h+1$.
Every reachable state has an access word of length at most $h$.
Use a shortest access representative of length at most $h$ per signature, and on each
letter go to a representative of the signature at the extended access
word. Use the empty access word as initial representative for its
signature. Impose deterministic caps and defaults on inconsistent data.

Run this machine on the first $m-t$ letters. At each replacement, the
true successor of the current representative and its new representative
have the same short signature. Their probability of giving different
final signs under the remaining independent continuation is at most
$\epsilon/(16m)$ by (*), since it has at least $t$ letters.
For a formal telescoping argument compare, for each $i$, the true
output after simulating $i$ representative steps and then following
the physical table for all remaining steps. Consecutive comparisons
condition only on the prefix through step $i$. Summing the disagreement
probabilities shows that the physical response from the final
representative on the $t$-letter suffix has error at most
$\epsilon/16$. Fit a binary lookup on (representative, last $t$ letters)
by empirical risk minimization on an independent panel. Its domain
size is at most $n2^t$, so Hoeffding and a union bound give
$O((n2^t+\log(2/\gamma))/\epsilon^2)$ samples. This allows signature
collisions and does not require that the finite signatures form a
right congruence.

\paragraph{An exterior absorbing state.}
Suppose instead $x\notin\mathcal K$. Split the first
$g=O_{A,D}(\log n+\log(2/\epsilon))$ letters into prefix bins.
Take $g$ large enough that the clean mass off $\mathcal K$ is at
most $\epsilon/16$, using the ordinary mixing bound from the centered
block at successive powers. Use a whole-word dictionary for $m\le g$.
For every bin train the ordinary clean learner with its unknown
prescribed start on the suffix, together with the two constant
predictors and independent validation. For any prefix that has already
hit $x$ a constant is correct. If it has not hit $x$ and its clean
endpoint lies in $\mathcal K$, the suffix labels are exactly those
of the clean target from that endpoint. The remaining bins have total
input probability at most $\epsilon/16$. The assumptions of the
clean conditional learner hold simultaneously for all initial states
on its target event, since its burn, dictionary and contraction are
uniform. Splitting the sample confidence across $2^g$ bins and collecting
enough independent uniform suffixes in each costs a polynomial factor.
A common independent validation panel chooses between these learners,
the dictionary and constants, with internal accuracies reduced by
fixed factors. All choices use full words with a single final sign.

Finally choose $A>2$ in the large-component and stratum results and
pool their candidates by validation. If neither stratum applies,
a closed set of some size $s\in[2,K]$ exists (two absorbing states
also give such a set). Its probability is at most
$\binom n2(2/n)^4+\sum_{s=3}^K\binom ns(s/n)^{2s}
\le (8+o(1))n^{-2}$. This gives the claimed uniform exception.
\end{proof}

\subsubsection{Uniformity over polynomial word lengths}\label{sec:poly-lengths}
Here are the parameter changes for polynomial lengths in the fixed-schedule version of
Theorem~\ref{thm:poly-learning}. Fix the integer $D\ge1$ before drawing the table.
Use $t=4096DR$, $h=t+1$ and $L=2t+j+5$ throughout this construction.
For fixed $D$ the first-meeting bound through $t+1$ gives $H_l=1-o_{\Pr}(1)$
uniformly for $l\le t+1$ by monotonicity. The normalized histories contract
after the same first 40 blocks. The residual after conversion to the entrywise
norm and division by $q\asymp 1/n$ is
$O(n^{2-(4096D-40)/25})\le n^{-150D}$ for large $n$.
The history and feature singular-value bounds therefore still give $s_*\ge n^{-2}$.
The target events for this one $t$ apply to every $m\le n^D$.

If $m<L+2500DR$, collect the length-$m$ truth table with the coupon bound:
$2^m=n^{O(D)}$. Otherwise the padding bias is at most $4n^{1-125D}$.
With $C2^L n^{242D}(L+\log(8/\delta))$ samples the bin errors are
$O(n^{-120D})$. After calibration and normalization the array errors are
bounded by $\tau=n^{-100D}$. Set $\lambda=n^{-26D}$.
The ridge bias is at most $n^{6-26D}$ and the resolvent perturbation is
$O(n^{-48D})$, so the error of one estimated letter is $a_*=O(n^{-20D})$.
In a word of length $l\le n^D$, the product bound with $B_*\le n^3$ has
$lB_*a_*=O(n^{3-19D})$. After applying the readout, the word and terminal-vector
errors are $O(n^{8-19D}+n^{5-100D})=o(1)$. Thus the same margin holds throughout
this range. Use the same $b$ as above to fit the prefixes, switching to the
truth table if $b\ge m$ since $2^m\le 2^b$. The bit length of a product of $m$ rational matrices is at most $m$ times a polynomial bound per letter. Thus both sample size and bit running time are polynomial throughout the stated range. The confidence bound applies at the length used for training.

\end{document}